\documentclass[a4paper,fleqn]{cas-dc}

\usepackage[numbers]{natbib}
\usepackage{tabularx}
\usepackage{float}

\usepackage{subcaption}

\appendix
\usepackage{wrapfig} 

\def\tsc#1{\csdef{#1}{\textsc{\lowercase{#1}}\xspace}}
\tsc{WGM}
\tsc{QE}
\tsc{EP}
\tsc{PMS}
\tsc{BEC}
\tsc{DE}

\begin{document}
\let\WriteBookmarks\relax
\def\floatpagepagefraction{1}
\def\textpagefraction{.001}

\shorttitle{When Metrics Disagree: A Meta-Analysis of KGC Model Benchmarking}
\shortauthors{H. Gul et al.}

\title[mode=title]{When Metrics Disagree: A Meta-Analysis of Knowledge-Graph-Completion Model Benchmarking}





\author[1]{Haji Gul}[orcid=0000-0002-2227-6564]
\ead{23h1710@ubd.edu.bn}

\author[1]{Ajaz Ahmad Bhat}[orcid=0000-0002-6992-8224]
\cormark[1]
\ead{ajaz.bhat@ubd.edu.bn}

\affiliation[1]{
    organization={School of Digital Science, Universiti Brunei Darussalam},
    addressline={Jalan Tungku Link, Gadong BE1410},
    country={Brunei Darussalam}
}

\cortext[cor1]{Corresponding author}

\begin{abstract}
Evaluating Knowledge Graph Completion (KGC) models has reached a critical challenge due to the fragmented nature of standard performance assessment. Current evaluation still relies primarily on isolated rank-based metrics such as Mean Reciprocal Rank (MRR), Hits$@$k, and Mean Rank (MR), which often yield conflicting model orderings across datasets. For example, a model that leads on MRR may fall behind on Hits@1, and one that performs strongly on one dataset may fail to generalize on another. This fragmentation makes comparisons difficult, enables selective metric reporting, and ultimately obscures real progress in the field. This paper reframes KGC evaluation as a Multi-Criteria Decision-Making (MCDM) problem and introduces a meta-analysis evaluating seven aggregators across five tests: consistency with base metrics, cross-dataset stability, metric independence (leave-one-metric-out), robustness under noise injection, and generalizability (leave-one-dataset-out). To enhance robustness, each test is averaged over leave-one-KGC-model-out (LOMO) and leave-one-group-out (LOGO) removal analyses, ensuring that reliability scores reflect aggregator behavior across diverse model subsets rather than a single fixed collection. Across both tail $(h,r,?)$ and relation $(h,?,t)$ prediction tasks, the results indicate that Pareto optimal analysis effectively identifies optimal aggregators, with Z-score being the most balanced aggregator across both tasks. Of the models we tested, Z-score identified DualE as the top-performing KGC method for tail prediction and FMS (Flow-Modulated Scoring) as the top-performing method for relation prediction. Using the same LOMO and LOGO removals, we further conduct a test-sensitivity analysis that quantifies how much each meta-evaluation test shifts when individual models or whole architectural families are eliminated, revealing that consistency and stability are largely removal-invariant, while generalizability and independence are the most sensitive. The framework resolves evaluation inconsistencies and provides evidence-based guidance for both aggregator selection and model benchmarking in KGC.\\
Follow the anonymous link for the code:
\url{https://anonymous.4open.science/r/When-Metrics-Disagree-4380/README.md}

\end{abstract}


\maketitle

\section{Introduction}
\label{sec:introduction}
Knowledge graph completion, the task of inferring missing relational facts, has advanced rapidly across diverse architectures from geometric embeddings~\cite{sun2019rotate} and tensor factorization~\cite{wang2019relational} to pre-trained language models~\cite{yao2019kgbert} and large language model (LLM)-augmented systems~\cite{wei2024kicgpt,gul2025muco,li2025kermit}. Yet, despite this architectural diversity, evaluation protocols remain fragmented; performance is typically assessed via isolated rank-based metrics (MRR, Hits@$k$ ($k \in \{1,3,10\}$), MR) on individual benchmarks, lacking a holistic comparative framework that can align conflicting results across metrics and datasets.

This fragmented evaluation creates a fundamental challenge. Since each metric captures a distinct aspect of ranking performance, no single metric provides a holistic view~\cite{rossi2021knowledge}. Consequently, model rankings often contradict each other; one model may lead on MRR while trailing on MR; another may dominate FB15k-237 yet underperform on WN18RR. This inconsistency makes it difficult to identify a universally superior model, encourages manual selective procedures, and reduces fairness and reliability~\cite{toutanova2015observed,kadlec2017knowledge}.

Prior work has proposed useful but still incomplete solutions to evaluation inconsistencies. Standardized training protocols help reduce implementation-related noise ~\cite{ruffinelli2020you}, while uncertainty-aware benchmarking~\cite{sun2020benchmarking}, and calibration metrics~\cite{rao2024calibration} deal with randomness in results and overly confident predictions, respectively. Similarly, embedding stability measures assess geometric reliability ~\cite{egger2024relik}. Recently, Gul et al.~\cite{gul2025kg} advanced this direction by introducing KG-EDAS, which employs the EDAS method from MCDM as a unified meta-metric aggregator. However, no existing work validates the use of EDAS or any other MCDM for KGC metric aggregation. The literature lacks a careful analysis of MCDM utility in KGC evaluations and how aggregators are influenced by cross-dataset stability, robustness to metric variation, and generalizability to unseen benchmarks. 

Despite the rapid growth of KGC architectures, including recent hierarchical graph attention networks~\cite{xu2025knowledge}, the critical question remains open: which aggregation strategy offers the most reliable performance assessment, and under what conditions? Beyond methodological fragmentation, the absence of a unified evaluation criterion weakens accurate benchmarking and slows progress in knowledge graph analysis, where reliable model selection directly affects downstream reasoning, recommendation, and decision-making pipelines. Recent efforts in uncertainty-aware benchmarking, embedding stability analysis, and standardized training methods have reduced implementation noise, but they remain isolated and focused on specific metrics. Importantly, no existing framework systematically evaluates how multiple, often conflicting, rank-based metrics should be aggregated into a single, trustworthy ranking. This gap motivates our central research question: Which aggregation strategy yields the most reliable model rankings under realistic evaluation conditions, and how do trade-offs across reliability dimensions inform this choice?

MCDM provides a principled framework from operations research for ranking alternatives under conflicting and heterogeneous objectives~\cite{gyani2022mcdm,zavadskas2016integrated}. Therefore, we propose reframing KGC evaluation as an MCDM problem. Specifically, given $n$ models evaluated on $m$ rank-based metrics, a performance matrix $\mathbf{X} \in \mathbb{R}^{n \times m}$ is constructed, and seven aggregators are applied: EDAS, TOPSIS, VIKOR, MOORA, WASPAS, Borda Count, and Z-score. Critically, the reliability of each aggregator is systematically assessed along five test dimensions to determine the most robust aggregator for each prediction task, which is then used to identify top-performing KGC models. 
In other words, the proposed evaluation has two stages. First, multiple MCDM aggregators combine standard KGC metrics into unified model scores for each prediction task, and the aggregators are assessed using five tests. 
Each of these five tests is itself evaluated under a LOMO and LOGO removal scheme so that the resulting reliability scores are averaged across many model subsets rather than depending on one fixed collection of KGC models. Second, Pareto-based trade-off analysis selects the most balanced aggregator across five tests, which is then used to produce final consensus rankings of KGC models.
This paper makes the following contributions:

\begin{enumerate}
\item \textbf{First comprehensive MCDM meta-analysis in KGC.}
This study evaluates seven aggregators on multiple KGC models covering six datasets (2.2M+ triples) and two tasks: tail $(h,r,?)$ and relation $(h,?,t)$ prediction.

\item \textbf{Five-dimensional evaluation framework for aggregator assessment:}
Each aggregator-provided ranking is tested across five key axes:
\begin{itemize}
    \item \emph{Consistency:} Assess alignment with individual base metrics.
    \item \emph{Stability:} Cross-dataset ranking variance across benchmarks.
    \item \emph{Metric independence:} Sensitivity to individual metric removal via leave-one-metric-out criteria. 
    \item \emph{Robustness:} Resilience to measurement uncertainty via noise injection (5\%-30\% over 1,000 independent trials for each noise level).
    \item \emph{Generalizability:} Predictive accuracy on unseen benchmarks via leave-one-dataset-out validation.
\end{itemize}

\item \textbf{Removal-based robust evaluation analysis:} Rather than scoring each aggregator on a single fixed set of KGC models, each one of the five tests is averaged over a model-removal criterion, LOMO, and LOGO, where entire architectural families (translational/geometric, semantic-matching, neural-CNN, and GNN-based) are removed. 

\item \textbf{Pareto-based aggregator selection and consensus ranking:} Aggregator selection is formulated as a multi-objective optimization problem to identify a balanced aggregator method that avoids privileging any single reliability criterion. The resulting Pareto-optimal aggregator is then employed to produce unified, consensus-driven rankings of KGC models across both prediction tasks.

\item \textbf{Test-sensitivity analysis:}
A leave-one-KGC-model-out study covering single-model removal and group-level (by architectural family) removal quantifies how much each test score shifts when a model/group is removed from MCDM rankings.
\end{enumerate}

\section{Related Work}
\label{sec:related}
Prior research has improved KGC evaluation from various directions, but most studies still analyze metrics individually rather than addressing how multiple conflicting metrics should be combined into a single reliable ranking~\cite{toutanova2015observed,kadlec2017knowledge,rossi2021knowledge}. In this section, we organize the literature into four strands directly relevant to the problem of this study: (1) Standardization and bias mitigation, (2) popularity-aware evaluation, (3) domain-specific and non-benchmark evaluation, and (4) meta-evaluation through MCDM. This structure helps to separate improvements to individual metrics from the broader question addressed in this work: how to derive a robust, consensus-driven ranking when base metrics disagree across metrics and datasets.

\textbf{Standardization and Bias Mitigation:}
Evaluation results can vary significantly due to small implementation choices. Sun et al.~\cite{sun-etal-2020-evaluation} show that ranking metrics are affected by scoring methods, ranking procedures, and bias handling. They also show that some models report higher results because of weak evaluation settings. To address this, they propose a standard evaluation method and find that model performance drops when evaluation is utilized properly. Similarly, Pezeshkpour et al.~\cite{pezeshkpour2020revisiting}, show that ranking-based evaluation does not fully measure triple classification under the Open World Assumption (OWA). They introduce YAGO3-TC for direct triple classification and find that simple type-based methods can perform as well as more complex embedding models. These studies reveal clear problems in current evaluation methods and improve fairness. However, they still focus on separate metrics. They do not combine different metric results into one ranking or explain how to compare models across datasets.

\textbf{Popularity-Aware Evaluation:}
A second critical stream addresses the popularity bias inherent in standard metrics. Mohamed et al.~\cite{pmlr-v124-mohamed20a} present that MRR and Hits@$k$ are systematically biased toward high-popularity entities and relations due to power-law distributions in knowledge graphs. They propose stratified metrics (strat-MRR, strat-Hits@$k$) that reweight predictions by entity frequency, revealing that embedding models perform significantly worse on long-tail entities than standard metrics suggest. Moon, Sooho, and Ko, Yunyongk~\cite{10.1145/3773966.3779401}, advance this direction with PROBE, a dual-perspective framework that evaluates models on predictive sharpness and robustness to popularity bias. PROBE introduces parametrized rank transformation and popularity-based aggregation, demonstrating that models like RNNLogic are overestimated by standard metrics while RotatE exhibits superior popularity-robustness. These approaches improve metric fairness but introduce additional complexity rather than simplification. Stratified metrics require tuning parameters to control emphasis on popularity, and PROBE demands domain knowledge to set sharpness ($\alpha$) and bias-robustness ($\beta$) values. More critically, they generate multiple metric streams (head vs. tail, sharp vs. lenient) that themselves require reconciliation, leaving the fundamental challenge of producing a single, interpretable consensus ranking unresolved.

\textbf{Domain-Specific and Non-Benchmark Evaluation:}
M. Habiburahman et al. \cite{habiburahman2024beyond} demonstrate that KGC models optimized on standard benchmarks (FB15k-237, WN18RR) show substantially different performance on real-world enterprise data with higher sparsity and skewed distributions. More recently, Peng et al.~\cite{peng2026causal} proposed CKGCNA, a causal knowledge graph completion model with multi-head attention for fault diagnosis in industrial settings. Their work highlights that standard embedding models frequently capture frequency biases rather than true causal relations. These domain-specific studies underscore a critical limitation: without a unified aggregation framework, practitioners must manually reconcile conflicting results across metrics, datasets, and application domains, hindering systematic progress.

\textbf{Meta-Evaluation via MCDM:}
Recently, Gul et al.~\cite{gul2025kg} introduced KG-EDAS, applying MCDM to KGC evaluation by aggregating MRR, Hits@$k$, and MR into a unified score. This marks a crucial paradigm shift from metric critique to integration. However, KG-EDAS leaves critical questions unanswered. It relies on a single aggregator without systematic comparison to established alternatives (TOPSIS, VIKOR, Borda Count, Z-score, etc.). Validation is limited to Pearson correlation, ignoring rank-aware statistics essential for ranking reliability. The study evaluates a single dataset split, neglecting cross-dataset stability, robustness to metric fluctuation, noise resilience, and generalisability. Most importantly, it provides no principled guidance on aggregator preference, nor does it separate aggregator validation from model ranking.

Recent advances in AI evaluation emphasize reliability, reproducibility, and decision transparency. While frameworks for dynamic benchmarking and aggregation-aware model selection have emerged, their adaptation to KGC remains limited. Studies on evaluation bias and domain-specific validation highlight the need for holistic, criteria-driven ranking over isolated metric reporting. This study bridges the gap by formalizing KGC evaluation as a multi-criteria decision problem, explicitly quantifying aggregator reliability across orthogonal dimensions, and providing a reproducible framework for consensus-driven model selection. This aligns with broader KBS objectives of transparent, evidence-based evaluation pipelines for intelligent systems.

Existing literature falls into three categories: (i) critiques of specific evaluation metrics without offering aggregation solutions~\cite{sun-etal-2020-evaluation,pezeshkpour2020revisiting,pmlr-v124-mohamed20a,10.1145/3773966.3779401,habiburahman2024beyond}; (ii) applications of a single aggregator without cross-method comparison or rigorous multi-dimensional validation~\cite{gul2025kg}; and (iii) meta-evaluation frameworks developed for other domains that have not been adapted to KGC's unique challenges.

This study addresses these gaps through a two-stage evaluation framework. First, seven MCDM aggregators are evaluated across different KGC models, standard benchmarks, and two prediction tasks using five reliability dimensions. Second, Pareto optimality analysis is used to select the best aggregators, which are then used to produce KGC model rankings. 
\section{Methodology}
\label{sec:methodology}
Current KGC model evaluation often relies on separate single-metric results, such as MRR, Hits@$k$, and MR. This leads to unstable model rankings across datasets and tasks. Since each metric measures a different part of ranking quality, using only one metric gives an incomplete view of model performance. To address this limitation, the KGC evaluation is reformulated as an MCDM problem, allowing multiple metrics to be combined into a single, clear, and structured evaluation framework.

\textbf{Problem Formulation:}
Let $\mathcal{M} = \{M_1, \dots, M_n\}$ denote a set of $n$ KGC models and $\mathcal{J} = \{c_1, \dots, c_m\}$ denote a set of $m$ evaluation metrics such as $\mathrm{MR},\mathrm{MRR},\mathrm{H@k}$ while ${\mathrm{k}=1,3,10}$. The performance of all models across metrics is represented by a matrix:
\begin{equation}
\mathbf{X} \in \mathbb{R}^{n \times m}, \quad X_{ij} = \text{performance of} ~ \mathcal{M}_i ~ \text{on} ~ \mathcal{J}_j.
\end{equation}
The objective is to derive a unified scoring function that aggregates multiple metric values into a single scalar score per model, thereby producing a consistent ranking across models. Formally, we define an aggregation function:
\begin{equation}
f: \mathbb{R}^{n \times m} \rightarrow \mathbb{R}^{n},
\end{equation}
which maps the performance matrix $\mathbf{X}$ to a score vector $\mathbf{s} = f(\mathbf{X})$, where each entry $s_i$ corresponds to the aggregated performance of model $\mathcal{M}_i$. Evaluation metrics are categorized into beneficial metrics (higher is better), such as MRR and Hits@$k$, and non-beneficial metrics (lower is better), such as MR. The standard definitions of these metrics are given by:
\begin{equation}
\label{eq:eva}
\text{MR} = \frac{1}{N} \sum_{i=1}^{N} \text{rank}_i
\end{equation}
\begin{equation}
\label{eq:eva}
\text{MRR} = \frac{1}{N} \sum_{i=1}^{N} \frac{1}{\text{rank}_i}, 
\end{equation}
\begin{equation}
\label{eq:evab}
 \quad   \text{Hits@k} = \frac{1}{N} \sum_{i=1}^{N} \mathbf{1}(\text{rank}_i \leq k)
\end{equation}
The main notations and symbols used throughout  this paper are summarized in Table~\ref{tab:notations}.

\begin{table}
\centering
\caption{Main notations used in this paper.}
\label{tab:notations}
\renewcommand{\arraystretch}{1.3}
\resizebox{\columnwidth}{!}{%
\begin{tabular}{ll ll}
\hline
\textbf{Symbol} & \textbf{Description} & \textbf{Symbol} & \textbf{Description} \\
\hline
\(\mathcal{M}\) & Set of \(n\) KGC models 
& \(n\) & Number of KGC models \\

\(\mathcal{J}\) & Set of \(m\) evaluation metrics 
& \(m\) & Number of metrics \\

\(\mathbf{X} \in \mathbb{R}^{n\times m}\) & Performance (decision) matrix 
& \(f(\cdot)\) & Aggregation function \\

\(\mathbf{s} = f(\mathbf{X})\) & Vector of aggregated scores 
& \(w_j = 1/m\) & Equal weight per metric \\

\(\mathcal{D}\) & Set of benchmark datasets 
& \(r_{a,i}^{(d)}\) & Rank of model \(i\) by aggregator \(a\) on dataset \(d\) \\

\(\mathbf{r}_a\) & Global ranking vector of aggregator \(a\) 
& \(C_a\) & Consistency score \\

\(S_a\) & Stability score 
& \(I(a)\) & Metric independence score \\

\(R(a)\) & Robustness score 
& \(G(a)\) & Generalizability score \\

\(E_i\) & EDAS score for model \(i\) 
& \(C_i\) & TOPSIS closeness score \\

\(Q_i\) & VIKOR compromise index 
& \(B_i\) & Borda Count score \\

\(\mathrm{ZS}_i\) & Z-score aggregated score 
& \(\mathbf{q}_a^{(t)}\) & Performance vector of aggregator \(a\) for task \(t\) \\

\(\mathcal{P}_t\) & Pareto frontier for task \(t\) & & \\
\hline
\end{tabular}%
}
\end{table}

To ensure fair comparison, the study adopts equal metric weights ($w_j = 1/m$). Equal weighting is used as a neutral choice because no universally accepted weighting scheme exists for KGC metrics. However, this does not mean that MR, MRR, and Hits@k are equally important in all applications. In practice, different downstream settings may place greater emphasis on early precision, average reciprocal rank, or robustness to outlier ranks. Therefore, sensitivity analysis using other weighting schemes is recommended for future benchmark reporting. Each MCDM aggregator computes composite scores from $\mathbf{X}$ and produces unified rankings for cross-model benchmarking. For each dataset and task, the original source of every MR, MRR, and Hits@k value is reported. 

\begin{figure*}[h!]
    \centering
    \includegraphics[width=0.99\textwidth]{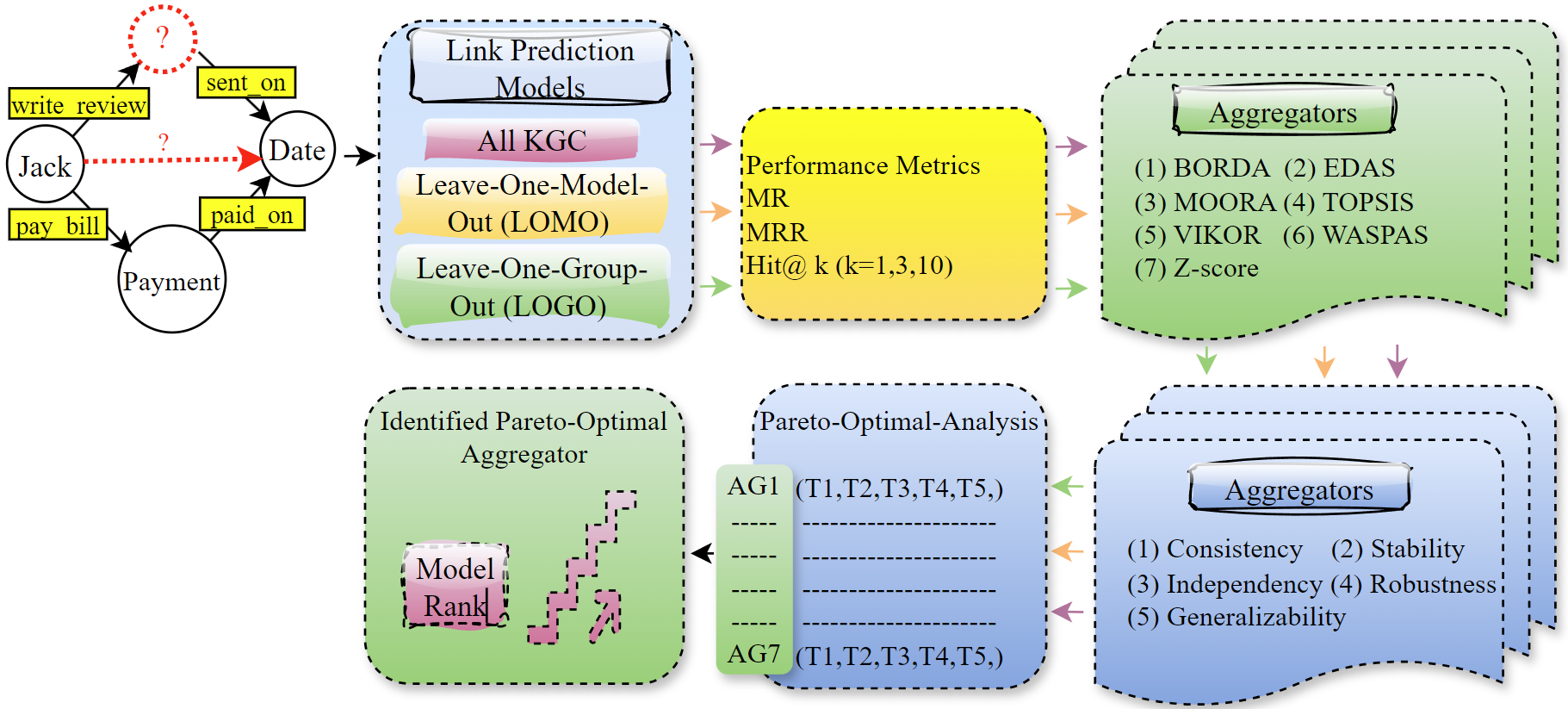}
\caption{Overview of the proposed Meta Analysis Model Benchmarking framework. 
\textit{(1)} Knowledge graph completion (KGC) models generate link prediction results for both tail $(h,r,?)$ and relation $(h,?,t)$ tasks, covering diverse model families such as translational, semantic, deep learning, and graph neural network (GNN)-based approaches. 
\textit{(2)} Multi-Criteria Decision-Making (MCDM) aggregator performance is evaluated under three patterns of KGC methods: Full-Set of KCG models, Leave-One-Model-Out (LOMO), and Leave-One-Group-Out (LOGO), using standard metrics (MR, MRR, Hit@$k$ for $k=1,3,10$).
\textit{(3)} To enhance the robustness, MCDM aggregators (BORDA, EDAS, MOORA, TOPSIS, VIKOR, WASPAS, Z-score)  are evaluated through each pattern of KGC methods using five tests: consistency, stability, independency (leave-one-metric-out), robustness (different level of noise injection), and generalisability (leave-one-dataset-out). 
\textit{(4)} A Pareto optimal analysis jointly assesses performance across all tests. 
\textit{(5)} Pareto optimal analysis identified the most balanced aggregator. 
\textit{(6)} The Pareto-optimal aggregator produces the final consensus ranking of KGC models.}
    \label{fig:methodology}
\end{figure*}

\subsection{Methodological Pipeline}
\label{subsec:pipeline}
Figure~\ref{fig:methodology} outlines the complete evaluation pipeline. First, knowledge graph completion models produce link prediction results for tail $(h,r,?)$ and relation $(h,?,t)$ tasks, evaluated using standard rank-based metrics (MR, MRR, Hits@$k$ ($k\in\{1,3,10\})$). 
Second, the performance of seven MCDM aggregators (BORDA, EDAS, MOORA, TOPSIS, VIKOR, WASPAS, Z-score) is assessed under three KGC method patterns: All-Base, Leave-One-Model-Out (LOMO), and Leave-One-Group-Out (LOGO). Third, each aggregator is evaluated across five tests: consistency, stability, independence (leave-one-metric-out), robustness (noise injection), and generalizability (leave-one-dataset-out). 
Fourth, a Pareto optimal analysis jointly compares aggregator performance across all five tests. 
Fifth, the most balanced (Pareto-optimal) aggregator is identified. 
Sixth, this optimal aggregator produces the final consensus ranking of KGC models.

\subsubsection{MCDM Aggregators}
\label{subsec:aggregators}
We now detail the mathematical formulations of the seven MCDM aggregators used to transform the decision matrix into unified model rankings.

\textbf{Aggregator 1 (EDAS):} Evaluation based on Distance from Average Solution (EDAS) computes a normalized score $E_i \in [0,1]$ by measuring each model’s positive and negative deviations from the average performance of each metric; higher values indicate better overall performance. The average performance for metric $j$ is given by: $\text{Avg}_j = \frac{1}{n} \sum_{i=1}^{n} X_{ij}, \quad j = 1, \ldots, m.$
Positive ($\text{PDA}_{ij}$) and negative ($\text{NDA}_{ij}$) deviations from $\text{Avg}_j$ are computed as:
For beneficial:
\begin{equation}
\text{PDA}_{ij} = \frac{\max\!\left(0, X_{ij} - \text{Avg}_j\right)}{\text{Avg}_j}
\end{equation}
\begin{equation}
\text{NDA}_{ij} = \frac{\max\!\left(0, \text{Avg}_j - X_{ij}\right)}{\text{Avg}_j}
\end{equation}
Non-beneficial
\begin{equation}
\text{PDA}_{ij} = \frac{\max\!\left(0, \text{Avg}_j - X_{ij}\right)}{\text{Avg}_j}
\end{equation}
\begin{equation}
\text{NDA}_{ij} = \frac{\max\!\left(0, X_{ij} - \text{Avg}_j\right)}{\text{Avg}_j}
\end{equation}
where $\mathrm{dir}(j)$ indicates preference direction (higher = beneficial, lower = non-beneficial) and equal weights $w_j = 1/m$ are used for fairness, the weighted deviations are:
\begin{equation}
\text{WPDA}_i=\sum_{j=1}^{m} w_j \cdot \text{PDA}_{ij},
\end{equation}
\begin{equation}
\text{WNDA}_i=\sum_{j=1}^{m} w_j \cdot \text{NDA}_{ij}
\end{equation}
After normalization to $[0,1]$:
\begin{equation}
NWPDA_i = \frac{\text{WPDA}_i}{\max_k \text{WPDA}_k}
\end{equation}
\begin{equation}
NWNDA_i = \frac{\text{WNDA}_i}{\max_k \text{WNDA}_k}
\end{equation}

The final score for model $i$ is:
\begin{equation}
 E_i = \frac{1}{2} \left[ N(\text{WPDA}_i) + (1 - N(\text{WNDA}_i)) \right].
\end{equation}
$E_i $ is the final EDAS value for each model. 

\textbf{Aggregator 2 (TOPSIS):} Technique for Order Preference by Similarity to Ideal Solution (TOPSIS) ranks models by closeness to the ideal solution. The decision matrix is vector-normalized to accommodate heterogeneous scales (e.g., MRR $\in [0,1]$ vs.\ MR $\in [1,|E|]$). Beneficial and non-beneficial criteria are distinguished in defining the positive ($A_j^+$) and negative ($A_j^-$) ideal solutions. With equal weights $w_j=1/m$, the weighted normalized values are
\begin{equation}
    v_{ij} = w_j \cdot \frac{x_{ij}}{\sqrt{\sum_{k=1}^n x_{kj}^2}}.
\end{equation}
\begin{align}
A_j^+ &=
\begin{cases}
\max_i v_{ij} & \text{beneficial} \\
\min_i v_{ij} & \text{non-beneficial}
\end{cases}
\end{align}

\begin{align}
A_j^- &=
\begin{cases}
\min_i v_{ij} & \text{beneficial} \\
\max_i v_{ij} & \text{non-beneficial}
\end{cases}
\end{align}
The Euclidean distances to PIS and NIS are:
\begin{equation}
D_i^+ = \sqrt{\sum_{j=1}^m (v_{ij} - A_j^+)^2}, 
\end{equation}
\begin{equation}
D_i^- = \sqrt{\sum_{j=1}^m (v_{ij} - A_j^-)^2}
\end{equation}

and the closeness score is $C_i = D_i^- / (D_i^+ + D_i^-)$. Models are ranked by descending $C_i$.

\textbf{Aggregator 3 (VIKOR):} VIsekriterijumsko KOmpromisno Rangiranje (VIKOR) ranks models by balancing group utility ($S_i$, average performance) and individual regret ($R_i$, worst-metric performance). It identifies the best ($f_j^*$) and worst ($f_j^-$) values per metric and computes:
\begin{equation}
S_i = \sum_{j=1}^{m} w_j \frac{f_j^* - X_{ij}}{f_j^* - f_j^-}, \quad R_i = \max_{j} \left( w_j \frac{f_j^* - X_{ij}}{f_j^* - f_j^-} \right),
\end{equation}
where $w_j = 1/m$ (equal weights). The compromise index is:
\begin{equation}
Q_i = v \frac{S_i - S^*}{S^- - S^*} + (1 - v) \frac{R_i - R^*}{R^- - R^*},
\end{equation}
with $S^* = \min_i S_i$, $S^- = \max_i S_i$, $R^* = \min_i R_i$, $R^- = \max_i R_i$, and $v = 0.5$ (standard setting). Models are ranked by ascending $Q_i$.
VIKOR is well-suited for KGC as it penalizes models with catastrophic failures in any metric (via $R_i$). 

\textbf{Aggregator 4 (BORDA):} Borda Count aggregates ordinal rankings across metrics into a consensus score, making it robust to scale incompatibilities (e.g., MRR vs.\ Hits@10). For each metric \(j\), models are ranked \(\rho_{ij} \in \{1,\dots,n\}\) (1 = best), and the Borda  Count is:
\begin{equation}
B_i = \sum_{j=1}^{m} (n + 1 - \rho_{ij}).
\end{equation}
with descending $B_i$ ranking.

\textbf{Aggregator 5 (WASPAS):}  WASPAS combines the Weighted Sum Model (WSM) and Weighted Product Model (WPM) for robust aggregation. After normalization, $\bar{x}_{ij}$ are calculated as:
\begin{equation}
\bar{x}_{ij} =
\begin{cases}
\frac{x_{ij}}{\max_i x_{ij}} & \text{for benefit criteria} \\
\frac{\min_i x_{ij}}{x_{ij}} & \text{for cost criteria}
\end{cases}
\end{equation}
The WSM score ($Q_i^{(1)}$) and WPM score ($Q_i^{(2)}$) are:
\begin{equation}
Q_i^{(1)} = \sum_{j=1}^{m} \bar{x}_{ij} w_j, \quad Q_i^{(2)} = \prod_{j=1}^{m} (\bar{x}_{ij})^{w_j}
\end{equation}
The final aggregated preference score is:
\begin{equation}
Q_i = \lambda Q_i^{(1)} + (1 - \lambda) Q_i^{(2)},
\end{equation}
where we use the standard balanced setting $\lambda = 0.5$.

\textbf{Aggregator 6 (MOORA):} Multi-Objective Optimization on the Basis of Ratio Analysis (MOORA)
uses vector normalization to enable cross-metric comparison:  
\begin{equation}
\bar{x}_{ij} = \frac{x_{ij}}{\sqrt{\sum_{k=1}^{n} x_{kj}^2}}
\end{equation}
\begin{equation}
y_i^* = \sum_{j=1}^{g} w_j \bar{x}_{ij} - \sum_{j=g+1}^{m} w_j \bar{x}_{ij}
\end{equation}
where $g = m-1$ (only MR is non-beneficial) and equal weights $w_j = 1/m$ are applied.  

\textbf{Aggregator 7 (ZS):} Z-score (ZS) standardizes metrics using mean \(\mu_j\) and standard deviation \(\sigma_j\):  
\begin{equation}
z_{ij} = \frac{x_{ij} - \mu_j}{\sigma_j}, \quad
\text{ZS}_i = \frac{1}{m} \sum_{j=1}^{m} z_{ij}.
\end{equation}  
Models are ranked by descending \(\text{ZS}_i\).

\textbf{Computational Complexity:} Table~\ref{MCDM:complexity} compares time and space complexity of the MCDM methods. EDAS, VIKOR, Z-score, WASPAS, and MOORA are linear $\mathcal{O}(nm)$ in both time and space (highly efficient for KGC). TOPSIS and Borda Count incur slightly higher costs due to distances/sorting, while Pareto Frontier (not shown) scales poorly with $n$.
\begin{table}
\centering
\caption{Aggregators' Complexity}
\label{MCDM:complexity}
\small
\renewcommand{\arraystretch}{1.3}
\resizebox{\columnwidth}{!}{%
\begin{tabular}{l c c >{\raggedright\arraybackslash}p{4.8cm}}
\hline
\textbf{Method}      & \textbf{Time}         & \textbf{Space}      & \textbf{Notes} \\
\hline
EDAS         & $\mathcal{O}(nm)$     & $\mathcal{O}(nm)$   & Linear in models $\times$ metrics \\
TOPSIS       & $\mathcal{O}(nm+n^2)$ & $\mathcal{O}(nm+n^2)$ & Distance to ideal solutions \\
VIKOR        & $\mathcal{O}(nm)$     & $\mathcal{O}(nm)$   & Max and weighted sum operations \\
Z-score      & $\mathcal{O}(nm)$     & $\mathcal{O}(nm)$   & Mean and standard deviation \\
Borda Count  & $\mathcal{O}(nm\log m)$ & $\mathcal{O}(nm)$ & Per-metric sorting; possible ties \\
WASPAS       & $\mathcal{O}(nm)$     & $\mathcal{O}(nm)$   & Weighted sum + weighted product \\
MOORA        & $\mathcal{O}(nm)$     & $\mathcal{O}(nm)$   & Vector normalization + sum difference\\
\hline
\end{tabular}%
}
\end{table}

\subsubsection{Five-Dimensional Meta-Evaluation of Aggregators}
\label{subsec:meta_evaluation}
The key part of this study is a meta-evaluation framework that assesses each aggregator using five complementary tests and examines whether it preserves the original metric information, produces stable rankings across datasets, remains independent of individual metrics, resists noisy metric values, and transfers well to unseen datasets. Together, these criteria quantify whether an aggregator is reliable enough to support unified KGC benchmarking. The following section provides brief details of each test. 

\textbf{Consistency}:
Consistency measures whether the ranking produced by each MCDM aggregator agrees with the rankings induced by the original evaluation metrics. After each aggregator computes its score, the models are ranked in descending order of performance, where rank 1 indicates the best model. The same ranking rule is applied to the base metrics: MR is treated as a non-beneficial metric, so a lower MR receives a better rank, while MRR and Hits@$k$ are treated as beneficial metrics, so higher values receive better ranks. Therefore, consistency is computed using rank vectors rather than raw score vectors. The ranks used for consistency were computed across all datasets. 

Let $\mathbf{r}_{a}$ denote the rank vector produced by aggregator $a$ across all datasets, and let $\mathbf{r}_{j}$ denote the rank vector induced by base metric $j$ across all datasets, where $j \in \mathcal{J}$ and $\mathcal{J}=\{\mathrm{MR},\mathrm{MRR},\mathrm{H@1},\mathrm{H@3},\mathrm{H@10}\}$. The consistency score of aggregator $a$, denoted by $C_a$, is computed by averaging Kendall, Pearson, and Spearman correlations across all base metrics:
\begin{equation}
C_a
=
\frac{1}{3 . |\mathcal{J}|}
\sum_{j \in \mathcal{J}}
\left|\tau\left(\mathbf{r}_{a},\mathbf{r}_{j}\right)\right|
+
\left|r\left(\mathbf{r}_{a},\mathbf{r}_{j}\right)\right|
+
\left|\rho\left(\mathbf{r}_{a},\mathbf{r}_{j}\right)\right|
\end{equation}
Here, $\tau$, $r$, and $\rho$ denote Kendall, Pearson, and Spearman correlation coefficients, respectively. The absolute values are used so that the strength of agreement is captured regardless of sign.
For comparison across aggregators, the final score of the aggregator $a$ (for any test) is normalized by dividing it by the maximum score among all aggregators:
\begin{equation}
\label{eq:nor}
T_a^{(k)} = \frac{\mathcal{T}_a^{(k)}}{\max_{b \in \mathcal{A}} \mathcal{T}_b^{(k)}},
\end{equation}
where $T_a^{(k)} \in [0,1]$ is the normalized score (higher values indicate better performance).

\textbf{Stability}:
Stability measures whether an aggregator produces similar model rankings across different datasets. A stable aggregator should not cause the same model to move sharply up or down in ranking when the benchmark dataset changes.  Unlike the consistency test, which uses global ranks computed across all datasets, stability is evaluated using dataset-wise ranks (i.e., models are ranked separately on each dataset according to the aggregator's score on that dataset).

Let $r_{a,i}^{(d)}$ denote the rank of model $i$ produced by aggregator $a$ on dataset $d$. The average rank of model $i$ across all datasets is computed as:
\begin{equation}
\bar{r}_{a,i}
=
\frac{1}{|\mathcal{D}|}
\sum_{d \in \mathcal{D}}
r_{a,i}^{(d)}.
\end{equation}
Here, $\mathcal{D}$ is the set of datasets and $|\mathcal{D}|$ is the number of datasets. 
The rank variance of model $i$ under aggregator $a$ is then:
\begin{equation}
\mathrm{Var}_{i}(a)
=
\frac{1}{|\mathcal{D}|}
\sum_{d \in \mathcal{D}}
\left(
r_{a,i}^{(d)}
-
\bar{r}_{a,i}
\right)^2.
\end{equation}

A smaller variance indicates that the model maintains a more consistent rank position across datasets. Since lower variance is better, the final stability score for each aggregator is obtained by averaging the inverted variances across all models, $S_a$ followed by normalization using Equation \ref{eq:nor} (higher is better).
\begin{equation}
S_a = \frac{1}{N} \sum_{i=1}^{N} \frac{1}{\mathrm{Var}_{i}(a)}.
\end{equation}

\textbf{Independency (Leave-One-Metric-Out)}:
Metric independence evaluates whether an aggregator depends too strongly on any single metric family. A reliable aggregator should produce a similar ranking even when one metric family is removed from the decision matrix. Therefore, a leave-one-metric-family-out analysis is performed.  First, each aggregator is computed using all available metrics to obtain the baseline ranking. Then, one metric family is removed, the aggregator is recomputed, and the new ranking is compared with the baseline ranking.  The baseline rank vector is computed across all datasets (global ranking).

Let $r_{a}$ denote the baseline rank vector produced by aggregator $a$ across all datasets, and let $r_{a,-j}$ denote the rank vector after removing metric family $j$, where $j \in \mathcal{J}$ and $\mathcal{J}$ is the set of metric families. The independence score is computed by first averaging the absolute values of Kendall's $\tau$, Pearson's $r$, and Spearman's $\rho$ for each removed metric family, and then averaging across all metric families:
\begin{equation}
\begin{aligned}
I(a)
&=
\frac{1}{3 \cdot |\mathcal{J}|}
\sum_{j \in \mathcal{C}} \Bigl(
\left|\tau\left(r_{a}, r_{a,-j}\right)\right|
+
\left|r\left(r_{a}, r_{a,-j}\right)\right| \\
&\quad
+
\left|\rho\left(r_{a}, r_{a,-j}\right)\right|
\Bigr).
\end{aligned}
\end{equation}
For comparison across aggregators, this score is normalized using Equation \ref{eq:nor}.
In addition to rank correlation, the positional change in model rankings is measured using average absolute rank displacement. For aggregator $a$ and removed metric family $j$, the rank displacement is:
\begin{equation}
\Delta_{a,j}
=
\frac{1}{|\mathcal{N}|}
\sum_{i=1}^{|\mathcal{N}|}
\left|
r_{a,i}
-
r_{a,-j,i}
\right|,
\end{equation}
where $r_{a,i}$ is the baseline rank of model $i$ and $r_{a,-j,i}$ is its rank after removing metric family $j$. A smaller value indicates less sensitivity to metric removal.
The average rank displacement across all metric families is:
\begin{equation}
\overline{\Delta}(a)
=
\frac{1}{|\mathcal{C}|}
\sum_{j \in \mathcal{C}}
\Delta_{a,j}.
\end{equation}
Since lower displacement is better, this value is inverted and normalized so that higher values indicate stronger metric independence. The final independence score combines both the correlation-based and displacement-based components (both higher-is-better after normalization).

\textbf{Robustness (Under Noise Injection)}:
Robustness measures whether an aggregator maintains its ranking when the metric values change. In this test, random noise is added to all metric values at four levels: 5\%, 10\%, 20\%, and 30\%. For each noise level, the MCDM aggregators are recomputed over 1,000 Monte Carlo iterations, and the noisy rankings are compared with the original baseline rankings. The baseline and noisy rankings used for comparison are global rankings computed across all datasets.

Let $r_{a}$ denote the baseline (noise-free) rank vector produced by aggregator $a$ across all datasets, and let $\tilde{r}_{a,\eta,q}$ denote the rank vector produced after adding noise level $\eta$ in Monte Carlo iteration $q$. The robustness score is computed by first averaging the absolute values of Kendall's $\tau$, Pearson's $r$, and Spearman's $\rho$ correlations for each noise level and iteration, and then averaging across all noise levels and Monte Carlo iterations:
\begin{equation}
\begin{aligned}
R(a)
&=
\frac{1}{3 \cdot |\mathcal{E}| R}
\sum_{\eta \in \mathcal{E}}
\sum_{q=1}^{R}
\Bigl(
\left|\tau\left(r_{a}, \tilde{r}_{a,\eta,q}\right)\right|
+
\left|r\left(r_{a}, \tilde{r}_{a,\eta,q}\right)\right| \\
&\quad
+
\left|\rho\left(r_{a}, \tilde{r}_{a,\eta,q}\right)\right|
\Bigr)
\end{aligned}
\end{equation}
Here, $\mathcal{E}=\{0.05,0.10,0.20,0.30\}$ is the set of noise levels, and $R=1000$ is the number of Monte Carlo iterations. A higher value $R(a)$ indicates that the noisy rankings remain more strongly aligned with the baseline ranking. For comparison across aggregators, this correlation-based score is normalized using Equation \ref{eq:nor}. In addition to correlation, robustness is also evaluated using average absolute rank displacement:
\begin{equation}
\Delta_{a,\eta,q}
=
\frac{1}{|\mathcal{N}|}
\sum_{i=1}^{|\mathcal{N}|}
\left|
r_{a,i}
-
\tilde{r}_{a,i,\eta,q}
\right|.
\end{equation}
The mean rank displacement of aggregator $a$ across all noise levels and Monte Carlo iterations is:
\begin{equation}
\overline{\Delta}(a)
=
\frac{1}{|\mathcal{E}| R}
\sum_{\eta \in \mathcal{E}}
\sum_{q=1}^{R}
\Delta_{a,\eta,q}.
\end{equation}
Since lower displacement is better, this value is inverted and normalized so that higher values indicate stronger robustness. The final robustness score combines both the correlation-based and displacement-based components (both higher-is-better after normalization).

\textbf{Generalizability (Leave-One-Dataset-Out):}
Generalizability evaluates whether an aggregator can identify strong models on an unseen dataset. One dataset is left out, and the rankings from the remaining datasets are used to predict the model ranking for the eliminated dataset. The predicted ranking for the eliminated dataset is compared with reference rankings derived from the base metrics on that same dataset.

Let $r_{i,-d}^{(a)}$ denote the \textit{predicted} rank of the model $i$ produced by the aggregator $a$ for the eliminated dataset $d$, derived from its aggregated performance across the remaining datasets.

The predicted ranking is then compared with the reference rankings induced by each base metric on the held-out dataset (using the same beneficial/non-beneficial ranking direction as in the consistency test). The generalizability score is computed by averaging the absolute values of Kendall's $\tau$, Pearson's $r$, and Spearman's $\rho$ across all base metrics and all leave-one-dataset-out folds:

\begin{equation}
\begin{aligned}
G(a)
&=
\frac{1}{3 \cdot |\mathcal{J}| \cdot |\mathcal{D}|}
\sum_{d \in \mathcal{D}}
\sum_{j \in \mathcal{J}}
\Bigl(
\left|\tau\left(r_{a,-d}, r_{j,d}\right)\right|
\\
&\quad
+
\left|r\left(r_{a,-d}, r_{j,d}\right)\right|
\\
&\quad
+
\left|\rho\left(r_{a,-d}, r_{j,d}\right)\right|
\Bigr).
\end{aligned}
\end{equation}

The final score of aggregators is normalised using Equation \ref{eq:nor}.
In addition to correlation, top-1 and top-3 accuracy are recorded (i.e., whether the best model or the top-3 models on the held-out dataset are correctly predicted), along with the mean absolute rank displacement between the predicted and reference rankings. Since lower-rank displacements are better, they are inverted and normalized so that higher values indicate stronger generalizability. The final generalizability score combines the correlation-based measure with these additional indicators (all higher-is-better after normalization).

\subsubsection{KGC Model-Removal Analysis (LOMO + LOGO)}
\label{lab:kgc-removal}
To enhance robustness, each test is averaged across a model-removal analysis (one model and group KGC model removal). Let $\mathcal{M} = \{M_1,\dots,M_n\}$ be the full set of $(n)$ KGC models. In leave-one-KGC-model-out (LOMO), each model $(M_u)$ is removed in turn; the aggregator is re-scored on the remaining $(n-1)$ models and computes ranks. In leave-one-group-out (LOGO), the models are partitioned into different groups (see Table \ref{tab:tail_link_prediction_all} for group detail): translational/geometric, semantic-matching, neural-CNN, and GNN-based, and each group is removed as a block; the remaining are executed, the score is computed, and the ranks are recorded. 
The final score of a test for an aggregator is the mean of its comparison across all runs, and the standard deviation across runs is reported as the error bar. Models removed in a given run are excluded from that run's comparison. 

For broader validation, we evaluate four configurations: (1) the integrated Full-Set+LOMO+LOGO, the main focus of this paper; while (2) LOMO only; (3) LOGO only; and (4) Full-Set without removal results are reported in the supplementary file. Consequently, all five meta-evaluation tests reported in the main text are averaged across the Full-Set+LOMO+LOGO runs, ensuring that reliability estimates remain robust to the specific composition of the evaluated model set. The analysis of these tests begins with consistency.
\subsubsection{Pareto-Based Aggregator Selection:}
The five reliability scores used in this stage have already been averaged across all LOMO and LOGO removal iterations, ensuring that each score reflects consistent performance across diverse model subsets rather than a single fixed collection. Because no single evaluation criterion can fully capture an aggregator’s overall quality, we treat aggregator selection as a multi-objective optimization problem. For each prediction task $t$ (tail and relation prediction), every aggregator $a$ is represented by a normalized five-dimensional performance vector:

\begin{equation}
\mathbf{q}_{a}^{(t)} =
\bigl[
C_{(a)},\,
S_{(a)},\,
I_{(a)},\,
R_{(a)},\,
G_{(a)}
\bigr],
\end{equation}
where each component is scaled to the range $[0,1]$ across all aggregators for the same task (higher values indicate better performance).
An aggregator $a$ dominates another aggregator $b$ if it performs at least as well as $b$ on all five criteria and strictly better on at least one criterion. The set of non-dominated aggregators forms the Pareto frontier $\mathcal{P}_t$.

Since multiple aggregators may lie on the Pareto frontier, the final aggregator is selected as the one closest to the ideal point using Euclidean distance:
\begin{equation}
a_t^{*}
=
\arg\min_{a \in \mathcal{P}_t}
\left\|
\mathbf{q}_{a}^{(t)} - \mathbf{1}
\right\|_2.
\end{equation}
This approach selects the most balanced aggregator that does not excessively favour any single reliability dimension.

\textbf{Final Ranking of KGC Models:}
The selected optimal aggregator $a_t^{*}$ is then used to compute the final consensus ranking of KGC models for each prediction task (tail prediction and relation prediction).

\subsubsection{Tests Sensitivity Analysis}
Following section \ref{lab:kgc-removal}, we know that the reliability of an aggregator may depend on the specific set of KGC models included in the evaluation. To assess this, sensitivity analyses, LOMO, and LOGO are performed. For each individual model $u$, the five-dimensional performance vector of every aggregator is recomputed after removing $u$ from all decision matrices. Let 
\[
\mathbf{g}(a) =
\bigl[
C_{(a)},\,
S_{(a)},\,
I_{(a)},\,
R_{(a)},\,
G_{(a)}
\bigr]
\]
denote the original normalized score vector of aggregator $a$, and let $\mathbf{g}^{(-u)}(a)$ denote the score vector after removing model $u$. The model-level sensitivity is defined as:
\begin{equation}
\Delta_{\mathrm{model}}(u,a)
=
\frac{1}{5}
\left\|
\mathbf{g}(a) - \mathbf{g}^{(-u)}(a)
\right\|_1.
\end{equation}
Additionally, models are grouped into architectural families (e.g., translational, semantic, neural, rule-based, and GNN-based). For a family $F$, all models belonging to $F$ are removed simultaneously and the scores are recomputed. The family-level sensitivity is defined as:
\begin{equation}
\Delta_{\mathrm{family}}(F,a)
=
\frac{1}{5}
\left\|
\mathbf{g}(a) - \mathbf{g}^{(-F)}(a)
\right\|_1,
\end{equation}
where $\mathbf{g}^{(-F)}(a)$ is the score vector after removing all models in family $F$.
These analyses quantify how much the aggregator evaluation scores change when parts of the model set are excluded. Smaller sensitivity values indicate that the aggregator assessments are more stable and less prone to variation across the evaluated KGC models.

\begin{table*}
\centering
\caption{Link prediction results (head/tail entity prediction) on FB15k, WN18, FB15k-237, WN18RR, and YAGO3-10. 
Results reported in this table are published in \cite{10.1016/j.eswa.2024.125260,Cao_Xu_Yang_Cao_Huang_2021,electronics11233866,10.1145/3424672,zhang2020hake}.}
\label{tab:tail_link_prediction_all}
\resizebox{\textwidth}{!}{%
\renewcommand{\arraystretch}{1.25}
\LARGE
\begin{tabular}{lcccccccccccccccccccc}
\toprule
\multirow{2}{*}{\textbf{Models}} 
& \multicolumn{4}{c}{\textbf{FB15k}} 
& \multicolumn{4}{c}{\textbf{WN18}} 
& \multicolumn{4}{c}{\textbf{FB15k-237}} 
& \multicolumn{4}{c}{\textbf{WN18RR}} 
& \multicolumn{4}{c}{\textbf{YAGO3-10}} \\
\cmidrule(lr){2-5} \cmidrule(lr){6-9} \cmidrule(lr){10-13} \cmidrule(lr){14-17} \cmidrule(lr){18-21}
& \textbf{MR} $\downarrow$ & \textbf{MRR} $\uparrow$ & \textbf{H@1} $\uparrow$ & \textbf{H@10} $\uparrow$
& \textbf{MR} $\downarrow$ & \textbf{MRR} $\uparrow$ & \textbf{H@1} $\uparrow$ & \textbf{H@10} $\uparrow$
& \textbf{MR} $\downarrow$ & \textbf{MRR} $\uparrow$ & \textbf{H@1} $\uparrow$ & \textbf{H@10} $\uparrow$
& \textbf{MR} $\downarrow$ & \textbf{MRR} $\uparrow$ & \textbf{H@1} $\uparrow$ & \textbf{H@10} $\uparrow$
& \textbf{MR} $\downarrow$ & \textbf{MRR} $\uparrow$ & \textbf{H@1} $\uparrow$ & \textbf{H@10} $\uparrow$ \\
\midrule
\multicolumn{21}{l}{\textbf{Translational / Geometric Models}} \\
TransE  & 45 & 0.628 & 0.494 & 0.847 & 279 & 0.646 & 0.406 & 0.949 & 209 & 0.310 & 0.217 & 0.497 & 3936 & 0.206 & 0.028 & 0.495 & 1187 & 0.501 & 0.406 & 0.674 \\
STransE & 69 & 0.543 & 0.398 & 0.796 & 208 & 0.656 & 0.431 & 0.935 & 357 & 0.315 & 0.225 & 0.496 & 5172 & 0.226 & 0.101 & 0.422 & 5797 & 0.049 & 0.033 & 0.074 \\
CrossE  & 136 & 0.702 & 0.601 & 0.862 & 441 & 0.834 & 0.733 & 0.950 & 227 & 0.298 & 0.212 & 0.471 & 5212 & 0.405 & 0.381 & 0.450 & 3839 & 0.446 & 0.331 & 0.654 \\
TorusE  & 143 & 0.746 & 0.689 & 0.840 & 525 & 0.947 & 0.943 & 0.954 & 211 & 0.281 & 0.196 & 0.447 & 4873 & 0.463 & 0.427 & 0.534 & 1945 & 0.342 & 0.274 & 0.474 \\
RotatE  & 42 & 0.791 & 0.739 & 0.881 & 274 & 0.949 & 0.943 & 0.960 & 178 & 0.336 & 0.238 & 0.531 & 3318 & 0.475 & 0.426 & 0.573 & 1827 & 0.498 & 0.405 & 0.671 \\
HakE    & 128 & 0.714 & 0.639 & 0.834 & 304 & 0.934 & 0.919 & 0.957 & 160 & 0.346 & 0.250 & 0.542 & 2250 & 0.497 & 0.452 & 0.582 & 1130 & 0.545 & 0.462 & 0.694 \\
DualE   & 21 & 0.813 & 0.766 & 0.896 & 156 & 0.952 & 0.946 & 0.962 & 91 & 0.365 & 0.268 & 0.559 & 2270 & 0.492 & 0.444 & 0.584 & 2068 & 0.280 & 0.096 & 0.591 \\
\midrule
\multicolumn{21}{l}{\textbf{Semantic Matching Models}} \\
DistMult & 173 & 0.784 & 0.736 & 0.863 & 675 & 0.824 & 0.726 & 0.946 & 199 & 0.313 & 0.224 & 0.490 & 5913 & 0.433 & 0.397 & 0.502 & 1107 & 0.501 & 0.413 & 0.661 \\
ComplEx  & 34 & 0.848 & 0.816 & 0.905 & 3623 & 0.949 & 0.945 & 0.955 & 202 & 0.349 & 0.257 & 0.530 & 4907 & 0.458 & 0.426 & 0.521 & 1112 & 0.576 & 0.505 & 0.704 \\
ANALOGY  & 126 & 0.726 & 0.656 & 0.837 & 808 & 0.934 & 0.926 & 0.944 & 476 & 0.202 & 0.126 & 0.354 & 9266 & 0.366 & 0.358 & 0.380 & 2423 & 0.283 & 0.192 & 0.457 \\
SimplE   & 138 & 0.726 & 0.661 & 0.836 & 759 & 0.938 & 0.933 & 0.946 & 651 & 0.179 & 0.100 & 0.344 & 8764 & 0.398 & 0.383 & 0.427 & 2849 & 0.453 & 0.358 & 0.632 \\
HolE     & 211 & 0.800 & 0.759 & 0.868 & 650 & 0.938 & 0.931 & 0.949 & 186 & 0.303 & 0.214 & 0.476 & 8401 & 0.432 & 0.403 & 0.488 & 6489 & 0.502 & 0.418 & 0.652 \\
TuckER   & 39 & 0.788 & 0.729 & 0.889 & 510 & 0.951 & 0.946 & 0.958 & 162 & 0.352 & 0.259 & 0.536 & 6239 & 0.459 & 0.430 & 0.514 & 2417 & 0.544 & 0.466 & 0.681 \\
\midrule
\multicolumn{21}{l}{\textbf{Neural / CNN-based Models}} \\
ConvE  & 51 & 0.688 & 0.595 & 0.849 & 413 & 0.945 & 0.939 & 0.957 & 281 & 0.305 & 0.219 & 0.476 & 4944 & 0.427 & 0.390 & 0.508 & 2429 & 0.488 & 0.399 & 0.658 \\
ConvKB & 109 & 0.567 & 0.462 & 0.762 & 516 & 0.824 & 0.743 & 0.945 & 257 & 0.396 & 0.300 & 0.517 & 2554 & 0.248 & 0.150 & 0.525 & 2626 & 0.419 & 0.323 & 0.586 \\
ConvR  & 70 & 0.773 & 0.723 & 0.870 & 471 & 0.950 & 0.946 & 0.959 & 251 & 0.346 & 0.256 & 0.526 & 5646 & 0.467 & 0.437 & 0.527 & 2582 & 0.527 & 0.446 & 0.673 \\
RSN    & 51 & 0.777 & 0.706 & 0.886 & 346 & 0.928 & 0.912 & 0.951 & 248 & 0.280 & 0.198 & 0.444 & 4210 & 0.395 & 0.346 & 0.483 & 1339 & 0.511 & 0.427 & 0.664 \\
\midrule
\multicolumn{21}{l}{\textbf{GNN-based Models}} \\
R-GCN      & 68 & 0.651 & 0.541 & 0.825 & 510 & 0.819 & 0.697 & 0.964 & 600 & 0.164 & 0.100 & 0.300 & 6700 & 0.123 & 0.080 & 0.200 & 15000 & 0.120 & 0.060 & 0.211 \\
CompGCN & 69 & 0.419 & 0.270 & 0.498 & 539 & 0.676 & 0.518 & 0.921 & 197 & 0.355 & 0.264 & 0.535 & 3533 & 0.479 & 0.443 & 0.546 & 1591 & 0.227 & 0.126 & 0.442 \\
NodePiece & 420 & 0.148 & 0.078 & 0.158 & 633 & 0.435 & 0.263 & 0.647 & 230 & 0.256 & 0.190 & 0.420 & 5000 & 0.403 & 0.400 & 0.515 & 3800 & 0.247 & 0.200 & 0.488 \\
\bottomrule
\end{tabular}
}
\end{table*}

\begin{table*}
\centering
\caption{\small Aggregator scores for knowledge graph embedding models (ZS: Z-score, ED: EDAS, MO: MOORA, BS: Borda Score, TP: TOPSIS, WA: WASPAS, VQ: VIKOR).}
\label{tab:tail_mcdm-kgc-scores}
\setlength{\tabcolsep}{12pt}
\renewcommand{\arraystretch}{1.0}
\resizebox{\textwidth}{!}{%
\begin{tabular}{lccccccccccc}
\toprule
\textbf{Model} & \textbf{ED} & \textbf{TP} & \textbf{VQ} & \textbf{BS} & \textbf{ZS} & \textbf{MO} & \textbf{WA} & \textbf{MR} & \textbf{MRR} & \textbf{H1} & \textbf{H10} \\
\midrule
DualE     & 0.9397 & 0.7670 & 0.1633 & 1.0000 & 1.0000 & 0.2861 & 0.8403 & 0.9053 & 0.5804 & 0.5040 & 0.7184 \\
HakE      & 0.8548 & 0.6197 & 0.1446 & 0.8920 & 0.8968 & 0.2609 & 0.8083 & 0.6451 & 0.6072 & 0.5444 & 0.7218 \\
ComplEx   & 0.8269 & 0.5981 & 0.1031 & 0.9379 & 0.8641 & 0.2543 & 0.5132 & 0.5130 & 0.6360 & 0.5898 & 0.7230 \\
RotatE    & 0.7997 & 0.6503 & 1.0000 & 0.9246 & 0.8608 & 0.2551 & 0.7947 & 0.5729 & 0.6098 & 0.5502 & 0.7232 \\
TuckER    & 0.7459 & 0.5970 & 0.1412 & 0.9275 & 0.8149 & 0.2449 & 0.7476 & 0.4449 & 0.6188 & 0.5660 & 0.7156 \\
ConvR     & 0.6842 & 0.5415 & 0.1534 & 0.8328 & 0.7593 & 0.2323 & 0.7122 & 0.3642 & 0.6126 & 0.5616 & 0.7110 \\
RSN       & 0.6673 & 0.5792 & 0.1707 & 0.6494 & 0.7007 & 0.2337 & 0.6947 & 0.5181 & 0.5782 & 0.5178 & 0.6856 \\
DistMult  & 0.6309 & 0.5211 & 0.1004 & 0.6317 & 0.6758 & 0.2243 & 0.6529 & 0.4381 & 0.5710 & 0.4992 & 0.6924 \\
ConvE     & 0.6007 & 0.5374 & 0.1386 & 0.6257 & 0.6663 & 0.2230 & 0.6682 & 0.4048 & 0.5706 & 0.5084 & 0.6896 \\
HolE      & 0.5608 & 0.4808 & 0.0885 & 0.6080 & 0.6323 & 0.2108 & 0.5861 & 0.2534 & 0.5950 & 0.5450 & 0.6866 \\
TorusE    & 0.5349 & 0.4860 & 0.0983 & 0.6109 & 0.5996 & 0.2094 & 0.6150 & 0.3812 & 0.5558 & 0.5058 & 0.6498 \\
TransE    & 0.4915 & 0.5446 & 0.1165 & 0.5799 & 0.5438 & 0.2132 & 0.5863 & 0.5931 & 0.4582 & 0.3102 & 0.6924 \\
CrossE    & 0.4774 & 0.4729 & 0.0954 & 0.4882 & 0.5512 & 0.2032 & 0.5875 & 0.3258 & 0.5370 & 0.4516 & 0.6774 \\
ConvKB    & 0.4042 & 0.4394 & 0.0780 & 0.5163 & 0.5377 & 0.1954 & 0.3840 & 0.4303 & 0.4908 & 0.3356 & 0.6670 \\
SimplE    & 0.3933 & 0.4272 & 0.0714 & 0.3876 & 0.4034 & 0.1834 & 0.2853 & 0.2286 & 0.5388 & 0.4870 & 0.6370 \\
CompGCN   & 0.3867 & 0.4676 & 0.1049 & 0.5814 & 0.4569 & 0.1943 & 0.5386 & 0.4777 & 0.4312 & 0.3242 & 0.5884 \\
ANALOGY   & 0.3134 & 0.3957 & 0.0681 & 0.3402 & 0.3380 & 0.1721 & 0.3023 & 0.2501 & 0.5022 & 0.4516 & 0.5944 \\
STransE   & 0.1313 & 0.3802 & 0.0624 & 0.3358 & 0.2214 & 0.1557 & 0.2102 & 0.3870 & 0.3578 & 0.2376 & 0.5446 \\
NodePiece & 0.0321 & 0.3389 & 0.0567 & 0.2840 & 0.0010 & 0.1404 & 0.1550 & 0.2867 & 0.2978 & 0.2262 & 0.4456 \\
R-GCN     & 0.0071 & 0.2938 & 0.0564 & 0.2722 & 0.0258 & 0.1289 & 0.1524 & 0.2352 & 0.3754 & 0.2956 & 0.5000 \\
\bottomrule
\end{tabular}%
}
\end{table*}

\begin{table}
\centering
\caption{\small Ranking of knowledge graph completion models based on MCDM aggregator methods (1 = best).}
\label{tab:tail_mcdm-kgc-ranks}
\setlength{\tabcolsep}{3pt} 
\tiny
\renewcommand{\arraystretch}{0.9} 
\footnotesize
\resizebox{\columnwidth}{!}{%
\begin{tabular}{lccccccccccc}
\toprule
\textbf{Model} & \textbf{ED} & \textbf{TP} & \textbf{VQ} & \textbf{BS} & \textbf{ZS} & \textbf{MO} & \textbf{WA} & \textbf{MR} & \textbf{MRR} & \textbf{H1} & \textbf{H10} \\
\midrule
DualE     & 1 & 1 & 3 & 1 & 1 & 1 & 1 & 1 & 7  & 10 & 4  \\
HakE      & 2 & 3 & 5 & 5 & 2 & 2 & 2 & 2 & 5  & 6  & 3  \\
ComplEx   & 3 & 4 & 10& 2 & 3 & 4 & 14& 6 & 1  & 1  & 2  \\
RotatE    & 4 & 2 & 1 & 4 & 4 & 3 & 3 & 4 & 4  & 4  & 1  \\
TuckER    & 5 & 5 & 6 & 3 & 5 & 5 & 4 & 8 & 2  & 2  & 5  \\
ConvR     & 6 & 8 & 4 & 6 & 6 & 7 & 5 & 14& 3  & 3  & 6  \\
RSN       & 7 & 6 & 2 & 7 & 7 & 6 & 6 & 5 & 8  & 7  & 11 \\
DistMult  & 8 & 10& 11& 8 & 8 & 8 & 8 & 9 & 9  & 11 & 7  \\
ConvE     & 9 & 9 & 7 & 9 & 9 & 9 & 7 & 11& 10 & 8  & 9  \\
HolE      & 10& 12& 14& 11& 10& 11& 12& 17& 6  & 5  & 10 \\
TorusE    & 11& 11& 12& 10& 11& 12& 9 & 13& 11 & 9  & 14 \\
TransE    & 12& 7 & 8 & 13& 13& 10& 11& 3 & 16 & 17 & 7  \\
CrossE    & 13& 13& 13& 15& 12& 13& 10& 15& 13 & 13 & 12 \\
ConvKB    & 14& 15& 15& 14& 14& 14& 15& 10& 15 & 15 & 13 \\
SimplE    & 15& 16& 16& 16& 16& 16& 17& 20& 12 & 12 & 15 \\
CompGCN   & 16& 14& 9 & 12& 15& 15& 13& 7 & 17 & 16 & 17 \\
ANALOGY   & 17& 17& 17& 17& 17& 17& 16& 18& 14 & 13 & 16 \\
STransE   & 18& 18& 18& 18& 18& 18& 18& 12& 19 & 19 & 18 \\
NodePiece & 19& 19& 19& 19& 20& 19& 19& 16& 20 & 20 & 20 \\
R-GCN     & 20& 20& 20& 20& 19& 20& 20& 19& 18 & 18 & 19 \\
\bottomrule
\end{tabular}%
}
\end{table}

\section{Results and Analysis}
\label{sec:results}
This section reports results for two KGC prediction tasks: tail prediction $(h,r,?)$ and relation prediction $(h,?,t)$. To improve the reliability of the evaluation, all reported results average calculations over the Full-Set of selected KGC models and both LOMO and LOGO removal strategies, where individual KGC methods and entire KGC model groups are sequentially removed. The analysis begins with the tail prediction task, where seven MCDM aggregators are evaluated across five dimensions: consistency, stability, metric independence, robustness, and generalizability.

To ensure fair comparison across heterogeneous evaluation criteria, all metrics are normalized so that higher values indicate better performance and are bounded by 1.0. Specifically, MRR is retained in its native $[0,1]$ scale; Hits@1 and Hits@10 are scaled by dividing each model's value by the column maximum; and MR, being a non-beneficial metric, is transformed using $\text{MR}_{\text{norm}} = \min(\text{MR}) / \text{MR}$. This approach ensures that the best-performing model reaches 1.0 while preserving the relative ordering of models and avoiding zero values that could affect multiplicative aggregators.

\subsection{Tail Prediction Analysis}
\label{sec:results:tail}
Table~\ref{tab:tail_link_prediction_all} reports the base tail prediction results across five benchmark datasets. A clear pattern emerges: model performance is highly fragmented across metrics and datasets. No single KGC architecture dominates all evaluation criteria, and strong performance on one benchmark does not consistently transfer to others. For example, DualE achieves the best MR on FB15k, but this advantage is not consistent across other datasets. In contrast, models such as ComplEx, RotatE, TuckER, and HakE show varying strengths across metrics. This confirms that comparisons based on individual metrics can lead to inconsistent conclusions. The role of MCDM is therefore not to replace the original metrics but to combine these fragmented signals into a more stable and interpretable ranking.

The corresponding aggregator base scores and model rankings are reported in Tables~\ref{tab:tail_mcdm-kgc-scores} and~\ref{tab:tail_mcdm-kgc-ranks}. These results show that the choice of aggregator influences the final ranking. While the top models are generally similar, their ordering changes across EDAS, TOPSIS, VIKOR, Borda, Z-score, MOORA, and WASPAS. The next part of this section evaluates these aggregators using the five criteria defined in Section~\ref{subsec:meta_evaluation} to identify the most reliable method for tail prediction.

\begin{table*}
\centering
\caption{\small Correlation Matrices for MCDM Methods with Key Metrics, $(h, r, ?)$}
\label{tab:t_correlation-matrices-relation}
\scriptsize
\renewcommand{\arraystretch}{0.99}
\resizebox{\textwidth}{!}{%
\begin{tabular}{lcccccccccccc c}
\toprule
\textbf{MCDM} & \multicolumn{4}{c}{\textbf{Kendall $\tau$ ($\uparrow$)}} & \multicolumn{4}{c}{\textbf{Pearson $r$ ($\uparrow$)}} & \multicolumn{4}{c}{\textbf{Spearman $\rho$ ($\uparrow$)}} & \textbf{Overall} \\
\cmidrule(lr){2-5} \cmidrule(lr){6-9} \cmidrule(lr){10-13} \cmidrule(lr){14-14}
& H1 & H10 & MR & MRR
& H1 & H10 & MR & MRR
& H1 & H10 & MR & MRR & Corr. \\
\midrule
MOORA   & 0.9984 & 0.9984 & 0.9984 & 0.9984 & 0.9983 & 0.9983 & 0.9983 & 0.9983 & 0.9997 & 0.9997 & 0.9997 & 0.9997 & \textbf{0.9988} \\
EDAS    & 0.9944 & 0.9944 & 0.9944 & 0.9944 & 0.9974 & 0.9974 & 0.9974 & 0.9974 & 0.9986 & 0.9986 & 0.9986 & 0.9986 & 0.9968 \\
Z-Mean  & 0.9875 & 0.9875 & 0.9875 & 0.9875 & 0.9963 & 0.9963 & 0.9963 & 0.9963 & 0.9973 & 0.9973 & 0.9973 & 0.9973 & 0.9937 \\
TOPSIS  & 0.9836 & 0.9836 & 0.9836 & 0.9836 & 0.9956 & 0.9956 & 0.9956 & 0.9956 & 0.9964 & 0.9964 & 0.9964 & 0.9964 & 0.9919 \\
Borda   & 0.9817 & 0.9817 & 0.9817 & 0.9817 & 0.9956 & 0.9956 & 0.9956 & 0.9956 & 0.9968 & 0.9968 & 0.9968 & 0.9968 & 0.9914 \\
WASPAS  & 0.9803 & 0.9803 & 0.9803 & 0.9803 & 0.9926 & 0.9926 & 0.9926 & 0.9926 & 0.9943 & 0.9943 & 0.9943 & 0.9943 & 0.9891 \\
VIKOR   & 0.9479 & 0.9479 & 0.9479 & 0.9479 & 0.9778 & 0.9778 & 0.9778 & 0.9778 & 0.9786 & 0.9786 & 0.9786 & 0.9786 & 0.9681 \\
\bottomrule
\end{tabular}%
}
\end{table*}

\textbf{Consistency Test:}
To evaluate the consistency of each MCDM aggregator, three correlation measures, Kendall's $\tau$, Pearson's $r$, and Spearman's $\rho$, are computed between each aggregator's ranking and base performance metrics (MRR, Hits@1, Hits@10, MR). 
The final consistency score for an aggregator is the mean of the three correlation coefficients averaged across all runs, and the standard deviation across the three coefficients is reported as the stability error indicator.

\begin{figure}
\centering
\includegraphics[scale=.64]{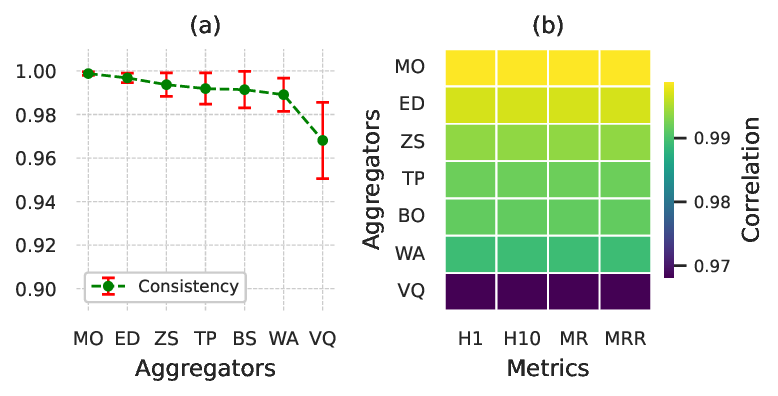}
\caption{\textit{(a)} Consistency scores of MCDM aggregators with standard-deviation error bars (over one run with the Full-Set of $20$ KGC models  and 20 LOMO runs with $19$ models at a time and $4$ LOGO runs with one group at a time). \textit{(b)} Heatmap of the average correlation between each aggregator and the base metrics across all $24$ runs. Brighter cells denote stronger correlation.}
\label{fig:consistency_hmp}
\end{figure}

Figure~\ref{fig:consistency_hmp}\,\textit{(a)} reports the consistency scores with standard-deviation error bars. MOORA, EDAS, and Z-score achieve the highest and most stable consistency, indicating that they almost perfectly preserve the base-metric ordering. TOPSIS and Borda remain highly consistent; WASPAS shows a noticeable drop, while VIKOR records a lower score among the others and the widest error bar, indicating high sensitivity to models or groups removed. MOORA, EDAS, and Z-score provide more stable performance because they use smooth, monotonic transformations, whereas VIKOR relies on dynamic reference points sensitive to extreme values.

Figure~\ref{fig:consistency_hmp}\,\textit{(b)} shows the heatmap of average correlation between each aggregator and each base metric across all $24$ runs. The vertical pattern matches panel\,\textit{(a)}: MOORA and EDAS are highly correlated, followed by Z-score, TOPSIS, Borda, and WASPAS. While VIKOR shows lower correlation among others across all metrics. 

The detailed correlation results are reported in Table~\ref{tab:t_correlation-matrices-relation}. Overall, MOORA, EDAS, and Z-score deliver the most robust and consistent aggregation.
\begin{table}
\centering
\scriptsize\caption{\small Variance of rankings for each knowledge graph embedding model across the five benchmark datasets, computed separately for each MCDM method. Lower variance indicates less fluctuation of ranks across datasets.}
\label{tab:variance_rankings}
\resizebox{\columnwidth}{!}{
\begin{tabular}{lccccccc}
\hline
\textbf{Model} & \textbf{BS} & \textbf{ZS} & \textbf{MO} & \textbf{TP} & \textbf{VQ} & \textbf{WA} & \textbf{ED} \\
\hline
ConvR & 2.724 & 2.747 & 2.427 & 2.388 & 2.273 & 2.536 & 2.636 \\
RotatE & 3.024 & 3.061 & 3.377 & 3.796 & 3.812 & 3.766 & 2.908 \\
CrossE & 5.574 & 6.183 & 5.185 & 4.944 & 5.498 & 5.575 & 4.492 \\
TuckER & 7.932 & 5.704 & 6.004 & 6.204 & 6.392 & 5.900 & 7.646 \\
ConvE & 7.949 & 5.906 & 6.621 & 5.952 & 7.398 & 6.412 & 6.235 \\
ComplEx & 7.171 & 6.834 & 7.456 & 7.189 & 8.102 & 9.412 & 7.208 \\
SimplE & 6.818 & 8.364 & 8.753 & 9.621 & 8.780 & 9.093 & 8.069 \\
DistMult & 12.640 & 10.499 & 10.376 & 10.681 & 10.344 & 11.273 & 10.099 \\
ANALOGY & 8.484 & 12.141 & 11.188 & 11.403 & 11.958 & 11.812 & 11.692 \\
RSN & 14.931 & 13.320 & 12.488 & 12.473 & 12.241 & 12.384 & 13.233 \\
TransE & 13.768 & 13.019 & 13.210 & 12.326 & 13.333 & 13.704 & 13.565 \\
NodePiece & 10.268 & 13.118 & 17.054 & 17.976 & 11.832 & 11.522 & 12.164 \\
HolE & 16.239 & 15.239 & 14.657 & 15.672 & 14.946 & 13.337 & 14.506 \\
HakE & 18.890 & 14.789 & 14.891 & 14.575 & 14.834 & 15.436 & 15.663 \\
TorusE & 17.866 & 17.920 & 17.626 & 17.738 & 17.400 & 16.486 & 17.476 \\
STransE & 17.811 & 18.648 & 16.606 & 16.331 & 19.442 & 17.995 & 19.183 \\
DualE & 21.660 & 19.795 & 20.556 & 20.616 & 20.397 & 20.446 & 19.813 \\
R-GCN & 19.482 & 22.212 & 23.201 & 24.053 & 20.719 & 21.234 & 19.170 \\
ConvKB & 24.771 & 28.963 & 27.306 & 25.416 & 26.607 & 26.979 & 27.788 \\
CompGCN & 43.048 & 40.023 & 37.826 & 37.978 & 39.248 & 39.326 & 39.758 \\
\hline
Mean Var. & 14.053 & 13.924 & 13.867& 13.840 & 13.778 & 13.731& 13.665\\
\hline
\end{tabular}
}
\end{table}

\textbf{Stability Test:}
To assess cross-dataset stability, we compute the variance of each model's rank across the five benchmark datasets, separately for every aggregator. For each aggregator, we first calculate the change in each model’s rank (its variance) and then take the average across all runs in which the model appears. The aggregator’s stability score is computed as the inverse of the average rank variance, normalized so that more stable rankings receive higher scores, with the most stable method scoring approximately $1.0$.

\begin{figure}
\centering
\includegraphics[scale=.78]{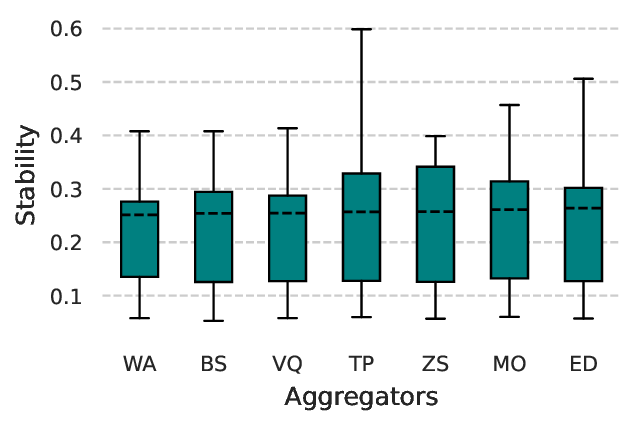}
\caption{Distribution of normalized rank variances across the five benchmark datasets.}
\label{fig:var_std}
\end{figure}
Figure~\ref{fig:var_std} presents the distribution of normalized rank variances as boxplots. TOPSIS, Z-score, MOORA, and EDAS exhibit progressively longer upper whiskers and higher maxima, reflecting a few models whose ranks shift more strongly under cross-dataset and cross-group variation. Importantly, the lower quartiles and medians remain close across all seven aggregators.

Table~\ref{tab:variance_rankings} summarizes the per-model rank variances and the normalized stability scores (the Mean Variance row, higher \,=\, more stable). The scores confirm that every aggregator delivers broadly stable rankings across the five datasets. Overall, no aggregator is destabilized by metric or group removal; the aggregator EDAS is marginally better.

\begin{table}
\caption{Independency analysis of MCDM aggregators under metric removal. Kendall's~$\tau$, Spearman's~$\rho$, and Pearson correlation indicate global ranking preservation (higher is better), and average absolute rank displacement ($|\Delta \text{Rank}|$) indicates local positional changes (lower is better).}
\label{tab:robustness-updated}
\centering
\scriptsize
\setlength{\tabcolsep}{2.5pt}
\renewcommand{\arraystretch}{1.15}
\resizebox{\columnwidth}{!}{%
\begin{tabular}{llccccccc}
\toprule
Removed & Metric & Borda & Z-Mean & TOPSIS & VIKOR & EDAS & MOORA & WASPAS \\
\midrule
\multirow{4}{*}{Kendall $\tau$}
& Without MR   & 0.853 & 0.832 & 0.579 & 0.561 & 0.800 & 0.705 & 0.674 \\
& Without MRR  & 0.874 & 0.937 & 0.947 & 0.989 & 0.947 & 0.958 & 0.968 \\
& Without H@1  & 0.853 & 0.895 & 0.821 & 0.758 & 0.811 & 0.811 & 0.895 \\
& Without H@10 & 0.968 & 0.979 & 0.958 & 0.979 & 0.937 & 0.958 & 0.958 \\
\midrule
\multirow{4}{*}{Spearman $\rho$}
& Without MR   & 0.961 & 0.952 & 0.752 & 0.707 & 0.941 & 0.862 & 0.785 \\
& Without MRR  & 0.962 & 0.985 & 0.988 & 0.998 & 0.989 & 0.992 & 0.995 \\
& Without H@1  & 0.947 & 0.971 & 0.940 & 0.908 & 0.932 & 0.950 & 0.970 \\
& Without H@10 & 0.991 & 0.997 & 0.991 & 0.997 & 0.989 & 0.992 & 0.989 \\
\midrule
\multirow{4}{*}{Pearson}
& Without MR   & 0.961 & 0.952 & 0.752 & 0.704 & 0.941 & 0.862 & 0.785 \\
& Without MRR  & 0.962 & 0.985 & 0.988 & 0.998 & 0.989 & 0.992 & 0.995 \\
& Without H@1  & 0.947 & 0.971 & 0.940 & 0.908 & 0.932 & 0.950 & 0.970 \\
& Without H@10 & 0.991 & 0.997 & 0.991 & 0.997 & 0.989 & 0.992 & 0.989 \\
\midrule
\multirow{4}{*}{$|\Delta$ Rank|}
& Without MR   & 1.100 & 1.200 & 3.100 & 3.300 & 1.500 & 2.400 & 2.400 \\
& Without MRR  & 1.000 & 0.600 & 0.500 & 0.100 & 0.500 & 0.400 & 0.300 \\
& Without H@1  & 1.400 & 0.900 & 1.500 & 1.900 & 1.400 & 1.300 & 0.900 \\
& Without H@10 & 0.300 & 0.200 & 0.400 & 0.200 & 0.500 & 0.400 & 0.400 \\
\bottomrule
\end{tabular}}
\end{table}

\textbf{Independency Test:}
To assess aggregator robustness under partial information loss, each base metric family is removed, and each aggregator is re-scored on the reduced metric set, and the correlation with the base ranks is computed. This isolates how much each aggregator's ordering depends on any single metric and on the availability of particular models or groups. For each scenario, we report Kendall 's~$\tau$, Spearman 's~$\rho$, and Pearson 's~$r$, which capture global ranking preservation (higher is better), together with the average absolute rank displacement ($|\Delta\text{Rank}|$), which captures local positional change (inverted to higher is better). The independency score for each aggregator is the mean of the three correlations across all scenarios.

\begin{figure}
\centering
\includegraphics[scale=.85]{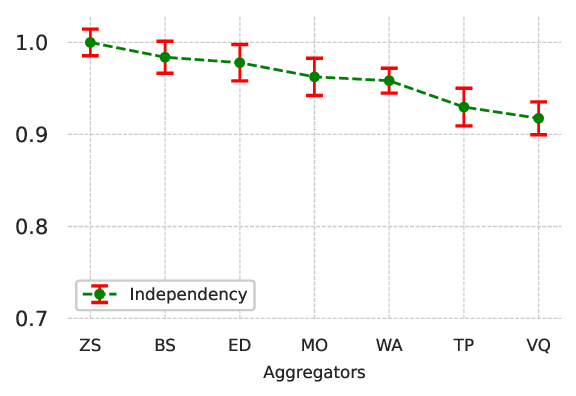}
\caption{Independence scores of MCDM aggregators, sorted from highest to lowest. Error bars indicate $\pm 1$ standard deviation across Kendall, Pearson, and Spearman correlations.}
\label{fig:robustness}
\end{figure}

Figure~\ref{fig:robustness} presents the normalized independency scores with error bars representing them, sorted from highest to lowest. The Z-score shows the highest independence with the tightest error bars, followed closely by Borda and EDAS, indicating that their rankings are largely preserved when an evaluation metric is removed or when models and groups of the KGC method are removed. While VIKOR shows the lowest scores, revealing high sensitivity to incomplete information. 

Table~\ref{tab:robustness-updated} reports the per-scenario values behind these scores. The higher sensitivity of each aggregator occurs when MR is removed, because MR is the most distinct, scale-sensitive criterion and provides information that bounds effectiveness metrics. Overall, the Z-score is a metric-independent aggregator, followed by Borda and EDAS for incomplete or fluctuating evaluation data.

\textbf{Robustness Test:}
To evaluate robustness under measurement uncertainty, multiplicative noise is injected into the base metrics (MRR, Hits@1, Hits@10, MR), with each value variation by an independent random factor and clipped to its valid range. Four noise levels ($5\%$, $10\%$, $20\%$, $30\%$) are applied, each over $1{,}000$ Monte Carlo iterations that re-sample the noise. In every iteration, the same LOMO and LOGO removals are run on the noisy data, the aggregator scores are averaged across scenarios, and the resulting ranking is compared against the clean, removal-free baseline using the mean absolute rank displacement ($|\Delta\mathrm{Rank}|$ inverted to higher is better) and the average correlation with the baseline. This jointly measures sensitivity to both noise and the absence of KGC models.

\begin{figure}
    \centering
    \includegraphics[scale=.80]{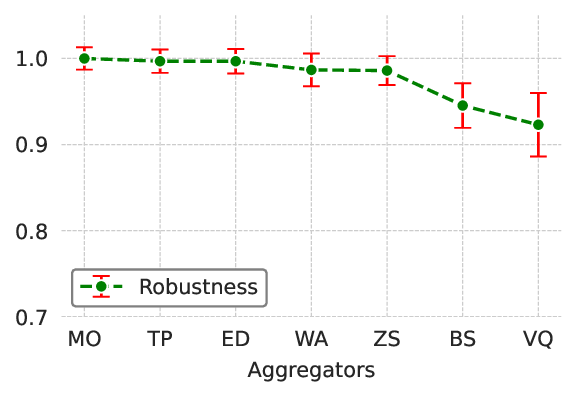}
    \caption{Robustness of MCDM aggregators under noise injection, averaged over all noise levels. Error bars indicate $\pm 1$ standard deviation across noise levels.}
    \label{fig:robustness}
\end{figure}

Figure~\ref{fig:robustness} shows the robustness scores averaged across all noise levels with error bars. MOORA, TOPSIS, and EDAS attain the highest baseline correlation with the tightest error bars, indicating that their rankings are least disturbed by noise and removal, followed by WASPAS and Z-score. Borda, and especially VIKOR, record the lowest scores, making them the most noise-sensitive.

\begin{figure}
    \centering
    \includegraphics[scale=.63]{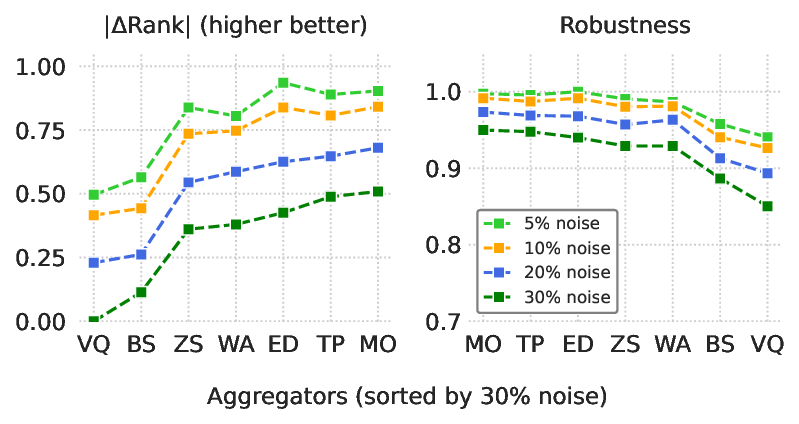}
    \caption{Robustness assessment with noise injection ($1{,}000$ iterations per level) and KGC method removal. \textit{Left}: mean absolute rank displacement ($|\Delta\mathrm{Rank}|$, inverted so higher\,=\,more robust). \textit{Right}: average correlation with the baseline ranking (higher\,=\,more robust). Aggregators are sorted by $30\%$ noise performance.}
    \label{fig:noise_5_10_20}
\end{figure}

Figure~\ref{fig:noise_5_10_20} presents the detailed results sorted by $30\%$ noise. The \textit{left} panel reports $|\Delta\mathrm{Rank}|$ inverted, so higher is better, and the \textit{right} panel reports the average correlation with the baseline ranking. Both panels agree and remain ordered consistently as the noise level rises: MOORA, TOPSIS, and EDAS remain at the top across all four noise levels, while VIKOR sits at the bottom and degrades sharply, with Borda close behind. The overall finding of this test suggests that MOORA, TOPSIS, and EDAS perform most reliably under noisy and partial-evaluation conditions, whereas VIKOR is notably sensitive to both noise and the removal of the KGC method.


\begin{figure}
    \centering
    \includegraphics[scale=.8]{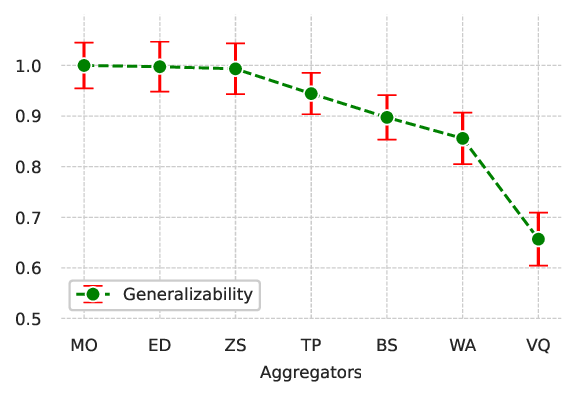}
    \caption{Generalizability of MCDM aggregators under leave-one-dataset-out validation. Error bars indicate $\pm 1$ standard deviation across folds.}
    \label{fig:generalizability}
\end{figure}

\begin{table}
\caption{Generalizability results for tail prediction using leave-one-dataset-out validation. Higher values indicate better cross-dataset transferability.}
\label{tab:final-generalizability}
\centering
\scriptsize
\setlength{\tabcolsep}{3pt}
\renewcommand{\arraystretch}{1.1}
\resizebox{\columnwidth}{!}{%
\begin{tabular}{lccccccc}
\hline
\textbf{Aggregator} & \textbf{Top-1} & \textbf{Top-3} & $\boldsymbol{\tau}$ & \textbf{Pearson} & \textbf{Spearman} & \textbf{Overall} & \textbf{95\% CI} \\
\hline
MOORA   & 0.250 & 0.300 & 0.435 & 0.572 & 0.572 & 0.526 & [0.480, 0.573] \\
EDAS    & 0.250 & 0.600 & 0.437 & 0.569 & 0.569 & 0.525 & [0.474, 0.576] \\
Z-Mean  & 0.250 & 0.300 & 0.432 & 0.568 & 0.568 & 0.523 & [0.471, 0.575] \\
TOPSIS  & 0.250 & 0.300 & 0.405 & 0.543 & 0.543 & 0.497 & [0.455, 0.539] \\
Borda   & 0.100 & 0.300 & 0.377 & 0.520 & 0.520 & 0.472 & [0.427, 0.517] \\
WASPAS  & 0.050 & 0.300 & 0.362 & 0.495 & 0.495 & 0.450 & [0.398, 0.503] \\
VIKOR   & 0.000 & 0.000 & 0.271 & 0.385 & 0.381 & 0.346 & [0.292, 0.400] \\
\hline
\end{tabular}
}
\end{table}

\textbf{Generalizability Test:}
This test evaluates the cross-dataset transferability of MCDM aggregators using leave-one-dataset-out (LODO) validation. For each fold, one dataset is eliminated as unseen. The remaining datasets are concatenated and used to compute a transferred (predicted) ranking for each aggregator, which is then compared against the true ranking obtained directly on the eliminated dataset. Models removed in removal-scenarios are excluded from the comparison.

Figure~\ref{fig:generalizability} shows the aggregated scores with error bars, sorted from highest to lowest.
For each fold, we compute Top-1 accuracy (whether the top-ranked model in the transferred ranking is also the best on the unseen dataset), Top-3 accuracy (whether that best model falls within the transferred top three), and the rank correlations. The overall generalizability score shown in Figure~\ref{fig:generalizability} is the mean of the three correlations and a $95\%$ confidence interval, where higher values indicate stronger transfer to unseen datasets. Table~\ref{tab:final-generalizability} reports the detailed LODO results. 
 
This test suggests that MOORA, EDAS, and Z-score consistently yield the highest cross-dataset transferability, whereas VIKOR is the least robust under both dataset shift and aggregator removal.
MOORA, EDAS, and Z-score generalize well because they use smooth, balanced criterion weighting, whereas VIKOR is less generalizable since its ranking depends heavily on dataset-specific best/worst reference values, leading to lower correlation and higher variance on unseen data.

\subsubsection{Pareto Analysis}
\label{subsec:pareto}
To identify the most suitable MCDM aggregator for KGC evaluation, we perform a Pareto optimality analysis across all five test dimensions: consistency, stability, robustness, generalizability, and independence. 

\begin{figure}
	\centering
	\includegraphics[scale=.69]{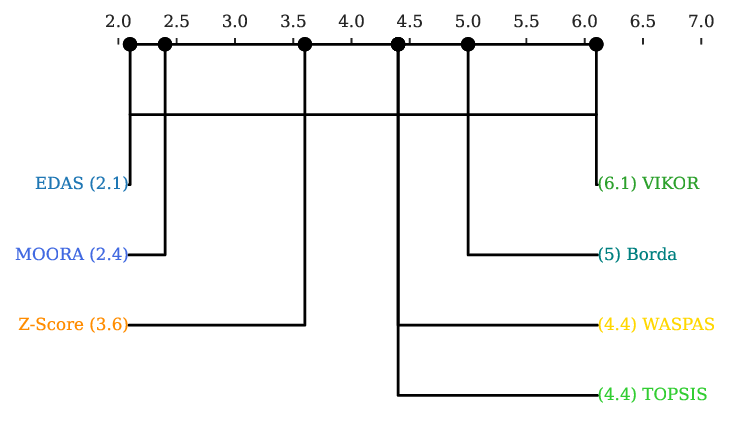}
	\caption{Critical difference (CD) diagram showing average ranks of MCDM aggregators across five evaluation dimensions. Methods connected by horizontal lines are not significantly different (Nemenyi post-hoc test, $\alpha=0.05$).}
	\label{fig:critical_difference}
\end{figure}
Figure~\ref{fig:critical_difference} presents a CD diagram based on Friedman ranking tests. The Friedman test determines whether observed differences are statistically significant ($p < 0.05$), followed by Nemenyi post-hoc analysis to compute the critical difference threshold. EDAS, MOORA, and Z-score achieve the best average ranks, while VIKOR performs the worst, confirming significant performance gaps across aggregators.

\begin{figure}
	\centering
	\includegraphics[scale=.66]{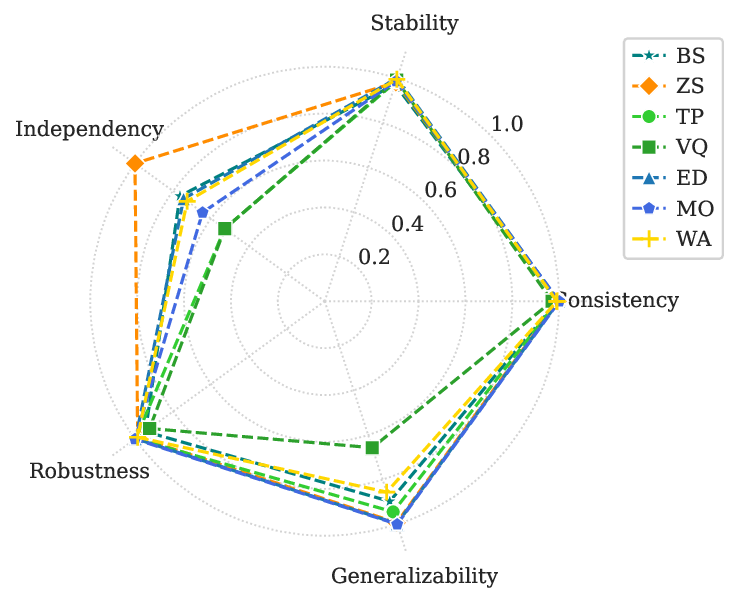}
	\caption{Radar plot of normalized performance scores across the five evaluation dimensions.}
	\label{fig:radar_tradeoff}
\end{figure}
Figure~\ref{fig:radar_tradeoff} shows the score (0–1 scale) for each aggregator across all five tests. The radar plot confirms that no aggregator dominates all dimensions, but Z-score and EDAS, followed by MOORA, consistently perform well across all criteria, whereas VIKOR underperforms in the independence and generalizability criteria.

\begin{figure}
	\centering
	\includegraphics[scale=.68]{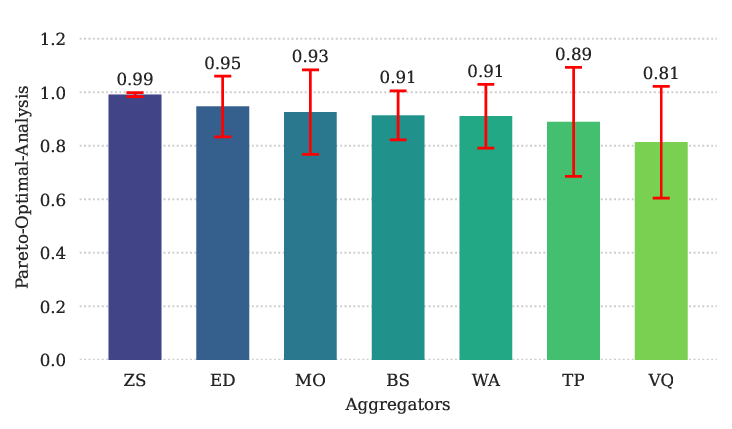}
	\caption{Overall performance scores for each MCDM aggregator are computed as the average normalized score across all five evaluation dimensions.}
	\label{fig:pareto}
\end{figure}
Figure~\ref{fig:pareto} presents the overall performance score for each aggregator.
Z-score achieves the highest overall score, indicating balanced performance across all dimensions. EDAS and MOORA follow the Z-score and also provide balanced performance compared to other aggregators. Error bars represent standard deviation across the five test metrics, highlighting variability across the five dimensions and confirming that the Z-score provides the most reliable aggregator. 

Based on the Pareto optimality analysis, \textbf{Z-score} is selected as the optimal aggregator for holistic KGC evaluation. Table~\ref{tab:zs-top3} presents the top three KGC methods for the tail prediction task as ranked exclusively by Z-score.
\begin{table}[t]
\centering
\caption{\small Top-3 KGC models ranked by the Z-score  aggregator (1 = best).}
\label{tab:zs-top3}
\footnotesize
\setlength{\tabcolsep}{6pt}
\begin{tabular*}{\columnwidth}{@{\extracolsep{\fill}}lccc}
\toprule
& \multicolumn{3}{c}{\textbf{Z-score }} \\
\cmidrule(lr){2-4}
\textbf{Model} & \textbf{DualE} & \textbf{HakE} & \textbf{ComplEx} \\
\midrule
\textbf{Rank} & 1 & 2 & 3 \\
\bottomrule
\end{tabular*}
\end{table}
DualE achieves the highest Z-score aggregate ranking (rank~1), indicating it provides the most balanced performance across all metrics (MR, MRR, hits@1, and hits@10) on the tail prediction task.

\subsection{Relation Prediction}
\label{sec:results:relation}
We apply the same meta-evaluation pipeline (introduced in Section~\ref{subsec:meta_evaluation} and fully detailed in the tail prediction analysis) to the relation prediction task $(h,?,t)$. This task uses MRR, Hits@1, and Hits@3 across six benchmark datasets: FB15K, FB15K-237, WN18, WN18RR, NELL995, and DDB14. Base results are reported in Table~\ref{tab:transductive_percentage}, while MCDM aggregation scores and induced rankings are presented in Tables~\ref{tab:mcdm-kgc-r_scores} and~\ref{tab:mcdm-kgc-r_ranks}.

\begin{table*}[h!]
\centering
\caption{Transductive performance comparison on all six datasets (results are taken from Flow-Modulated Scoring for Semantic-Aware Knowledge Graph Completion).}
\label{tab:transductive_percentage}
\resizebox{\textwidth}{!}{%
\renewcommand{\arraystretch}{1.2}
\begin{tabular}{ll|ccc|ccc|ccc|ccc|ccc|ccc}
\toprule
\textbf{Type} & \textbf{Model}
& \multicolumn{3}{c|}{\textbf{FB15K}} & \multicolumn{3}{c|}{\textbf{FB15K-237}} & \multicolumn{3}{c|}{\textbf{WN18}} & \multicolumn{3}{c|}{\textbf{WN18RR}} & \multicolumn{3}{c|}{\textbf{NELL995}} & \multicolumn{3}{c}{\textbf{DDB14}} \\
& & \textbf{MRR} & \textbf{H@1} & \textbf{H@3}
& \textbf{MRR} & \textbf{H@1} & \textbf{H@3}
& \textbf{MRR} & \textbf{H@1} & \textbf{H@3}
& \textbf{MRR} & \textbf{H@1} & \textbf{H@3}
& \textbf{MRR} & \textbf{H@1} & \textbf{H@3}
& \textbf{MRR} & \textbf{H@1} & \textbf{H@3} \\
\midrule
\multirow{6}{*}{Translational/Semantic-based}
& TransE & 96.2 & 94.0 & 98.2 & 96.6 & 94.6 & 98.4 & 97.1 & 95.5 & 98.4 & 78.4 & 66.9 & 87.0 & 84.1 & 78.1 & 88.9 & 96.6 & 94.8 & 98.0 \\
& ComplEx & 90.1 & 84.4 & 95.2 & 92.4 & 87.9 & 97.0 & 98.5 & 97.9 & 99.1 & 84.0 & 77.7 & 88.0 & 70.3 & 62.5 & 76.5 & 95.3 & 93.1 & 96.8 \\
& DistMult & 66.1 & 43.9 & 86.8 & 87.5 & 80.6 & 93.6 & 78.6 & 58.4 & 98.7 & 84.7 & 78.7 & 89.1 & 63.4 & 52.4 & 72.0 & 92.7 & 88.6 & 96.1 \\
& RotatE & 97.9 & 96.7 & 98.6 & 97.0 & 95.1 & 98.0 & 98.4 & 97.9 & 98.6 & 79.9 & 73.5 & 82.3 & 72.9 & 69.1 & 75.6 & 95.3 & 93.4 & 96.4 \\
& SimplE & 98.3 & 97.2 & 99.1 & 97.1 & 95.5 & 98.7 & 97.2 & 96.4 & 97.6 & 73.0 & 65.9 & 75.5 & 71.6 & 67.1 & 74.8 & 92.4 & 89.2 & 94.8 \\
& QuatE & 98.3 & 97.2 & 99.1 & 97.4 & 95.8 & 98.8 & 98.1 & 97.5 & 98.3 & 82.3 & 76.7 & 85.2 & 75.2 & 70.6 & 78.3 & 94.6 & 92.2 & 96.2 \\
&&&&&&&&&&&&&&&&&&&\\
\midrule
\multirow{2}{*}{Rule-based}
& DRUM & 94.5 & 94.5 & 97.8 & 95.9 & 90.5 & 95.8 & 96.9 & 95.6 & 98.0 & 85.4 & 77.8 & 91.2 & 71.5 & 64.0 & 74.0 & 95.8 & 93.0 & 98.7 \\
&&&&&&&&&&&&&&&&&&&\\
\midrule
\multirow{7}{*}{GNN-based}
& R-GCN & 69.6 & 60.1 & 76.0 & 93.1 & 90.3 & 94.2 & 90.9 & 82.4 & 97.9 & 82.2 & 75.0 & 84.8 & 78.5 & 71.2 & 82.6 & 96.2 & 93.8 & 97.9 \\
& Path & 93.7 & 91.8 & 95.1 & 97.2 & 95.7 & 98.6 & 98.1 & 97.1 & 98.9 & 93.3 & 89.7 & 96.1 & 73.7 & 68.5 & 76.4 & 96.9 & 94.8 & 99.1 \\
& Con & 96.2 & 93.4 & 98.8 & 97.8 & 96.1 & 99.5 & 96.0 & 92.7 & 99.2 & 94.3 & 89.4 & 99.3 & 87.5 & 81.5 & 92.8 & 97.7 & 96.1 & 99.4 \\
& PathCon & 98.4 & 97.4 & 99.5 & 97.9 & 96.4 & 99.4 & 99.3 & 98.8 & 99.8 & 97.4 & 95.4 & 99.4 & 89.6 & 84.4 & 94.1 & 98.0 & 96.6 & 99.5 \\
& FMS & 99.6 & 99.5 & 99.8 & 99.8 & 99.7 & 99.9 & 99.7 & 99.5 & 99.9 & 99.9 & 99.8 & 99.9 & 99.1 & 98.6 & 99.5 & 99.8 & 99.7 & 99.9 \\
&&&&&&&&&&&&&&&&&&&\\
\bottomrule
\end{tabular}
}
\end{table*}

\begin{table*}[h!]
\centering
\caption{\small Aggregation scores for knowledge graph models.}
\label{tab:mcdm-kgc-r_scores}
\setlength{\tabcolsep}{12pt}
\tiny
\renewcommand{\arraystretch}{1.0}
\resizebox{\textwidth}{!}{%
\begin{tabular}{lcccccccccc}
\toprule
\textbf{Model} & \textbf{ED} & \textbf{TP} & \textbf{VQ} & \textbf{BS} & \textbf{ZS} & \textbf{MO} & \textbf{WA} & \textbf{MRR} & \textbf{H1} & \textbf{H3} \\
\midrule
FMS & 1.0000 & 1.0000 & 1.0000 & 1.0000 & 1.0000 & 0.3195 & 1.0000 & 1.0000 & 1.0000 & 1.0000 \\
PathCon & 0.8450 & 0.8282 & 0.7753 & 0.9120 & 0.8545 & 0.3092 & 0.8757 & 0.9710 & 0.9533 & 0.9880 \\
Con & 0.7457 & 0.7613 & 0.7014 & 0.7199 & 0.7704 & 0.3030 & 0.8023 & 0.9524 & 0.9201 & 0.9834 \\
Path & 0.5972 & 0.6067 & 0.3982 & 0.5880 & 0.6377 & 0.2925 & 0.6622 & 0.9245 & 0.9005 & 0.9419 \\
TransE & 0.5662 & 0.6105 & 0.2973 & 0.5000 & 0.5713 & 0.2905 & 0.5768 & 0.9182 & 0.8778 & 0.9499 \\
QuatE & 0.5369 & 0.5848 & 0.3681 & 0.5509 & 0.5134 & 0.2883 & 0.5410 & 0.9129 & 0.8878 & 0.9281 \\
RotatE & 0.4996 & 0.5544 & 0.3080 & 0.4722 & 0.4934 & 0.2853 & 0.5168 & 0.9054 & 0.8806 & 0.9174 \\
DRUM & 0.4792 & 0.5395 & 0.2701 & 0.4120 & 0.4720 & 0.2842 & 0.5000 & 0.9030 & 0.8633 & 0.9274 \\
ComplEx & 0.4160 & 0.4971 & 0.2949 & 0.4120 & 0.4288 & 0.2799 & 0.4810 & 0.8873 & 0.8433 & 0.9226 \\
SimplE & 0.4170 & 0.5101 & 0.1679 & 0.3773 & 0.3463 & 0.2788 & 0.2159 & 0.8857 & 0.8565 & 0.9024 \\
R-GCN & 0.2476 & 0.3700 & 0.1283 & 0.3426 & 0.2710 & 0.2685 & 0.2206 & 0.8537 & 0.7921 & 0.8906 \\
DistMult & 0.0001 & 0.1776 & 0.0001 & 0.2130 & 0.0010 & 0.2489 & 0.0721 & 0.7909 & 0.6742 & 0.8954 \\
\bottomrule
\end{tabular}%
}
\end{table*}

\begin{table}[t]
\centering
\caption{\small Rhajianking of knowledge graph embedding models based on MCDM aggregator methods (1 = best).}
\label{tab:mcdm-kgc-r_ranks}
\setlength{\tabcolsep}{3pt} 
\renewcommand{\arraystretch}{0.9} 
\footnotesize
\resizebox{\columnwidth}{!}{%
\begin{tabular}{lcccccccccc}
\toprule
\textbf{Model} & \textbf{ED} & \textbf{TP} & \textbf{VQ} & \textbf{BS} & \textbf{ZS} & \textbf{MO} & \textbf{WA} & \textbf{MRR} & \textbf{H1} & \textbf{H3} \\
\midrule
FMS & 1 & 1 & 1 & 1 & 1 & 1 & 1 & 1 & 1 & 1 \\
PathCon & 2 & 2 & 2 & 2 & 2 & 2 & 2 & 2 & 2 & 2 \\
Con & 3 & 3 & 3 & 3 & 3 & 3 & 3 & 3 & 3 & 3 \\
Path & 4 & 5 & 4 & 4 & 4 & 4 & 4 & 4 & 4 & 5 \\
TransE & 5 & 4 & 7 & 6 & 5 & 5 & 5 & 5 & 7 & 4 \\
QuatE & 6 & 6 & 5 & 5 & 6 & 6 & 6 & 6 & 6 & 6 \\
RotatE & 7 & 7 & 6 & 7 & 7 & 7 & 7 & 7 & 6 & 9 \\
DRUM & 8 & 8 & 9 & 8 & 8 & 8 & 8 & 8 & 8 & 7 \\
ComplEx & 10 & 10 & 8 & 8 & 9 & 9 & 9 & 9 & 10 & 8 \\
SimplE & 9 & 9 & 10 & 10 & 10 & 10 & 11 & 10 & 9 & 10 \\
R-GCN & 11 & 11 & 11 & 11 & 11 & 11 & 10 & 11 & 11 & 12 \\
DistMult & 12 & 12 & 12 & 12 & 12 & 12 & 12 & 12 & 12 & 11 \\
\bottomrule
\end{tabular}%
}
\end{table}

\textbf{Consistency Test:}
To compute consistency for relation prediction, we follow the same procedure described in the tail prediction consistency test. 

Figure~\ref{fig:r_consistency_hmp}~\textit{(a)} shows consistency scores with standard-deviation error bars. MOORA and Z-score have the highest consistency scores, with minimal variation, while VIKOR shows the lowest consistency among the others. The reason VIKOR performs worse is the calculation method.  
Figure~\ref{fig:r_consistency_hmp}~\textit{(b)} shows a comparison between aggregators and base metrics, revealing a strong correlation among others, whereas, similarly, VIKOR produces the distinct ranking behavior. More detailed results are reported in Table~\ref{tab:correlation-matrices-relation}. 
Overall, these results confirm that all MCDM aggregators except VIKOR reliably preserve the information in the original evaluation metrics.

\begin{figure}
    \centering
    \includegraphics[scale=.60]{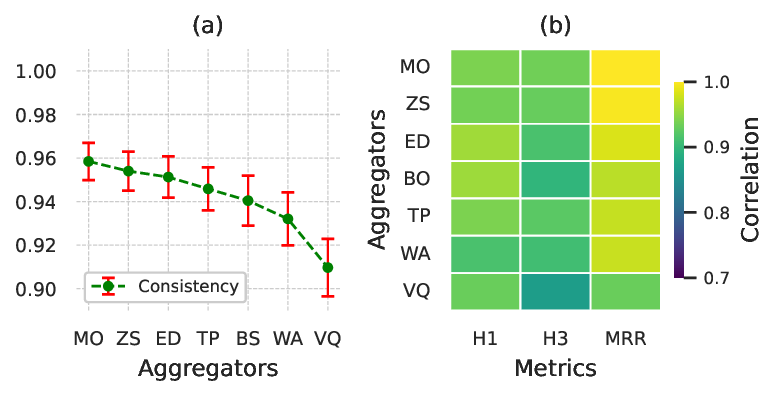}
    \caption{Relation consistency \textit{(a)} Consistency test of MCDM aggregators. \textit{(b)} Correlation between aggregators and base metrics.}
    \label{fig:r_consistency_hmp}
\end{figure}

\begin{table*}[h!]
\centering
\caption{\small Correlation Matrices for MCDM Methods with Key Metrics, (H1, H3, MRR)}
\label{tab:correlation-matrices-relation}
\scriptsize
\tiny
\renewcommand{\arraystretch}{0.99}
\resizebox{\textwidth}{!}{%
\begin{tabular}{l ccc ccc ccc c}
\toprule
\textbf{MCDM} & \multicolumn{3}{c}{\textbf{Kendall $\tau$ ($\uparrow$)}} & \multicolumn{3}{c}{\textbf{Pearson $r$ ($\uparrow$)}} & \multicolumn{3}{c}{\textbf{Spearman $\rho$ ($\uparrow$)}} & \textbf{Overall} \\
\cmidrule(lr){2-4} \cmidrule(lr){5-7} \cmidrule(lr){8-10} \cmidrule(lr){11-11}
& H1 & H3 & MRR
& H1 & H3 & MRR
& H1 & H3 & MRR & Corr. \\
\midrule
MOORA     & 0.8967 & 0.8800 & 1.0000 & 0.9625 & 0.9623 & 1.0000 & 0.9625 & 0.9623 & 1.0000 & 0.9585 \\
Z-score   & 0.8908 & 0.8740 & 0.9940 & 0.9580 & 0.9578 & 0.9978 & 0.9580 & 0.9578 & 0.9978 & 0.9540 \\
EDAS      & 0.9289 & 0.8478 & 0.9678 & 0.9714 & 0.9460 & 0.9910 & 0.9714 & 0.9460 & 0.9910 & 0.9513 \\
TOPSIS    & 0.9084 & 0.8683 & 0.9474 & 0.9572 & 0.9511 & 0.9859 & 0.9572 & 0.9511 & 0.9859 & 0.9458 \\
WASPAS    & 0.8566 & 0.8443 & 0.9553 & 0.9418 & 0.9400 & 0.9843 & 0.9418 & 0.9400 & 0.9843 & 0.9321 \\
Borda     & 0.9231 & 0.8224 & 0.9428 & 0.9714 & 0.9336 & 0.9822 & 0.9730 & 0.9325 & 0.9829 & 0.9404 \\
VIKOR     & 0.8964 & 0.7904 & 0.8895 & 0.9486 & 0.9036 & 0.9533 & 0.9486 & 0.9036 & 0.9533 & 0.9097 \\
\bottomrule
\end{tabular}%
}
\end{table*}

\textbf{Stability Test:}
For relation prediction, the stability test follows the same cross-dataset variance procedure described in the tail prediction stability analysis. 

\begin{figure}
	\centering
	\includegraphics[scale=.70]{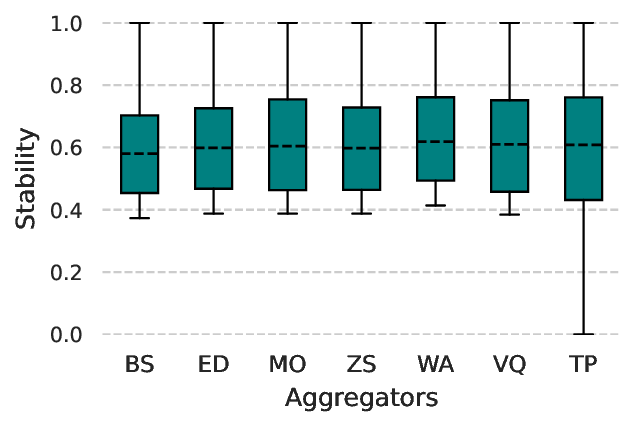}
	\caption{Distribution of ranking variance across relation prediction datasets for different aggregators.}
	\label{fig:r_var_std}
\end{figure}
Figure~\ref{fig:r_var_std} shows the distribution of ranking variance for each aggregator across the relation prediction datasets. A compact distribution indicates that the aggregator produces more stable rankings across benchmarks, whereas a wider box or longer whiskers indicate greater dataset-dependent fluctuation. Z-score and MOORA exhibit the most compact distributions, suggesting that their rankings remain more consistent across benchmark changes. In contrast, VIKOR and WASPAS show wider distributions and longer upper ranges. This suggests that their rankings are more sensitive to dataset-level changes.  The main finding of stability shows that Z-score and MOORA provide the most stable rankings for relation prediction.

\textbf{Independency Test:}
To assess aggregator robustness to partial information loss in relation prediction, we employ a leave-one-metric-out analysis following the same procedure as for tail prediction. Each base metric is systematically omitted, and the resulting aggregator rankings are compared against the baseline computed using the complete metric set.
\begin{figure}
	\centering
	\includegraphics[scale=.85]{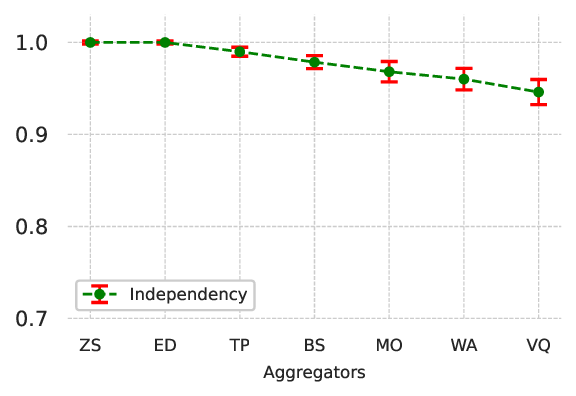}
	\caption{Metric independence of MCDM aggregators under leave-one-metric-family-out evaluation for relation prediction.}
	\label{fig:r_robustness}
\end{figure}

Figure~\ref{fig:r_robustness} presents independency scores with error bars representing $\pm 1$ standard deviation, sorted from highest to lowest correlation. Z-score and EDAS achieve the highest independency followed by TOPSIS, with minimal variation, indicating strong resistance to the removal of metrics. Overall, the independency test demonstrates that relation prediction rankings remain highly stable under metric removal, with Z-score and EDAS providing the strongest metric independence while VIKOR provides a very low independency score.

\begin{table}
\caption{Independency analysis of MCDM aggregators under metric removal for Relation Prediction. Kendall's~$\tau$, Spearman's~$\rho$, and Pearson correlation indicate global ranking preservation (higher is better), and average absolute rank displacement ($|\Delta \text{Rank}|$) indicates local positional changes (lower is better).}
\label{tab:r_robustness-relation}
\centering
\scriptsize
\setlength{\tabcolsep}{2.5pt}
\renewcommand{\arraystretch}{1.15}
\resizebox{\columnwidth}{!}{%
\begin{tabular}{llccccccc}
\toprule
Removed & Metric & Borda & Z-Mean & TOPSIS & VIKOR & EDAS & MOORA & WASPAS \\
\midrule
\multirow{3}{*}{Kendall $\tau$}
& Without MRR  & 0.992 & 0.970 & 1.000 & 0.909 & 0.970 & 0.939 & 0.909 \\
& Without H@1  & 0.992 & 1.000 & 0.970 & 0.879 & 1.000 & 0.939 & 0.909 \\
& Without H@3  & 0.870 & 1.000 & 0.939 & 0.909 & 1.000 & 0.909 & 0.939 \\
\midrule
\multirow{3}{*}{Spearman $\rho$}
& Without MRR  & 0.998 & 0.993 & 1.000 & 0.972 & 0.993 & 0.986 & 0.972 \\
& Without H@1  & 0.998 & 1.000 & 0.993 & 0.944 & 1.000 & 0.986 & 0.972 \\
& Without H@3  & 0.963 & 1.000 & 0.986 & 0.972 & 1.000 & 0.972 & 0.979 \\
\midrule
\multirow{3}{*}{Pearson}
& Without MRR  & 0.997 & 0.993 & 1.000 & 0.972 & 0.993 & 0.986 & 0.972 \\
& Without H@1  & 0.997 & 1.000 & 0.993 & 0.944 & 1.000 & 0.986 & 0.972 \\
& Without H@3  & 0.954 & 1.000 & 0.986 & 0.972 & 1.000 & 0.972 & 0.979 \\
\midrule
\multirow{3}{*}{$|\Delta$ Rank|}
& Without MRR  & 0.083 & 0.167 & 0.000 & 0.333 & 0.167 & 0.333 & 0.500 \\
& Without H@1  & 0.083 & 0.000 & 0.167 & 0.667 & 0.000 & 0.333 & 0.500 \\
& Without H@3  & 0.750 & 0.000 & 0.333 & 0.333 & 0.000 & 0.500 & 0.333 \\
\bottomrule
\end{tabular}}
\end{table}
Table~\ref{tab:r_robustness-relation} reports the per-scenario correlation and rank displacement values. Z-Mean and EDAS demonstrate the highest robustness across all metric-removal scenarios and zero-rank displacement when H@1 or H@3 is removed. The removal of H@3 is the most disruptive overall (average $|\Delta \text{Rank}| = 0.321$), while MRR removal has the mildest impact. Overall, Z-Mean and EDAS are the most metric-independent aggregators, followed by TOPSIS and Borda, making them suitable for relation prediction tasks with incomplete or fluctuating evaluation data.

\textbf{Robustness Test:}
The robustness test for relation prediction follows the same noise-injection procedure described in the tail prediction analysis. 

Figure~\ref{fig:r_robustness} shows the robustness scores with error bars representing $\pm 1$ standard deviation, sorted from highest to lowest. TOPSIS and MOORA achieve strong robustness with minimal variation, followed closely by Z-score. VIKOR again shows the lowest robustness, indicating substantially higher sensitivity to noise injection. 

\begin{figure}
	\centering
		\includegraphics[scale=.80]{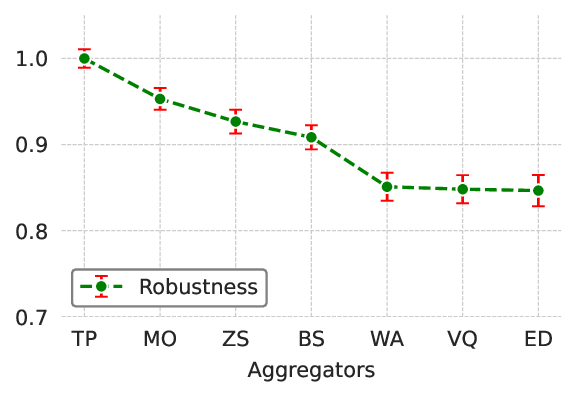}
	\caption{Robustness analysis under noise injection. }
	\label{fig:error_r_noise_5_10_20}
\end{figure}

\begin{figure}
	\centering
		\includegraphics[scale=.62]{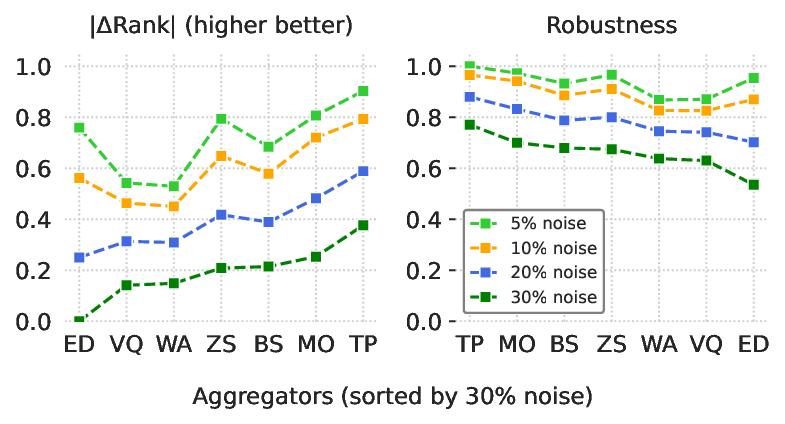}
	\caption{Robustness assessment with noise injection (1,000 iterations per level). \textit{Left}: Mean absolute rank displacement ($|\Delta\mathrm{Rank}|$, higher = more robust). \textit{Right}: average correlation with baseline ranking (higher = more robust).}
	\label{fig:r_noise_5_10_20}
\end{figure}

For more detail, Figure~\ref{fig:r_noise_5_10_20} presents two views: the left panel shows ranks changing in an inverted order, with higher being better, and the right panel shows average correlation with the baseline ranking. TOPSIS and MOORA are the most robust aggregators, maintaining high scores across both measures as noise increases due to their ratio-based scoring, which preserves relative separation among models. In contrast, VIKOR is the most sensitive, as its compromise-ranking mechanism amplifies small metric variations. Overall, TOPSIS and MOORA preserve ranking consistency more reliably under noise, whereas VIKOR is more affected.

\textbf{Generalizability Test:}
The relation prediction generalizability test follows the same leave-one-dataset-out validation, as tail prediction, examining whether an aggregator can predict rankings for an unseen benchmark using other existing results under KGC method removal robustness. 

Figure~\ref{fig:r_generalizability} presents the cross-dataset generalizability of aggregators for relation prediction. The Z-score achieves strong generalizability with minimal variance, demonstrating consistent ranking transferability across unseen benchmarks. VIKOR exhibits poor generalizability and high sensitivity to dataset distribution shifts, among other issues. 
\begin{figure}
	\centering
	\includegraphics[scale=.85]{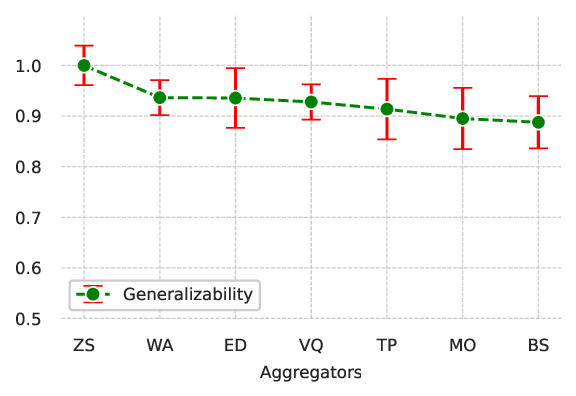}
	\caption{Generalizability analysis of MCDM aggregators under leave-one-dataset-out validation.}
	\label{fig:r_generalizability}
\end{figure}

\subsubsection{Pareto Analysis}
\label{sec:results:relation:pareto}
Combining the five evaluation dimensions (Figure~\ref{fig:r_pareto}), Z-score again achieves the best overall trade-off, followed by EDAS, TOPSIS, and MOORA. VIKOR remains poor.

Figure~\ref{fig:r_pareto} summarizes the overall trade-off among the aggregators after combining the normalized scores from the five evaluation dimensions. Z-score achieves the strongest overall Pareto performance. In contrast, VIKOR shows the weakest Pareto performance. Overall, the Pareto analysis identifies Z-score as the most suitable aggregator.

\begin{figure}
	\centering
	\includegraphics[scale=.68]{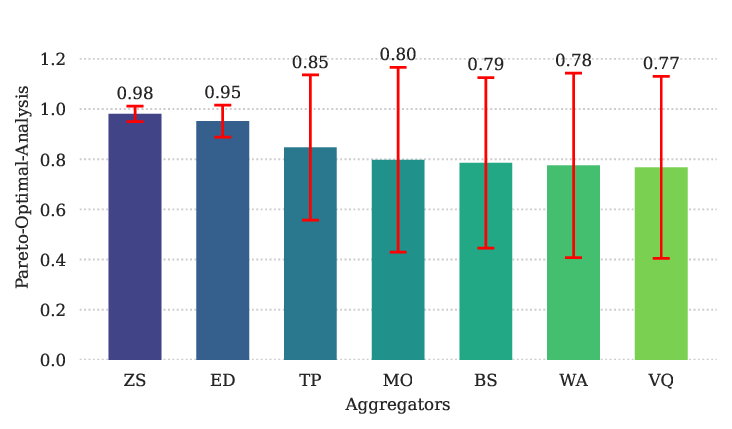}
	\caption{Pareto-based overall performance of MCDM aggregators for relation prediction.}
	\label{fig:r_pareto}
\end{figure}

\begin{table}[t]
\centering
\caption{\small Top-3 KGC models ranked by the Z-score aggregator for relation prediction (1 = best).}
\label{tab:zs-top3-relation}
\footnotesize
\setlength{\tabcolsep}{6pt}
\begin{tabular*}{\columnwidth}{@{\extracolsep{\fill}}lccc}
\toprule
& \multicolumn{3}{c}{\textbf{Z-score}} \\
\cmidrule(lr){2-4}
\textbf{Model} & \textbf{FMS} & \textbf{PathCon} & \textbf{Con} \\
\midrule
\textbf{Rank} & 1 & 2 & 3 \\
\bottomrule
\end{tabular*}
\end{table}
Using the Pareto-selected Z-score aggregator, Table~\ref{tab:zs-top3-relation} identifies FMS as the top-performing KGC model for relation prediction, followed by PathCon. 

\subsection{Test Sensitivity Analysis}
We evaluate how the five meta-evaluation tests respond to KGC model elimination for both tail and relation prediction tasks, using two strategies: LOMO and LOGO architectural families in Tables~\ref{tab:tail_link_prediction_all} and~\ref{tab:transductive_percentage}. For each removal, aggregated scores are recomputed on the remaining models, and the absolute change from the baseline (full set) is averaged across all removals to quantify test sensitivity.
\begin{figure}
    \centering
    \includegraphics[scale=.63]{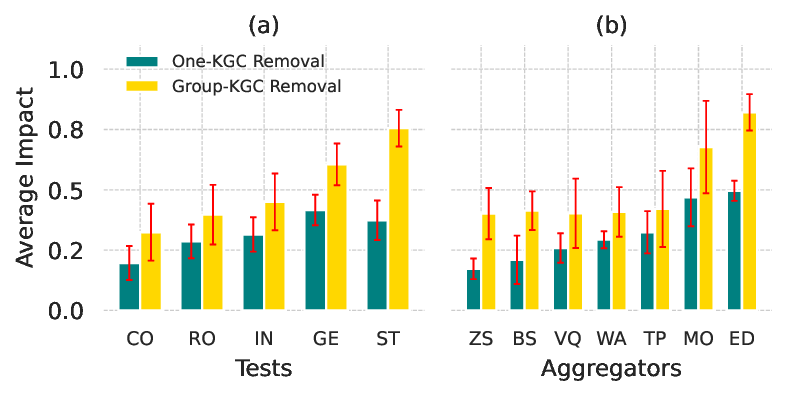}
    \caption{\textbf{\textit{Tail Prediction}} KGC-method removal sensitivity across the five meta-evaluation tests.}
    \label{fig:kgc_method_removal_change}
\end{figure}

\textbf{Tail prediction sensitivity analysis:} 
Figure~\ref{fig:kgc_method_removal_change} presents the average impact of KGC method removal for tests and aggregators under the tail prediction task. The error bars represent the 
standard error of the mean (SEM).  

Figure~\ref{fig:kgc_method_removal_change} \textbf{(a)} shows the impact of 
removal across tests. EDAS exhibits the highest sensitivity, followed by 
Generalizability and Stability. Consistency is the least sensitive test, 
making it the most robust against removal criteria. Independency and 
Robustness also show relatively low sensitivity.

Figure~\ref{fig:kgc_method_removal_change} \textbf{(b)} shows removal impact 
over aggregators. Z-score is the least sensitive, closely followed by 
Borda and VIKOR. In contrast, EDAS is the most sensitive aggregator, 
followed by MOORA and TOPSIS.

As observed, LOGO produces substantially larger average impacts than LOMO across all tests.
Removing an entire architectural family eliminates a set of models that exhibit systematic, dataset‑specific performance advantages. This alters the comparative rank ordering of the remaining models across datasets, directly inflating rank variance. Generalizability relies on learning transferable performance signatures from existing datasets to predict unseen benchmarks. Each architectural family contributes a distinct signature. Removing a family erases that predictive dimension from the existing set, significantly affecting it. In contrast, consistency measures static within‑dataset agreement between aggregators and base metrics, a property robust to compositional changes, explaining its minimal sensitivity.

\begin{figure}
    \centering
    \includegraphics[scale=.63]{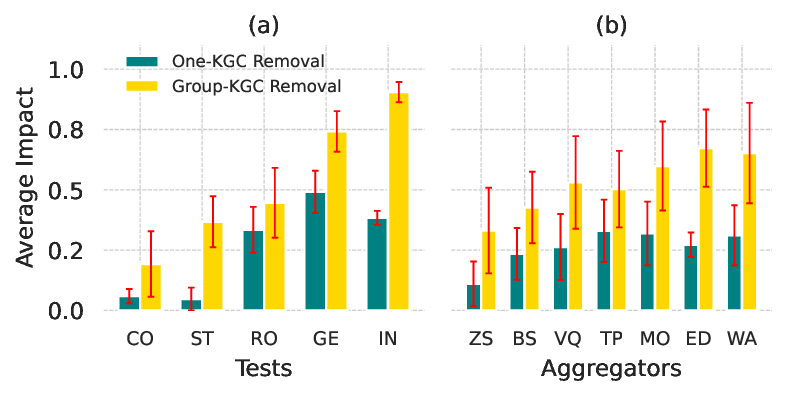}
    \caption{\textbf{\textit{Relation Prediction}} KGC-method removal sensitivity across the five meta-evaluation tests.}
    \label{fig:relation_kgc_method_removal_change}
\end{figure}

\textbf{Relation prediction sensitivity analysis:}
Figure~\ref{fig:relation_kgc_method_removal_change} shows relation prediction 
KGC method removal sensitivity analysis over given tests and aggregators. 

In Figure~\ref{fig:relation_kgc_method_removal_change} (a), Independency (IN) 
exhibits the highest sensitivity to KGC removal, followed by Robustness (RO) 
and Generalizability (GE). Conversely, Consistency (CO) and Stability (ST) 
are the least sensitive. Group-KGC removal consistently has a larger impact 
than One-KGC removal across all metrics.

Figure~\ref{fig:relation_kgc_method_removal_change} (b) shows that Z-score 
is the most robust aggregator (lowest sensitivity to KGC removal), followed 
by Borda (BS) and VIKOR (VQ). Group-KGC removal again shows substantially 
larger impact than One-KGC removal across all aggregators.

\section{Conclusion}
This study presented a five-dimensional meta-evaluation framework for metric aggregation in KGC. By formulating KGC benchmarking as an MCDM problem, we compared seven aggregation methods across consistency, stability, metric independence, robustness, and generalizability. The results support a no-free-lunch view: no aggregator is uniformly superior across all reliability dimensions and prediction tasks. However, Pareto-based trade-off analysis shows that Z-score provides the most balanced compromise in the evaluated setting, leading to consensus rankings in which DualE is strongest for tail prediction and FMS is strongest for relation prediction.
The main contribution of the work is not the replacement of standard KGC metrics, but rather the introduction of a principled procedure for deciding when and how to aggregate such metrics. By making aggregator trade-offs explicit, the framework helps reduce selective metric reporting and improves the interpretability of unified benchmark rankings. \textit{Future }work can extend the analysis to additional datasets, newer KGC and LLM-based reasoning models, alternative metric-weighting strategies, seed-level uncertainty, and application-specific evaluation criteria.

\section*{Authorship contribution statement}
\textit{Haji Gul:} Conceptualization, methodology, software, formal analysis, investigation, data curation, validation, Visualization, writing original draft.

\textit{Ajaz Ahmad Bhat:} Conceptualization, methodology, formal analysis, supervision, review, and improvement of methodological clarity and overall manuscript structure.

\section*{Data Availability}
Benchmark data are publicly available. Processed matrices and all code (aggregation, robustness, Pareto) will be released publicly upon acceptance.

\section*{Declaration of generative AI}
The authors used generative AI tools (including DeepSpeak and Qwen) to assist with drafting and refining portions of the manuscript and supporting aspects of code development. These tools were used as aids in the writing and implementation process; all substantive contributions, interpretations, and final decisions were made by the authors.

\bibliographystyle{model1-num-names}
\bibliography{cas-refs}

@article{gul2025kg,
  title={KG-EDAS: A Meta-Metric Framework for Evaluating Knowledge Graph Completion Models},
  author={Gul, Haji and Naim, Abul Ghani and Bhat, Ajaz Ahmad},
  journal={arXiv preprint arXiv:2508.15357},
  year={2025}
}

@article{ruffinelli2020you,
  title={You can teach an old dog new tricks! on training knowledge graph embeddings},
  author={Ruffinelli, Daniel and Broscheit, Samuel and Gemulla, Rainer},
  year={2020},
  publisher={ICLR}
}

@article{sun2020benchmarking,
  title={A benchmarking study of embedding-based entity alignment for knowledge graphs},
  author={Sun, Zequn and Zhang, Qingheng and Hu, Wei and Wang, Chengming and Chen, Muhao and Akrami, Farahnaz and Li, Chengkai},
  journal={arXiv preprint arXiv:2003.07743},
  year={2020}
}

@article{10.1145/3424672,
author = {Rossi, Andrea and Barbosa, Denilson and Firmani, Donatella and Matinata, Antonio and Merialdo, Paolo},
title = {Knowledge Graph Embedding for Link Prediction: A Comparative Analysis},
year = {2021},
issue_date = {April 2021},
publisher = {Association for Computing Machinery},
address = {New York, NY, USA},
volume = {15},
number = {2},
issn = {1556-4681},
url = {https://doi.org/10.1145/3424672},
doi = {10.1145/3424672},
journal = {ACM Trans. Knowl. Discov. Data},
month = jan,
articleno = {14},
numpages = {49}
}

@article{rossi2021knowledge,
  title={Knowledge graph embedding for link prediction: A comparative analysis},
  author={Rossi, Andrea and Barbosa, Denilson and Firmani, Donatella and Matinata, Antonio and Merialdo, Paolo},
  journal={TKDD},
  year={2021}
}

@inproceedings{yao2019kgbert,
  author    = {Yao, L. and Mao, C. and Luo, Y.},
  title     = {KG-BERT: BERT for knowledge graph completion},
  booktitle = {Proc. of AAAI},
  year      = {2019}
}

@inproceedings{zhang2020hake,
  author    = {Zhang, Z. and Cai, J. and Zhang, Y. and Wang, J.},
  title     = {Learning Hierarchy-Aware Knowledge Graph Embeddings for Link Prediction},
  booktitle = {Proc. of AAAI},
  year      = {2020}
}

@inproceedings{wei2024kicgpt,
  author    = {Wei, Y. and Huang, Q. and Zhang, Y. and Kwok, J.},
  title     = {KICGPT: Large Language Model with Knowledge in Context for KGC},
  booktitle = {Proc. of EMNLP},
  year      = {2023}
}

@inproceedings{rao2024calibration,
  title={Using Model Calibration to Evaluate Link Prediction in Knowledge Graphs},
  author={Rao, Aditya and Krishnan, Nithin A. and Rivero, Carlos R.},
  booktitle={Proceedings of the ACM Web Conference 2024 (WWW '24)},
  pages={2042--2051},
  year={2024},
  publisher={Association for Computing Machinery},
  address={New York, NY, USA},
  doi={10.1145/3589334.3645506}
}

@inproceedings{egger2024relik,
  title={ReliK: A Reliability Measure for Knowledge Graph Embeddings},
  author={Egger, Matthias K. and Ma, Wei and Mottin, Davide and Karras, Panagiotis and Bordino, Ilaria and Gullo, Francesco and Anagnostopoulos, Aris},
  booktitle={Proceedings of the ACM Web Conference 2024 (WWW '24)},
  year={2024},
  publisher={Association for Computing Machinery},
  address={New York, NY, USA},
  doi={10.1145/3589334.3645430}
}

@inproceedings{sun2019rotate,
  title={RotatE: Knowledge Graph Embedding by Relational Rotation in Complex Space},
  author={Sun, Zhiqing and Deng, Zhiyuan H and Nie, Jian-Yun and Tang, Jian},
  booktitle={ICLR},
  year={2019}
}

@inproceedings{toutanova2015observed,
  title={Observed versus latent features for knowledge base and text inference},
  author={Toutanova, Kristina and Chen, Danqi},
  booktitle={Proceedings of the 3rd workshop on continuous vector space models and their compositionality},
  pages={57--66},
  year={2015}
}

@inproceedings{kadlec2017knowledge,
  title={Knowledge base completion: Baselines strike back},
  author={Kadlec, Rudolf and Bajgar, Ond{\v{r}}ej and Kleindienst, Jan},
  booktitle={Proceedings of the 2nd Workshop on Representation Learning for NLP},
  pages={69--74},
  year={2017}
}

@article{zavadskas2016integrated,
  title={Integrated determination of objective criteria weights in MCDM},
  author={Zavadskas, Edmundas Kazimieras and Podvezko, Valentinas},
  journal={International Journal of Information Technology \& Decision Making},
  volume={15},
  number={02},
  pages={267--283},
  year={2016},
  publisher={World Scientific}
}

@article{wang2019relational,
  title={A relational tucker decomposition for multi-relational link prediction},
  author={Wang, Yanjie and Broscheit, Samuel and Gemulla, Rainer},
  journal={arXiv preprint arXiv:1902.00898},
  year={2019}
}

@inproceedings{gul2025muco,
  title={MuCo-KGC: Multi-Context-Aware Knowledge Graph Completion},
  author={Gul, Haji and Naim, Abdul Ghani and Bhat, Ajaz Ahmad},
  booktitle={Pacific-Asia Conference on Knowledge Discovery and Data Mining},
  pages={3--15},
  year={2025},
  organization={Springer}
}

@inproceedings{sun-etal-2020-evaluation,
    title = "A Re-evaluation of Knowledge Graph Completion Methods",
    author = "Sun, Zhiqing  and
      Vashishth, Shikhar  and
      Sanyal, Soumya  and
      Talukdar, Partha  and
      Yang, Yiming",
    editor = "Jurafsky, Dan  and
      Chai, Joyce  and
      Schluter, Natalie  and
      Tetreault, Joel",
    booktitle = "Proceedings of the 58th Annual Meeting of the Association for Computational Linguistics",
    month = jul,
    year = "2020",
    address = "Online",
    publisher = "Association for Computational Linguistics",
    url = "https://aclanthology.org/2020.acl-main.489/",
    doi = "10.18653/v1/2020.acl-main.489",
    pages = "5516--5522"
}

@InProceedings{pmlr-v124-mohamed20a,
  title = 	 {Popularity Agnostic Evaluation of Knowledge Graph Embeddings},
  author =       {Mohamed, Aisha and Parambath, Shameem and Kaoudi, Zoi and Aboulnaga, Ashraf},
  booktitle = 	 {Proceedings of the 36th Conference on Uncertainty in Artificial Intelligence (UAI)},
  pages = 	 {1059--1068},
  year = 	 {2020},
  editor = 	 {Peters, Jonas and Sontag, David},
  volume = 	 {124},
  series = 	 {Proceedings of Machine Learning Research},
  month = 	 {03--06 Aug},
  publisher =    {PMLR},
  url = 	 {https://proceedings.mlr.press/v124/mohamed20a.html}
}

@inproceedings{habiburahman2024beyond,
  author    = {M. Habiburahman and K. R. S. Wiharja and M. Fikriansyah},
  title     = {Beyond Benchmarks: Assessing Knowledge Graph Completion Methods on Non-Benchmark Employee Data},
  booktitle = {2024 International Conference on Data Science and Its Applications (ICODSA)},
  volume    = {11},
  pages     = {28--33},
  year      = {2024},
  month     = {July},
  doi       = {10.1109/icodsa62899.2024.10652136},
  publisher = {IEEE},
}

@inproceedings{10.1145/3773966.3779401,
author = {Moon, Sooho and Ko, Yunyong},
title = {How Sharp and Bias-Robust is a Model? Dual Evaluation Perspectives on Knowledge Graph Completion},
year = {2026},
isbn = {9798400722929},
publisher = {Association for Computing Machinery},
address = {New York, NY, USA},
url = {https://doi.org/10.1145/3773966.3779401},
doi = {10.1145/3773966.3779401},
booktitle = {Proceedings of the Nineteenth ACM International Conference on Web Search and Data Mining},
pages = {1211–1215},
numpages = {5},
location = {USA},
series = {WSDM '26}
}

@article{pezeshkpour2020revisiting,
  author    = {Pouya Pezeshkpour and Yifan Tian and Sameer Singh},
  title     = {Revisiting Evaluation of Knowledge Base Completion Models},
  journal   = {arXiv preprint arXiv:2002.00967},
  year      = {2020},
  month     = {February},
  url       = {https://arxiv.org/abs/2002.00967}
}

@Article{electronics11233866,
AUTHOR = {Ferrari, Ilaria and Frisoni, Giacomo and Italiani, Paolo and Moro, Gianluca and Sartori, Claudio},
TITLE = {Comprehensive Analysis of Knowledge Graph Embedding Techniques Benchmarked on Link Prediction},
JOURNAL = {Electronics},
VOLUME = {11},
YEAR = {2022},
NUMBER = {23},
ARTICLE-NUMBER = {3866},
URL = {https://www.mdpi.com/2079-9292/11/23/3866},
ISSN = {2079-9292},
DOI = {10.3390/electronics11233866}
}

@article{Cao_Xu_Yang_Cao_Huang_2021, 
title={Dual Quaternion Knowledge Graph Embeddings}, volume={35}, url={https://ojs.aaai.org/index.php/AAAI/article/view/16850}, 
DOI={10.1609/aaai.v35i8.16850}, 
number={8}, 
journal={Proceedings of the AAAI Conference on Artificial Intelligence}, 
author={Cao, Zongsheng and Xu, Qianqian and Yang, Zhiyong and Cao, Xiaochun and Huang, Qingming}, 
year={2021}, 
month={May}, 
pages={6894-6902} 
}

@article{10.1016/j.eswa.2024.125260,
author = {Cao, Zhanyue and Luo, Chao},
title = {Link prediction for knowledge graphs based on extended relational graph attention networks},
year = {2025},
issue_date = {Jan 2025},
publisher = {Pergamon Press, Inc.},
address = {USA},
volume = {259},
number = {C},
issn = {0957-4174},
url = {https://doi.org/10.1016/j.eswa.2024.125260},
doi = {10.1016/j.eswa.2024.125260},
journal = {Expert Syst. Appl.},
month = jan,
numpages = {12}
}

@article{gyani2022mcdm,
  title={MCDM and various prioritization methods in AHP for CSS: A comprehensive review},
  author={Gyani, Jayadev and Ahmed, Ahsan and Haq, Mohd Anul},
  journal={IEEE Access},
  volume={10},
  pages={33492--33511},
  year={2022},
  publisher={IEEE}
}

@article{peng2026causal,
  title={Causal inference-based knowledge graph completion using multi-head attention for fault diagnosis},
  author={Peng, Honglei and Yao, Liya and Yang, Shichen and Zhang, Zizhao and Shi, Jiancheng and Liao, Xu and Xiong, Hui},
  journal={Knowledge-Based Systems},
  pages={116031},
  year={2026},
  publisher={Elsevier}
}

@article{xu2025knowledge,
  title={Knowledge graph completion based on a hierarchical graph attention network with structural information},
  author={Xu, Jie and Zhang, Shumao and Xie, Haodiao and Zhang, Hailing and Miao, Ke and Fu, Qiuru},
  journal={Knowledge-Based Systems},
  pages={115164},
  year={2025},
  publisher={Elsevier}
}

@article{li2025kermit,
  title={Kermit: Knowledge graph completion of enhanced relation modeling with inverse transformation},
  author={Li, Haotian and Yu, Bin and Wei, Yuliang and Wang, Kai and Da Xu, Richard Yi and Wang, Bailing},
  journal={Knowledge-Based Systems},
  volume={324},
  pages={113500},
  year={2025},
  publisher={Elsevier}
}


\appendix

\section{Base Results}
\label{app:first}
This supplementary document provides the complete baseline and ablation results for the multi-criteria decision-making (MCDM) meta-evaluation of knowledge graph completion (KGC) models. Unlike the main manuscript, which reports results averaged across the combined leave-one-model-out (LOMO) and leave-one-group-out (LOGO) removal scheme, this supplement presents three distinct configurations: (1) \textbf{Baseline/Full-Set}: analysis with all models included (no removal), establishing reference rankings and correlation structures; (2) \textbf{LOMO-only}: overall scores computed after sequentially removing individual KGC models; and (3) \textbf{LOGO-only}: overall scores computed after removing entire architectural families (translational/geometric, semantic-matching, neural-CNN, GNN-based). All results cover the tail prediction task $(h,r,?)$ with 20 models across five benchmarks (\textsc{Fb15k}, \textsc{Wn18}, \textsc{Fb15k-237}, \textsc{Wn18rr}, \textsc{Yago3-10}) and the relation prediction task $(h,?,t)$ with 12 models across six benchmarks (adding \textsc{Nell995} and \textsc{Ddb14}), evaluated using seven MCDM aggregators (EDAS, TOPSIS, VIKOR, MOORA, WASPAS, Borda Count, Z-Score) across five reliability dimensions. The following sections report detailed aggregation scores, model rankings, consistency matrices, stability distributions, independence tests, robustness under noise injection, generalizability validation, and Pareto optimality summaries for each configuration.

This section reports results for two KGC prediction tasks: relation prediction $(h,?,t)$ and tail prediction $(h,r,?)$. The analysis begins with the tail prediction task, where seven MCDM aggregators are evaluated across five dimensions: consistency, stability, metric independence, robustness, and generalizability. 


\subsection{Tail Prediction Analysis}
\label{sec:results:tail}
Table~\ref{tab:tail_link_prediction_all} reports the base tail prediction results across five benchmark datasets. A clear pattern emerges: model performance is highly fragmented across metrics and datasets. No single KGC architecture dominates all evaluation criteria, and strong performance on one benchmark does not consistently transfer to others. For example, DualE achieves the best MR on FB15k, but this advantage is not consistent across other datasets. In contrast, models such as ComplEx, RotatE, TuckER, and HakE show varying strengths depending on the metric considered. This confirms that comparisons based on individual metrics can lead to inconsistent conclusions. The role of MCDM is therefore not to replace the original metrics, but to combine these fragmented signals into a more stable and interpretable ranking.

The corresponding aggregation scores and model rankings are reported in Tables~\ref{tab:tail_mcdm-kgc-scores} and~\ref{tab:tail_mcdm-kgc-ranks}. These results show that the choice of aggregator influences the final ranking. While the top models are generally similar, their ordering changes across EDAS, TOPSIS, VIKOR, Borda, Z-Score, MOORA, and WASPAS. The next part of this section evaluates these aggregators using the five criteria to identify the most reliable method for tail prediction.\\

\begin{table*}
\centering
\caption{\small Correlation Matrices for MCDM Methods with Key Metrics, $(h, ?, t)$}
\label{tab:t_correlation-matrices-relation}
\scriptsize
\renewcommand{\arraystretch}{0.99}
\resizebox{\textwidth}{!}{%
\begin{tabular}{lcccccccccccc c}
\toprule
\textbf{MCDM} & \multicolumn{4}{c}{\textbf{Kendall $\tau$ ($\uparrow$)}} & \multicolumn{4}{c}{\textbf{Pearson $r$ ($\uparrow$)}} & \multicolumn{4}{c}{\textbf{Spearman $\rho$ ($\uparrow$)}} & \textbf{Overall} \\
\cmidrule(lr){2-5} \cmidrule(lr){6-9} \cmidrule(lr){10-13} \cmidrule(lr){14-14}
& H1 & H10 & MR & MRR
& H1 & H10 & MR & MRR
& H1 & H10 & MR & MRR & Corr. \\
\midrule
EDAS    & 0.6596 & 0.8179 & 0.3684 & 0.7368 & 0.8442 & 0.9280 & 0.6112 & 0.9058 & 0.8409 & 0.9462 & 0.5729 & 0.9038 & \textbf{0.7613} \\
Z-Score & 0.6702 & 0.8074 & 0.3579 & 0.7474 & 0.8222 & 0.9466 & 0.6247 & 0.9029 & 0.8432 & 0.9379 & 0.5714 & 0.9038 & 0.7611 \\
MOORA   & 0.6069 & 0.8285 & 0.4211 & 0.6842 & 0.7802 & 0.9080 & 0.6650 & 0.8538 & 0.7943 & 0.9507 & 0.6436 & 0.8617 & 0.7498 \\
Borda   & 0.6596 & 0.7546 & 0.3684 & 0.7158 & 0.7436 & 0.8260 & 0.5792 & 0.7995 & 0.8424 & 0.9033 & 0.5789 & 0.8902 & 0.7218 \\
TOPSIS  & 0.5752 & 0.8074 & 0.4526 & 0.6316 & 0.6837 & 0.8404 & 0.6847 & 0.7573 & 0.7507 & 0.9395 & 0.6842 & 0.8120 & 0.7183 \\
WASPAS  & 0.5013 & 0.6596 & 0.5053 & 0.5579 & 0.7314 & 0.8640 & 0.6570 & 0.8012 & 0.6769 & 0.8041 & 0.7083 & 0.7278 & 0.6829 \\
VIKOR   & 0.4802 & 0.6491 & 0.5263 & 0.5158 & 0.3045 & 0.3312 & 0.3004 & 0.3176 & 0.6709 & 0.8011 & 0.7459 & 0.7023 & 0.5288 \\
\bottomrule
\end{tabular}%
}
\end{table*}

\textbf{\textit{Consistency Test:}}
To evaluate the consistency of each MCDM aggregator, three correlation measures, Kendall's $\tau$, Pearson's $r$, and Spearman's $\rho$, are computed between each aggregator’s ranking-assigned values and the base performance metrics (MRR, Hits@1, Hits@10, MR). These correlations are averaged across all benchmarks to assess how well each aggregator preserves the information contained in the original metrics. Higher values indicate stronger agreement and better consistency.

The consistency score for each aggregator is defined as the mean of these three correlations. To facilitate comparison, we normalize the mean scores across all methods. In addition, we quantify stability by computing the standard deviation of the three correlation coefficients, which serves as an error indicator of internal agreement among the correlation measures. 

Figure~\ref{fig:consistency_hmp} \textit{(a)} presents the normalized consistency scores together with their corresponding standard-deviation error bars. Z-Score achieves the highest consistency while exhibiting minimal variation, indicating strong and stable maintenance of the base performance ordering. Most aggregators maintain high consistency. However, VIKOR shows a noticeable drop in mean consistency and an increase in variability, suggesting greater sensitivity to the base metric characteristics. 
\begin{figure}
	\centering
	\includegraphics[scale=.64]{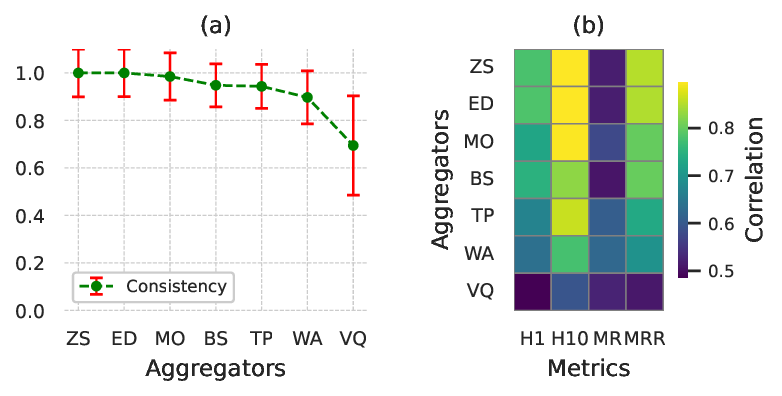}
    \caption{\textit{(a)} Consistency test of MCDM aggregators. \textit{(b)} Correlation between aggregators and base metrics.}
	\label{fig:consistency_hmp}
\end{figure}

Figure~\ref{fig:consistency_hmp} \textit{(b)} shows the correlation between each MCDM aggregator and each base evaluation metric. Each cell represents the average correlation across all benchmarks. A consistent pattern is observed: correlations with MR are generally lower across all aggregators. This occurs because MR is highly sensitive to large rank differences, which reduces its agreement with other metrics. In contrast, MRR and Hits@$k$ show stronger and more consistent alignment with the aggregated results. Overall, the results show that the Z-score is the most consistent and reliable aggregator. 

For tail prediction, EDAS, TOPSIS, and MOORA form a high-consensus cluster ($\tau \geq 0.911$) due to their shared reference-point normalization, while Borda and Z-Score also align closely. VIKOR and WASPAS exhibit lower agreement, reflecting their distinct formulations. 

\textbf{\textit{Stability Test:}}
To assess cross-dataset stability, model rankings are first computed independently for each dataset. For each model, the variance of its ranks across datasets is calculated. Since higher variance indicates greater instability, this value is normalized and inverted so that higher scores indicate more stable rankings. The final stability score is obtained by averaging across all models and tasks for each aggregator.

Figure~\ref{fig:var_std} presents these results using boxplots showing interquartile ranges and whisker spreads. TOPSIS exhibits the highest stability, followed by MOORA and EDAS, indicating stronger cross-dataset consistency, while Borda shows the lowest stability (see Table~\ref{tab:variance_rankings}). These results suggest that scoring-based aggregators tend to produce more stable rankings across heterogeneous benchmarks.\\

\begin{figure}
	\centering
		\includegraphics[scale=.70]{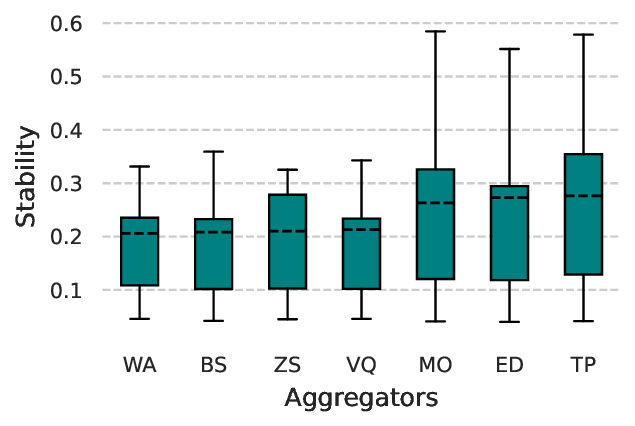}
	\caption{Distribution of stability across different aggregators.}
	\label{fig:var_std}
\end{figure}
\begin{table}
\centering
\footnotesize
\caption{\small Variance of rankings for each knowledge graph embedding model across the five benchmark datasets, computed separately for each MCDM method. Variance is inverted so that higher variance indicates higher ranking stability (more consistent performance) across datasets.}
\label{tab:variance_rankings}
\renewcommand{\arraystretch}{1.05}
\resizebox{\columnwidth}{!}{
\begin{tabular}{lccccccc}
\hline
\textbf{Model} & \textbf{VQ} & \textbf{WA} & \textbf{BC} & \textbf{ZS} & \textbf{ED} & \textbf{MO} & \textbf{TP} \\
\hline
ConvR     & 3.100 & 3.160 & 2.400 & 2.800 & 2.480 & 2.880 & 2.280 \\
RotatE    & 3.380 & 3.440 & 4.200 & 4.500 & 3.500 & 3.660 & 3.700 \\
CrossE    & 5.960 & 7.060 & 6.240 & 6.460 & 2.280 & 2.300 & 2.140 \\
TuckER    & 9.240 & 6.580 & 7.240 & 6.460 & 3.880 & 4.360 & 5.200 \\
ConvE     & 9.340 & 6.880 & 8.640 & 7.420 & 5.360 & 5.320 & 5.640 \\
ComplEx   & 9.060 & 7.920 & 9.340 & 11.020 & 8.240 & 7.120 & 6.180 \\
DistMult  & 13.740 & 12.280 & 11.680 & 12.940 & 9.360 & 10.040 & 9.100 \\
SimplE    & 7.920 & 9.840 & 9.460 & 9.840 & 13.980 & 14.860 & 14.040 \\
NodePiece & 11.940 & 15.740 & 13.680 & 13.600 & 10.880 & 10.880 & 11.200 \\
ANALOGY   & 10.380 & 13.960 & 14.300 & 14.300 & 12.960 & 13.740 & 14.100 \\
HolE      & 18.700 & 17.820 & 17.340 & 15.740 & 12.420 & 13.180 & 12.540 \\
RSN       & 17.280 & 15.580 & 14.180 & 14.380 & 18.400 & 16.680 & 16.160 \\
TransE    & 15.660 & 15.000 & 15.400 & 15.500 & 17.980 & 17.700 & 17.100 \\
HakE      & 21.460 & 17.000 & 16.980 & 17.600 & 19.700 & 17.920 & 16.200 \\
STransE   & 20.880 & 21.700 & 22.640 & 20.940 & 15.480 & 15.100 & 14.240 \\
TorusE    & 20.440 & 20.560 & 20.460 & 19.320 & 15.500 & 18.100 & 19.260 \\
R-GCN     & 23.340 & 26.200 & 25.000 & 25.420 & 17.540 & 16.620 & 16.440 \\
DualE     & 25.680 & 22.820 & 23.620 & 23.660 & 29.900 & 27.880 & 26.880 \\
ConvKB    & 29.180 & 34.060 & 31.020 & 31.500 & 31.720 & 32.720 & 32.240 \\
CompGCN   & 50.880 & 47.420 & 46.720 & 46.740 & 53.340 & 52.140 & 51.540 \\
\hline
\textbf{Mean Variance} & 0.9042 & 0.9113 & 0.9240 & 0.9252 & 0.9714 & 0.9768 & \textbf{1.0000} \\
\hline
\end{tabular}
}
\end{table}

\textbf{\textit{Independency Test:}}
To assess aggregator robustness under partial information loss, we employ a leave-one-metric-out analysis test. Each base metric is systematically omitted, and the resulting aggregator rankings are compared against the baseline computed using the complete metric set. This evaluates whether each method maintains ranking stability when evaluation information is incomplete.

For each aggregator, we compute three correlation coefficients between the reduced-metric and full-metric outputs. The overall independency score is the mean of these correlations. Scores are normalized by the maximum value across all methods for easier interpretation. To quantify stability, the analysis also reports the standard deviation across the three correlation measures as error bars, reflecting the internal consistency of agreement under metric removal.

Figure~\ref{fig:robustness} presents the normalized correlation with error bars representing $\pm 1$ standard deviation, sorted from highest to lowest correlation. WASPAS and VIKOR achieve the highest independency, indicating strong resistance to the removal of any single metric. The Z-score achieves near-perfect normalized correlation with the smallest variation. In contrast, TOPSIS and EDAS exhibit substantially lower scores and higher variability, revealing greater sensitivity to the incomplete evaluation metric. These results position WA and VQ, followed by Z-score, as the most robust aggregators for scenarios involving incomplete or fluctuating evaluation 

Table~\ref{tab:robustness-merged} reports the detailed metric-removal results. Kendall's~$\tau$ and Spearman's~$\rho$ measure global ranking preservation, average absolute rank displacement ($|\Delta \text{Rank}|$) measures local positional changes, and Top-5 Jaccard similarity evaluates whether the highest-performing models remain stable. This further verifies that Z-score and Borda provide stronger overall balance after normalization.\\

\begin{figure}
	\centering
	\includegraphics[scale=.75]{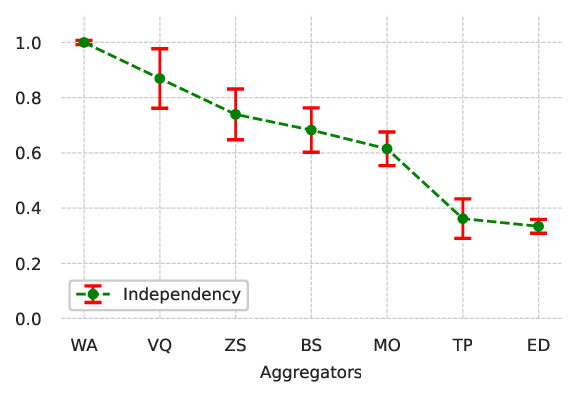}
    \caption{Independency analysis of MCDM aggregators under metric removal.}
	\label{fig:robustness}
\end{figure}

\begin{table}[h!]
\caption{Detailed independency analysis of MCDM aggregation methods under metric removal. For Kendall's~$\tau$, Spearmans~$\rho$, and Top-5 Jaccard, higher is better; for $|\Delta \text{Rank}|$, lower is better.}
\label{tab:robustness-merged}
\centering
\scriptsize
\setlength{\tabcolsep}{2.5pt}
\renewcommand{\arraystretch}{1.15}
\resizebox{\columnwidth}{!}{%
\begin{tabular}{llccccccc}
\toprule
Removed & Metric & Borda & Z-score & TOPSIS & VIKOR & EDAS & MOORA & WASPAS \\
\midrule
\multirow{4}{*}{Kendall $\tau$}
& Without MR  & 0.495 & 0.579 & 0.227 & 0.663 & 0.126 & 0.505 & 1.000 \\
& Without MRR & 0.635 & 0.674 & 0.164 & 0.684 & 0.189 & 0.474 & 1.000 \\
& Without H@1 & 0.638 & 0.579 & 0.575 & 0.853 & 0.600 & 0.589 & 1.000 \\
& Without H@10& 0.505 & 0.632 & 0.164 & 0.716 & 0.274 & 0.547 & 1.000 \\
\midrule
\multirow{4}{*}{Spearman $\rho$}
& Without MR  & 0.644 & 0.723 & 0.302 & 0.841 & 0.093 & 0.644 & 1.000 \\
& Without MRR & 0.780 & 0.791 & 0.180 & 0.836 & 0.266 & 0.586 & 1.000 \\
& Without H@1 & 0.776 & 0.735 & 0.743 & 0.941 & 0.698 & 0.708 & 1.000 \\
& Without H@10& 0.657 & 0.761 & 0.147 & 0.880 & 0.362 & 0.588 & 1.000 \\
\midrule
\multirow{4}{*}{$|\Delta ~ Rank|$}
& Without MR  & 1.50 & 1.60 & 4.85 & 3.50 & 7.60 & 2.80 & 2.90 \\
& Without MRR & 0.80 & 0.70 & 0.65 & 0.50 & 0.00 & 1.60 & 0.70 \\
& Without H@1 & 0.60 & 0.50 & 1.25 & 1.40 & 0.00 & 1.70 & 1.50 \\
& Without H@10& 1.40 & 1.20 & 0.55 & 0.90 & 0.00 & 1.10 & 0.80 \\
\midrule
\multirow{4}{*}{Top-5 Jaccard}
& Without MR  & 0.667 & 0.667 & 0.111 & 0.667 & 0.000 & 0.429 & 0.429 \\
& Without MRR & 1.000 & 1.000 & 1.000 & 1.000 & 1.000 & 0.667 & 0.667 \\
& Without H@1 & 1.000 & 1.000 & 1.000 & 0.667 & 1.000 & 0.667 & 0.667 \\
& Without H@10& 1.000 & 0.667 & 1.000 & 1.000 & 1.000 & 1.000 & 1.000 \\
\bottomrule
\end{tabular}}
\end{table}
\textbf{\textit{Robustness Test:}}
This test evaluates aggregator stability under measurement uncertainty by injecting noise into metric scores. Each noise level (5\%, 10\%, 20\%, and 30\%), repeating 1,000 times per level, to compute average rank displacement and correlation with baseline rankings.

\begin{figure}
	\centering
		\includegraphics[scale=.80]{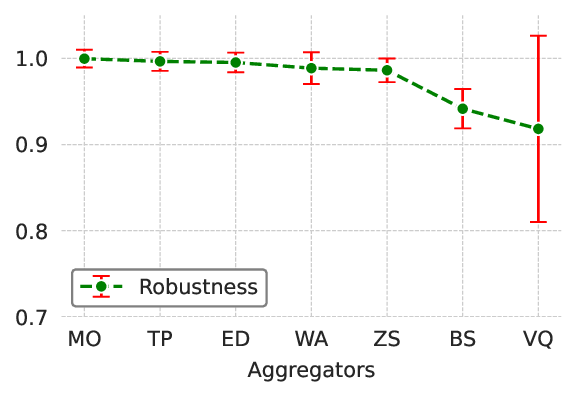}
	\caption{Robustness analysis under noise injection.}
	\label{fig:tnoise_5_10_20}
\end{figure}

As shown in Figure~\ref{fig:tnoise_5_10_20}, MOORA, TOPSIS, EDAS, WASPAS, and Z-Score maintain near-perfect normalized correlation with minimal variance, demonstrating strong robustness to metric noise injection across all levels. VIKOR shows the lowest correlation and the widest error bounds, indicating higher sensitivity to noise injection. The tight error margins for top-performing methods suggest that rank- and distance-based aggregators preserve ranking structure reliably even under substantial metric variation, making them suitable for deployment in noisy or uncertain evaluation environments.

For clarity, Figure~\ref{fig:noise_5_10_20} presents detailed results across noise levels, with aggregators sorted by performance at 30\% noise. The \textit{left} panel shows the mean absolute rank displacement ($|\Delta\mathrm{Rank}|$), which is inverted so that higher values indicate greater stability. VIKOR exhibits the highest sensitivity due to its distance-to-ideal formulation, while MOORA and TOPSIS demonstrate superior robustness, maintaining minimal displacement even at 30\% noise, as their normalization procedures inherently dampen noise effects. The \textit{right} panel shows Kendall's correlation with baseline rankings, where all aggregators maintain a strong correlation up to 20\% noise. MOORA and TOPSIS preserve the strongest correlation at 30\% noise, whereas VIKOR shows the weakest. Overall, MOORA emerges as the most robust aggregator for uncertain KGC evaluations, while VIKOR's compromise programming proves unsuitable for high-noise environments.\\
\begin{figure}
	\centering
		\includegraphics[scale=.60]{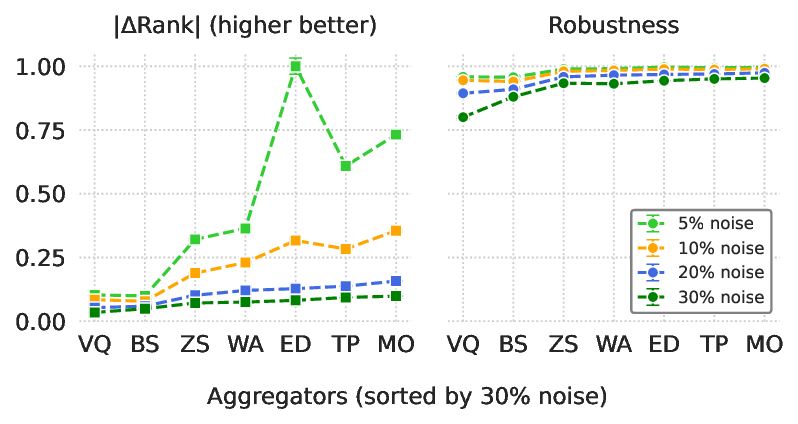}
	\caption{Robustness assessment with noise injection (1,000 iterations per level). \textit{Left}: Mean absolute rank displacement ($|\Delta\mathrm{Rank}|$, higher = more robust). \textit{Right}: average correlation with baseline ranking (higher = more robust).}
	\label{fig:noise_5_10_20}
\end{figure}

\textbf{\textit{Generalizability Test:}}
This test evaluates cross-dataset transferability using leave-one-dataset-out validation. For each eliminated benchmark, the aggregators are assessed according to how well rankings derived from the remaining datasets transfer to the unseen dataset. The aim is to examine whether each MCDM aggregation method produces rankings that remain reliable across changing evaluation contexts.

For each validation step, one dataset is treated as unseen, while the remaining datasets are used to predict the reference ranking for each aggregator. This transferred ranking is then compared with the ranking obtained directly on the eliminated dataset. The relationships are measured using correlation, which is averaged to obtain an overall generalizability and normalized with $value/Max$. 

Figure~\ref{fig:generalizability} presents the generalizability with error bars representing $\pm 1$ standard deviation, sorted from highest to lowest score. The results reveal that Z-scores achieve the highest generalizability with minimal variation. In contrast, TOPSIS and VIKOR yield the lowest scores, with the highest variation. This indicates that reference-distance-based methods are particularly sensitive to changes in the evaluation context. These results suggest that ZS-based methods offer the most reliable generalization behavior. 

Table~\ref{tab:final-generalizability} reports the detailed leave-one-dataset-out results. In each fold, Top-1 accuracy is computed by checking whether the model ranked first by the transferred ranking is also the best model on the eliminated dataset. Similarly, Top-3 accuracy checks whether the best model on the eliminated dataset appears among the top three models in the transferred ranking. The table also reports correlation to measure full-ranking agreement, while the last column provides the corresponding 95\% confidence interval. The overall findings of this test show that the Z-score and EDAS are more reliable when rankings learnt from observed datasets are transferred to unseen benchmarks.

\begin{figure}
	\centering
		\includegraphics[scale=.73]{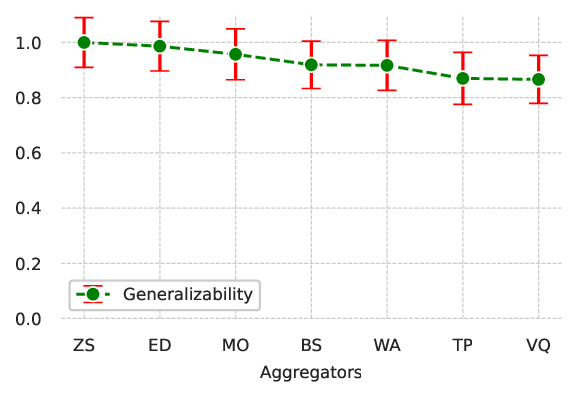}
	\caption{Generalizability analysis of MCDM aggregators under leave-one-dataset-out validation.}
	\label{fig:generalizability}
\end{figure}

\begin{table}[t]
\caption{Generalizability results for tail prediction using leave-one-dataset-out validation. Higher values indicate better cross-dataset transferability.}
\label{tab:final-generalizability}
\centering
\scriptsize
\setlength{\tabcolsep}{3pt}
\renewcommand{\arraystretch}{1.1}
\resizebox{\columnwidth}{!}{%
\begin{tabular}{lcccccc}
\hline
\textbf{Aggregator} & \textbf{Top-1} & \textbf{Top-3} & $\boldsymbol{\tau}$ & \textbf{Pearson} & \textbf{Spearman} & \textbf{95\% CI} \\
\hline
Mean Z-Score & 0.100 & 0.250 & 0.256 & 0.340 & 0.340 & [0.128, 0.496] \\
EDAS         & 0.050 & 0.150 & 0.249 & 0.337 & 0.337 & [0.125, 0.491] \\
MOORA        & 0.100 & 0.250 & 0.248 & 0.323 & 0.323 & [0.110, 0.486] \\
Borda Score  & 0.050 & 0.200 & 0.226 & 0.317 & 0.317 & [0.111, 0.462] \\
WASPAS       & 0.050 & 0.250 & 0.238 & 0.310 & 0.310 & [0.101, 0.471] \\
TOPSIS       & 0.050 & 0.250 & 0.222 & 0.296 & 0.296 & [0.079, 0.463] \\
VIKOR        & 0.050 & 0.100 & 0.220 & 0.295 & 0.295 & [0.093, 0.447] \\
\hline
\end{tabular}
}
\end{table}
\subsubsection{Pareto Analysis}
\label{subsec:pareto}
To identify the most suitable MCDM aggregator for KGC evaluation, we conduct a Pareto optimality analysis across all five test dimensions. Since no single aggregator dominates all criteria, we employ three complementary visualization and statistical techniques to determine which methods offer the best trade-offs.

\begin{figure}
	\centering
		\includegraphics[scale=.69]{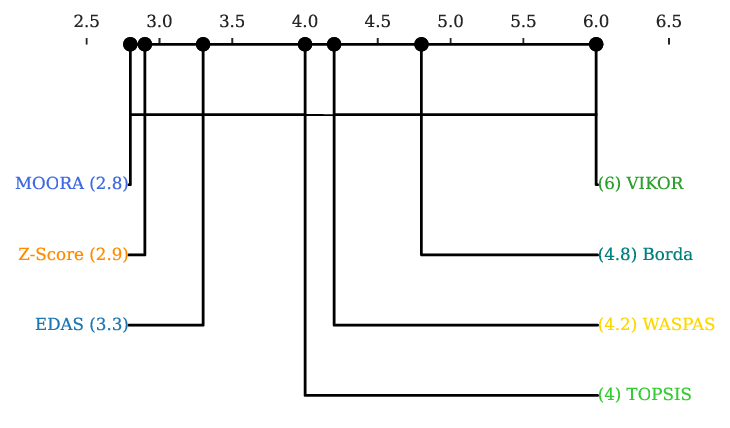}
	\caption{Critical difference (CD) diagram showing average ranks of MCDM aggregators across five evaluation dimensions. Methods connected by horizontal lines are not significantly different (Nemenyi post-hoc test, $\alpha=0.05$). Z-Score and MOORA achieve the top ranks performance.}
	\label{fig:critical_difference}
\end{figure}
Figure~\ref{fig:critical_difference} presents a critical difference (CD) diagram based on Friedman ranking tests. The Friedman test determines whether observed differences are statistically significant ($p < 0.05$), followed by Nemenyi post-hoc analysis to compute the critical difference threshold. Z-Score and MOORA achieve the best average ranks, followed by EDAS, while VIKOR performs the worst.

Figure~\ref{fig:radar_tradeoff} shows normalized performance scores (0–1 scale, higher is better) across all five tests. The radar plot further verified that no aggregator dominates all dimensions, but Z-Score performs most consistently, with consistently high scores across all criteria.
\begin{figure}
	\centering
		\includegraphics[scale=.60]{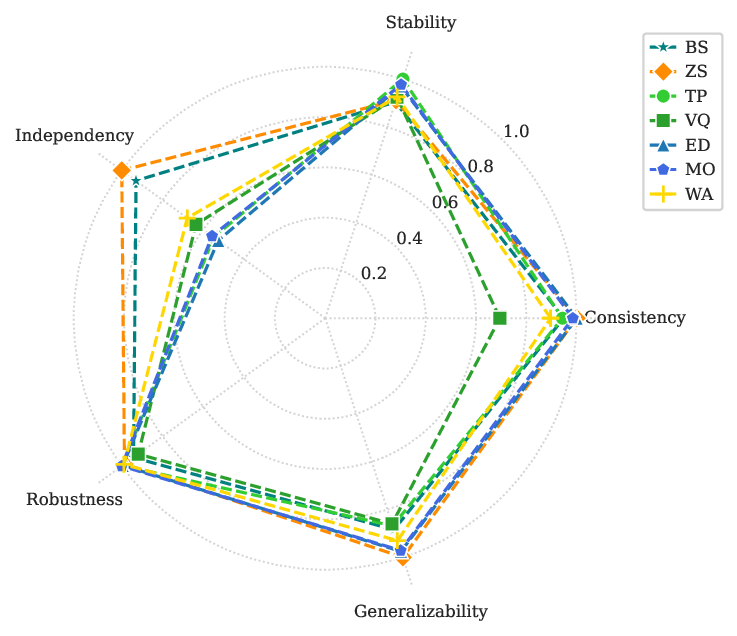}
	\caption{A radar plot showing normalized performance (0–1 scale, higher is better) across five evaluation dimensions. The Z-score demonstrates the most balanced performance with strong scores across consistency, stability, and generalisability.}
	\label{fig:radar_tradeoff}
\end{figure}

Figure~\ref{fig:pareto} presents the overall performance score for each aggregator across all five tests: $\text{Overall}_i = \frac{1}{5}\sum_{j=1}^{5} s_{ij}$. Z-Score achieves the highest overall performance, followed by Borda, MOORA, WASPAS, and EDAS, while VIKOR performs the worst.
\begin{figure}
	\centering
		\includegraphics[scale=.60]{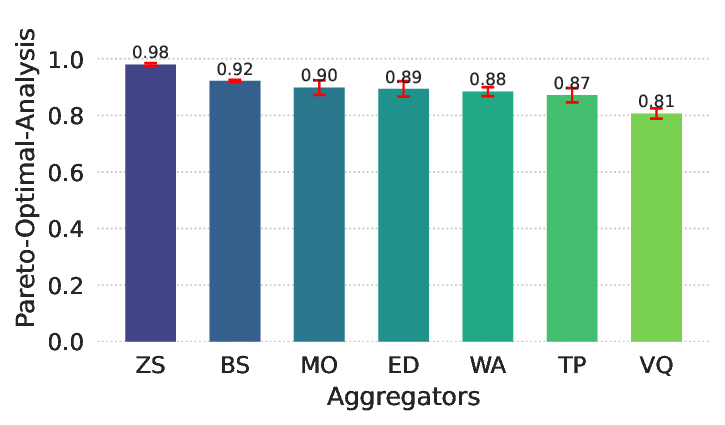}
	\caption{Overall performance scores for each MCDM aggregator. Z-Score achieves the highest score, indicating balanced performance across all tests.}
	\label{fig:pareto}
\end{figure}

\subsubsection{KGC method removal method analysis}
\begin{figure}
	\centering
		\includegraphics[scale=.60]{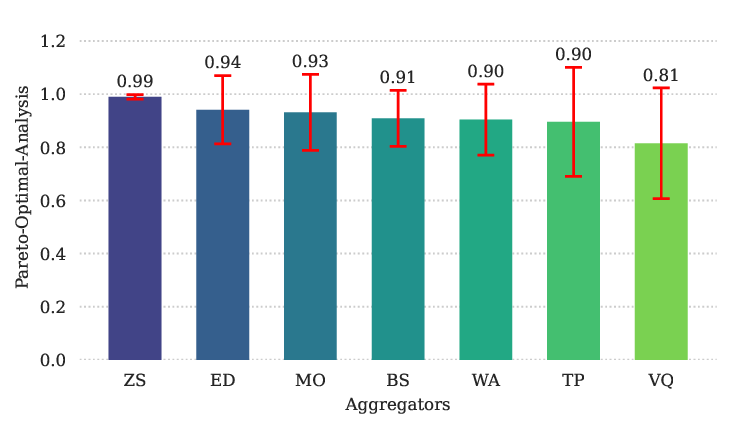}
	\caption{One KGC method removal overall performance.}
	\label{fig:one_pareto}
\end{figure}

Additionally, the overall score is computed after one KGC method removal analysis, followed by all five dimensional tests, and the output is reported in Figure \ref{fig:one_pareto}. 

\begin{figure}
	\centering
		\includegraphics[scale=.60]{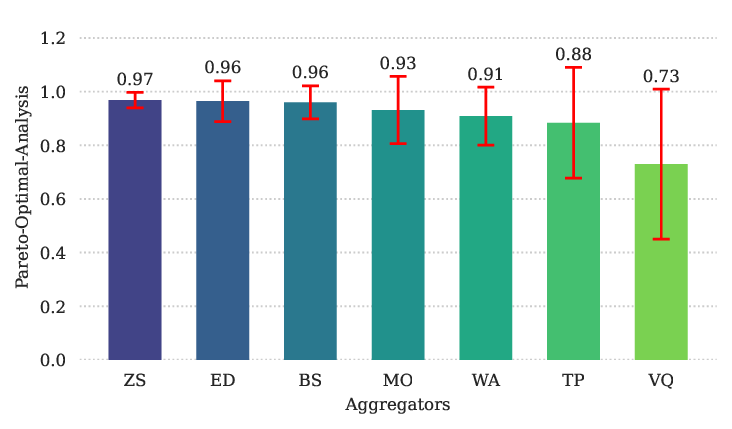}
	\caption{Group KGC method removal overall performance.}
	\label{fig:group_pareto}
\end{figure}

Similarly, the group KGC method removal analysis was performed under each of the five dimensional tests, and the overall score is shown in Figure \ref{fig:group_pareto}. 

Based on the Pareto optimality and KGC method removal analysis, \textbf{Z-Score} is selected as the optimal aggregator for holistic KGC evaluation. Table~\ref{tab:zs-top3} presents the top three KGC methods for the tail prediction task as ranked exclusively by Z-Score.
\begin{table}[t]
\centering
\caption{\small Top-3 KGC models ranked by the Z-Score aggregator (1 = best).}
\label{tab:zs-top3}
\footnotesize
\setlength{\tabcolsep}{6pt}
\begin{tabular*}{\columnwidth}{@{\extracolsep{\fill}}lccc}
\toprule
& \multicolumn{3}{c}{\textbf{Z-Score}} \\
\cmidrule(lr){2-4}
\textbf{Model} & \textbf{DualE} & \textbf{HakE} & \textbf{ComplEx} \\
\midrule
\textbf{Rank} & 1 & 2 & 3 \\
\bottomrule
\end{tabular*}
\end{table}
DualE achieves the highest Z-Score aggregate ranking (rank~1), indicating it provides the most balanced performance across all metrics (MR, MRR, Hits@1, and Hits@10) on the tail prediction task.

\subsection{Relation Prediction}
\label{sec:results:relation}
We apply the same meta-evaluation pipeline to the relation prediction task $(h,?,t)$. This task uses MRR, Hits@1, and Hits@3 across six benchmark datasets: FB15K, FB15K-237, WN18, WN18RR, NELL995, and DDB14. Base results are reported in Table~\ref{tab:transductive_percentage}, while MCDM aggregation scores and induced rankings appear in Tables~\ref{tab:mcdm-kgc-r_scores} and~\ref{tab:mcdm-kgc-r_ranks}.

\textbf{\textit{Consistency Test:}}
To compute consistency for relation prediction, we follow the same procedure described in the tail prediction consistency test. Figure~\ref{fig:r_consistency_hmp} presents the consistency analysis for relation prediction. 

Figure~\ref{fig:r_consistency_hmp}~\textit{(a)} shows normalized consistency scores with standard-deviation error bars: all aggregators maintain near-perfect consistency ($\approx 1.0$) with minimal variation, indicating strong preservation of base metric rankings. MOORA, Z-Score, EDAS, TOPSIS, WASPAS, and Borda Score exhibit virtually identical performance, while VIKOR shows a marginal decline with slightly elevated variance. 
Figure~\ref{fig:r_consistency_hmp}~\textit{(b)} depicts pairwise agreement between aggregators and base metrics, revealing a strong correlation between Z-Score and MOORA, whereas VIKOR produces the most distinct ranking behavior. Overall, these results confirm that all MCDM aggregators reliably preserve the information in the original evaluation metrics, with Z-Score and MOORA demonstrating the highest consistency.
\begin{figure}
    \centering
    \includegraphics[scale=.60]{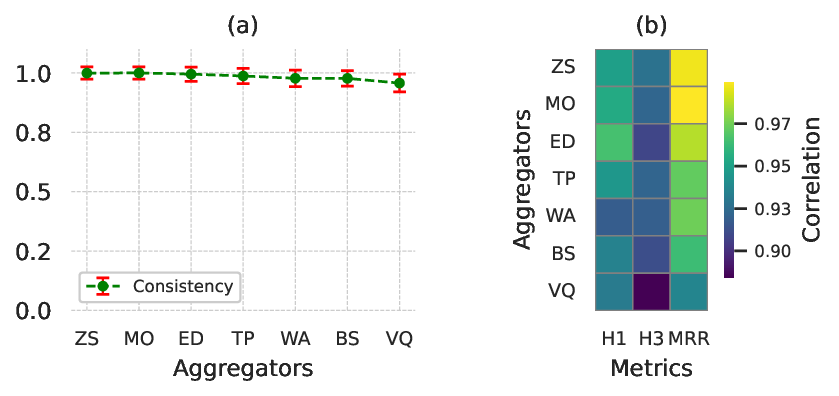}
    \caption{Relation consistency \textit{(a)} Consistency test of MCDM aggregators. \textit{(b)} Correlation between aggregators and base metrics.}
    \label{fig:r_consistency_hmp}
\end{figure}

\begin{table*}[h!]
\centering
\caption{\small Correlation Matrices for MCDM Methods with Key Metrics, (H1, H3, MRR)}
\label{tab:correlation-matrices-relation}
\scriptsize
\tiny
\renewcommand{\arraystretch}{0.99}
\resizebox{\textwidth}{!}{%
\begin{tabular}{l ccc ccc ccc c}
\toprule
\textbf{MCDM} & \multicolumn{3}{c}{\textbf{Kendall $\tau$ ($\uparrow$)}} & \multicolumn{3}{c}{\textbf{Pearson $r$ ($\uparrow$)}} & \multicolumn{3}{c}{\textbf{Spearman $\rho$ ($\uparrow$)}} & \textbf{Overall} \\
\cmidrule(lr){2-4} \cmidrule(lr){5-7} \cmidrule(lr){8-10} \cmidrule(lr){11-11}
& H1 & H3 & MRR
& H1 & H3 & MRR
& H1 & H3 & MRR & Corr. \\
\midrule
MOORA     & 0.9091 & 0.8788 & 1.0000 & 0.9814 & 0.9212 & 0.9989 & 0.9720 & 0.9650 & 1.0000 & 0.9585 \\
Z-Score   & 0.9091 & 0.8788 & 1.0000 & 0.9652 & 0.9394 & 0.9920 & 0.9720 & 0.9650 & 1.0000 & 0.9579 \\
EDAS      & 0.9394 & 0.8485 & 0.9697 & 0.9783 & 0.9259 & 0.9973 & 0.9790 & 0.9510 & 0.9930 & 0.9536 \\
TOPSIS    & 0.9091 & 0.8788 & 0.9394 & 0.9687 & 0.9268 & 0.9906 & 0.9580 & 0.9580 & 0.9860 & 0.9462 \\
WASPAS    & 0.8788 & 0.8485 & 0.9697 & 0.9172 & 0.9563 & 0.9592 & 0.9580 & 0.9510 & 0.9930 & 0.9369 \\
Borda     & 0.9313 & 0.8397 & 0.9619 & 0.8924 & 0.9459 & 0.9372 & 0.9807 & 0.9492 & 0.9912 & 0.9366 \\
VIKOR     & 0.9394 & 0.7879 & 0.9091 & 0.8772 & 0.9556 & 0.9283 & 0.9790 & 0.9091 & 0.9720 & 0.9175 \\
\bottomrule
\end{tabular}%
}
\end{table*}

Figure~\ref{fig:r_consistency_hmp}(a) and Table~\ref{tab:correlation-matrices-relation} present the consistency results. All aggregators achieve near-perfect normalized consistency scores ($\approx$ 1.0) with very low variation. MOORA attains the highest overall correlation, followed closely by Z-Score and EDAS. VIKOR shows a marginal drop. This indicates stronger and more uniform preservation of base metric rankings compared to tail prediction. 

For relation prediction, MOORA and Z-Score are perfectly correlated and align closely with EDAS, TOPSIS, and WASPAS, indicating strong consensus among top aggregators. VIKOR remains the most divergent, consistent with its compromise-oriented ranking strategy. Overall, these results confirm that high-performing aggregators preserve rather than contradict the underlying metric signals.

\textbf{\textit{Stability Test:}}
For relation prediction, the stability test follows the same cross-dataset variance procedure described in the tail prediction stability analysis. 

\begin{figure}
	\centering
	\includegraphics[scale=.70]{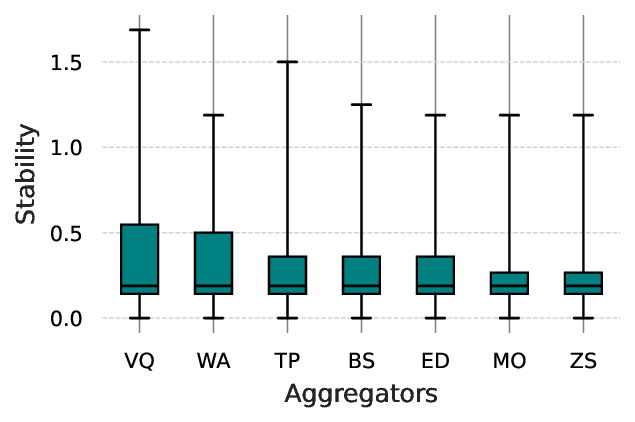}
	\caption{Distribution of ranking variance across relation prediction datasets for different aggregators.}
	\label{fig:r_var_std}
\end{figure}
Figure~\ref{fig:r_var_std} shows the distribution of ranking variance for each aggregator across the relation prediction datasets. A compact distribution indicates that the aggregator produces more stable rankings across benchmarks, whereas a wider box or longer whiskers indicate greater dataset-dependent fluctuation. Z-Score and MOORA exhibit the most compact distributions, suggesting that their rankings remain more consistent across benchmark changes. In contrast, VIKOR and WASPAS show wider distributions and longer upper ranges. This suggests that their rankings are more sensitive to dataset-level changes.  The main finding of stability shows that Z-Score and MOORA provide the most stable rankings for relation prediction.

\textbf{\textit{Independency Test:}}
To assess aggregator robustness to partial information loss in relation prediction, we employ a leave-one-metric-out analysis following the same procedure as for tail prediction. Each base metric is systematically omitted, and the resulting aggregator rankings are compared against the baseline computed using the complete metric set.

Figure~\ref{fig:r_robustness} presents the normalized independency scores with error bars representing $\pm 1$ standard deviation, sorted from highest to lowest correlation. VIKOR, WASPAS, Borda Score, and Z-Score achieve the highest independency with minimal variation, indicating strong resistance to the removal of any single metric. MOORA, EDAS, and TOPSIS follow closely with near-perfect scores, though TOPSIS shows slightly higher variability. Overall, the independency test demonstrates that relation prediction rankings remain highly stable under metric removal, with VIKOR, Borda, and Z-Score providing the strongest metric independence.
\begin{figure}
	\centering
	\includegraphics[scale=.80]{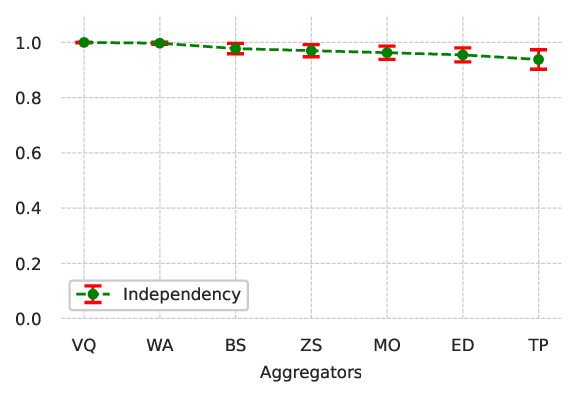}
	\caption{Metric independence of MCDM aggregators under leave-one-metric-family-out evaluation for relation prediction.}
	\label{fig:r_robustness}
\end{figure}

\textbf{\textit{Robustness Test:}}
The robustness test for relation prediction follows the same noise-injection procedure described in the tail prediction analysis. 

Figure~\ref{fig:r_robustness} shows the robustness scores with error bars representing $\pm 1$ standard deviation, sorted from highest to lowest. TOPSIS and MOORA achieve near-perfect robustness with minimal variation, followed closely by EDAS and Z-Score. VIKOR shows the lowest robustness and the widest error bounds, indicating substantially higher sensitivity to noise injection. The tight error margins for top-performing methods confirm that distance- and score-based aggregators reliably preserve the ranking structure, even under substantial metric variation, for relation prediction.
\begin{figure}
	\centering
		\includegraphics[scale=.75]{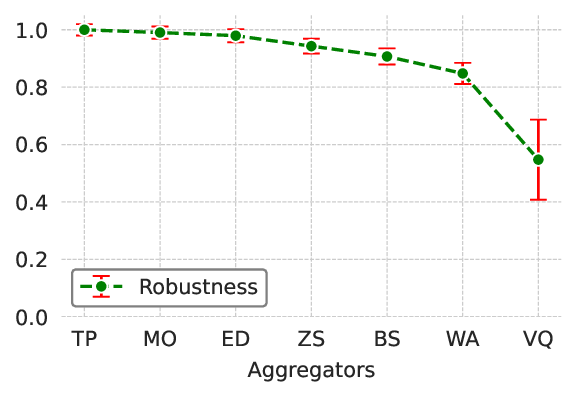}
	\caption{Robustness analysis under noise injection. }
	\label{fig:error_r_noise_5_10_20}
\end{figure}

For more detail, Figure~\ref{fig:r_noise_5_10_20} presents two views: the left panel shows ranks changing in an inverted order, with higher being better, and the right panel shows average correlation with the baseline ranking. TOPSIS and MOORA are the most robust aggregators, maintaining high scores across both measures as noise increases due to their ratio-based scoring, which preserves relative separation among models. In contrast, VIKOR is the most sensitive, as its compromise-ranking mechanism amplifies small metric variations. Overall, TOPSIS and MOORA preserve ranking consistency more reliably under noise, whereas VIKOR is more affected.
\begin{figure}
	\centering
		\includegraphics[scale=.58]{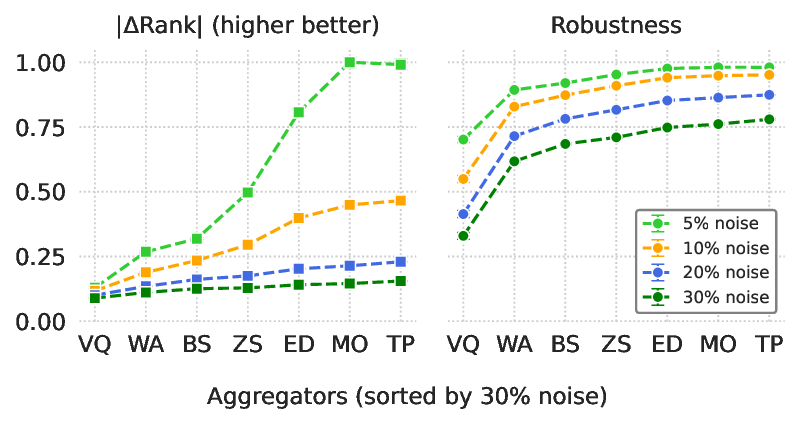}
	\caption{Robustness assessment with noise injection (1,000 iterations per level). \textit{Left}: Mean absolute rank displacement ($|\Delta\mathrm{Rank}|$, higher = more robust). \textit{Right}: average correlation with baseline ranking (higher = more robust).}
	\label{fig:r_noise_5_10_20}
\end{figure}

\textbf{\textit{Generalizability Test:}}
The relation prediction generalizability test follows the same leave-one-dataset-out validation procedure as tail prediction, examining whether an aggregator can predict rankings for an unseen benchmark using other existing results. 

Figure~\ref{fig:r_generalizability} presents the cross-dataset generalizability of MCDM aggregators for relation prediction. Z-Score, TOPSIS, MOORA, Borda, EDAS, and WASPAS achieve near-perfect normalized scores ($\approx 1.0$) with minimal variance, demonstrating consistent ranking transferability across unseen benchmarks. VIKOR exhibits a sharp decline with notably larger error bounds, indicating poor generalization and high sensitivity to dataset distribution shifts. 

Table~\ref{tab:r_final-generalizability-final} reports the detailed results, where all aggregators achieve perfect Top-1 and Top-3 accuracy, with 95\% confidence intervals. Overall, distance- and score-based aggregators reliably preserve model rankings across unseen benchmarks, whereas VIKOR's compromise-based method struggles with cross-dataset transfer.
\begin{figure}
	\centering
	\includegraphics[scale=.74]{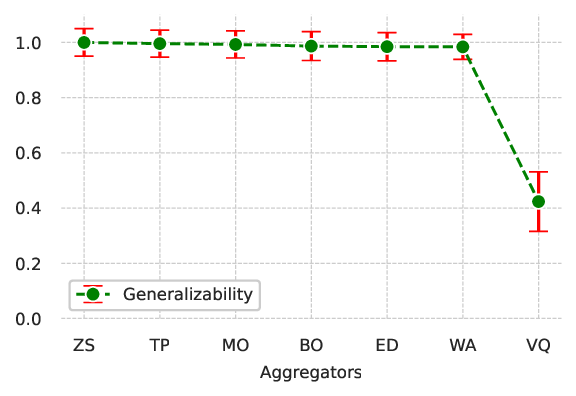}
	\caption{Generalizability analysis of MCDM aggregators under leave-one-dataset-out validation.}
	\label{fig:r_generalizability}
\end{figure}

\begin{table}[t]
\caption{Generalizability results for relation prediction using leave-one-dataset-out validation. Higher values indicate better cross-dataset transferability.}
\label{tab:r_final-generalizability-final}
\centering
\scriptsize
\setlength{\tabcolsep}{3pt}
\renewcommand{\arraystretch}{1.1}
\resizebox{\columnwidth}{!}{%
\begin{tabular}{lcccccc}
\hline
\textbf{Aggregator} & \textbf{Top-1} & \textbf{Top-3} & $\boldsymbol{\tau}$ & \textbf{Pearson} & \textbf{Spearman} & \textbf{95\% CI} \\
\hline
Mean Z-Score & 1.000 & 1.000 & 0.510 & 0.634 & 0.634 & [0.453, 0.568] \\
TOPSIS       & 1.000 & 1.000 & 0.517 & 0.635 & 0.635 & [0.458, 0.576] \\
WASPAS       & 1.000 & 1.000 & 0.498 & 0.630 & 0.630 & [0.446, 0.551] \\
Borda Score  & 1.000 & 1.000 & 0.503 & 0.630 & 0.630 & [0.443, 0.564] \\
MOORA        & 1.000 & 1.000 & 0.510 & 0.632 & 0.632 & [0.452, 0.568] \\
EDAS         & 1.000 & 1.000 & 0.505 & 0.627 & 0.627 & [0.445, 0.565] \\
VIKOR        & 1.000 & 1.000 & 0.210 & 0.273 & 0.273 & [0.112, 0.309] \\
\hline
\end{tabular}
}
\end{table}

\subsubsection{Pareto Analysis}
\label{sec:results:relation:pareto}
Combining the five evaluation dimensions (Figure~\ref{fig:r_pareto}), Z-Score again achieves the best overall trade-off, followed by MOORA, EDAS, and TOPSIS. VIKOR remains the weakest.
\begin{figure}
	\centering
	\includegraphics[scale=.68]{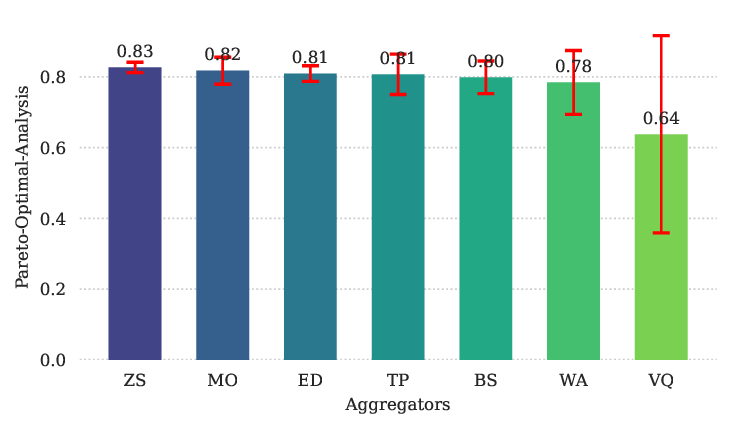}
	\caption{Pareto-based overall performance of MCDM aggregators for relation prediction.}
	\label{fig:r_pareto}
\end{figure}

Figure~\ref{fig:r_pareto} summarizes the overall trade-off among the aggregators after combining the normalized scores from the five evaluation dimensions. Z-Score achieves the strongest overall Pareto performance, followed closely by MOORA, EDAS, and TOPSIS. This indicates that these methods maintain strong performance across multiple criteria rather than performing well in only one test. Borda Score and WASPAS remain competitive but show slightly weaker overall balance. In contrast, VIKOR shows the weakest overall Pareto performance, mainly because its strength in metric independence is not enough to make up for its weaker performance in robustness and generalizability. Overall, the Pareto analysis identifies Z-Score as the most suitable aggregator; it remains consistently strong across the full evaluation framework, which makes it the most balanced and reliable choice for unified relation prediction ranking.

\begin{table}[t]
\centering
\caption{\small Top-3 KGC models ranked by the Z-Score aggregator for relation prediction (1 = best).}
\label{tab:zs-top3-relation}
\footnotesize
\setlength{\tabcolsep}{6pt}
\begin{tabular*}{\columnwidth}{@{\extracolsep{\fill}}lccc}
\toprule
& \multicolumn{3}{c}{\textbf{Z-Score}} \\
\cmidrule(lr){2-4}
\textbf{Model} & \textbf{FMS} & \textbf{PathCon} & \textbf{Con} \\
\midrule
\textbf{Rank} & 1 & 2 & 3 \\
\bottomrule
\end{tabular*}
\end{table}
Using the Pareto-selected Z-Score aggregator, Table~\ref{tab:zs-top3-relation} identifies FMS as the top-performing KGC model for relation prediction, followed by PathCon.

\subsection{Statistical Comparison (Z-Score vs MOORA)}
\begin{figure}
	\centering
	\includegraphics[scale=.45]{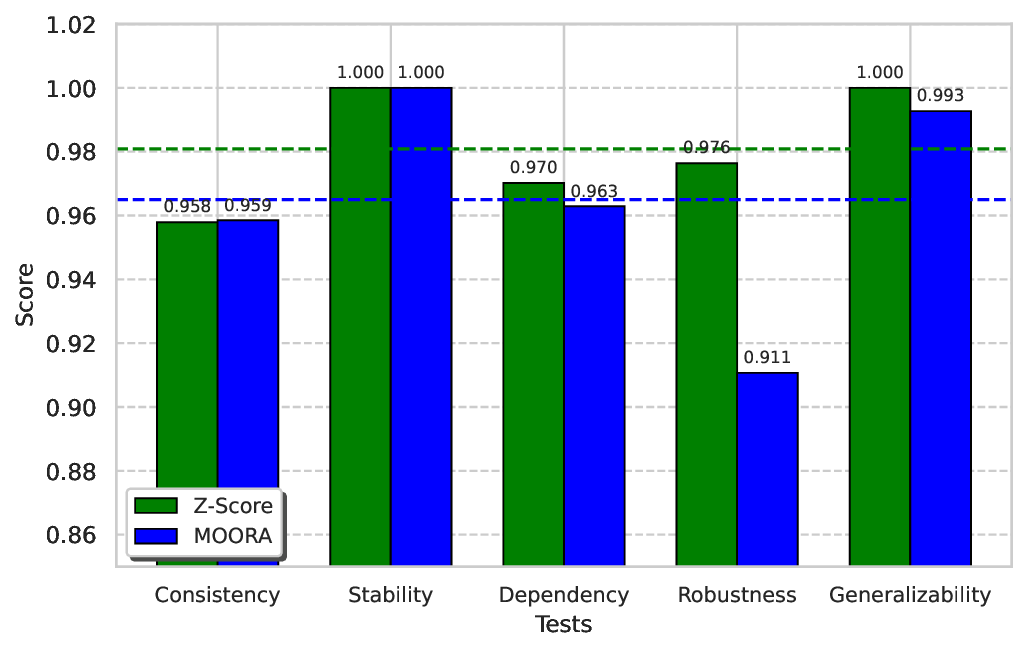}
    \caption{Comparison of Z-Score and MOORA aggregators on relation prediction. Z-Score outperforms MOORA on four of five criteria, with MOORA showing competitive results only on Stability.}
	\label{fig:p-test}
\end{figure}

This section provides the brief detail comparing the performance of Z-Score and MOORA aggregators across five evaluation criteria for relation and relation prediction tasks. 
For relation prediction (Figure \ref{fig:p-test}), Z-Score achieved a higher mean performance (M = 0.9809) compared to MOORA (M = 0.9650), outperforming it on three criteria: independence, robustness, and generalizability, while tying on stability. However, statistical tests confirmed that this difference could easily happen by chance (p = 0.273). Even so, the moderate effect size (Cohen's d = 0.564) shows that Z-Score still has a meaningful real-world advantage, especially in robustness.

\end{document}